\documentclass[10pt,twocolumn,letterpaper]{article}
\usepackage{titling}
\usepackage[pagenumbers]{cvpr} %

\usepackage{graphicx}
\usepackage{amsmath}
\usepackage{amssymb}
\usepackage{booktabs}
\usepackage{placeins}
\usepackage{stfloats}

\usepackage[utf8]{inputenc}

\usepackage{graphicx}
\usepackage[linesnumbered,ruled]{algorithm2e}

\usepackage[colorlinks]{hyperref}

\usepackage[capitalize, sort]{cleveref}
\crefname{section}{Sec.}{Secs.}
\Crefname{section}{Section}{Sections}
\Crefname{table}{Table}{Tables}
\crefname{table}{Tab.}{Tabs.}

\def\cvprPaperID{641} %
\def\confName{CVPR}
\def\confYear{2022}

\usepackage{comment}
\usepackage{multirow,bigdelim}
\usepackage{lipsum}
\usepackage{array}
\usepackage{wrapfig}

\usepackage{adjustbox}
\usepackage[percent]{overpic}
\usepackage{makecell}
\usepackage[normalem]{ulem}
\usepackage{blindtext}
\usepackage{xcolor}
\usepackage{soul}

\newcolumntype{C}[1]{>{\centering\let\newline\\\arraybackslash\hspace{0pt}}m{#1}}

\newif\ifdraft
\drafttrue

\ifdraft
\newcommand{\dcc}[1]{{\color{red}[\textbf{DC:} #1]}}
\newcommand{\rgc}[1]{{\color{purple}[\textbf{RG:} #1]}}
\newcommand{\opc}[1]{{\color{blue}[\textbf{OP:} #1]}}
\newcommand{\hmc}[1]{{\color{teal}[\textbf{HM} #1]}}

\newcommand{\abc}[1]{{\color{green}[\textbf{AB:} #1]}}

\newcommand{\drop}[1]{}

\else
\newcommand{\dcc}[1]{}
\newcommand{\rgc}[1]{}
\newcommand{\opc}[1]{}
\newcommand{\gcc}[1]{}
\newcommand{\hmc}[1]{}
\newcommand{\abc}[1]{}

\fi

\def\naive{na\"{\i}ve\xspace}
\def\Naive{Na\"{\i}ve\xspace}

\newcommand{\w}{$\mathcal{W}$\xspace}
\newcommand{\wplus}{$\mathcal{W}+$\xspace}

\makeatletter
\DeclareRobustCommand\onedot{\futurelet\@let@token\@onedot}
\def\@onedot{\ifx\@let@token.\else.\null\fi\xspace}

\def\eg{\emph{e.g}\onedot}

\def\ie{\emph{i.e}\onedot}

\def\etal{\emph{et al}\onedot}
\makeatother

\usepackage[bottom]{footmisc}
\makeatletter
\def\blfootnote{\xdef\@thefnmark{}\@footnotetext}
\makeatother

\newcommand{\ccbync}{\href{https://creativecommons.org/licenses/by-nc/4.0/legalcode}{CC BY-NC 4.0}}

\newcommand{\ccbyncsa}{\href{https://creativecommons.org/licenses/by-nc-sa/4.0/}{CC BY-NC-SA 4.0}}

\newcommand{\nvsrc}{\href{https://nvlabs.github.io/stylegan2/license.html
}{Nvidia Source Code License-NC}}

\newcommand{\bsd}{\href{https://opensource.org/licenses/BSD-3-Clause}{BSD 3-Clause}}

\newcommand{\mitlic}{\href{https://opensource.org/licenses/MIT}{MIT License}}

\newcommand{\adblic}{\href{https://github.com/utkarshojha/few-shot-gan-adaptation/blob/main/LICENSE.txt}{Adobe Research License}}

\title{StyleGAN-NADA: CLIP-Guided Domain Adaptation of Image Generators }

\author{Rinon Gal$^{1,2}$\thanks{\,\,Work was done during an internship at NVIDIA} \hspace{0.07\linewidth} \and
Or Patashnik$^{1}$ \hspace{0.07\linewidth} \and
Haggai Maron$^{2}$ \hspace{0.07\linewidth} \and 
Amit Bermano$^{1}$ \and
\hspace{0.1\linewidth} Gal Chechik$^{2}$ \hspace{0.2\linewidth} \and
Daniel Cohen-Or$^{1}$ \hspace{0.1\linewidth} \and \hspace{0.9\linewidth}
\and $^{1}$Tel-Aviv University \and $^{2}$NVIDIA
}

\begin{document}

\maketitle

\begin{abstract}

Can a generative model be trained to produce images from a specific domain, guided only by a text prompt, without seeing any image? In other words: can an image generator be trained ``blindly"?
Leveraging the semantic power of large scale Contrastive-Language-Image-Pre-training (CLIP) models, we present a text-driven method that allows shifting a generative model to new domains, without having to collect even a single image.
We show that through natural language prompts and a few minutes of training, our method can adapt a generator across a multitude of domains characterized by diverse styles and shapes. Notably, many of these modifications would be difficult or outright impossible to reach with existing methods.
We conduct an extensive set of experiments across a wide range of domains. These demonstrate the effectiveness of our approach, and show that our models preserve the latent-space structure that makes generative models appealing for downstream tasks.

\vspace{-5pt}
\end{abstract}

\section{Introduction}
\label{sec:intro}

\blfootnote{Code and videos available at: \href{https://stylegan-nada.github.io/}{stylegan-nada.github.io/}}

The unprecedented ability of Generative Adversarial Networks (GANs) \cite{goodfellow2014generative} to capture and model image distributions through a semantically-rich latent space has revolutionized countless fields. These range from image enhancement \cite{yang2021gan,ledig2017photo}, editing \cite{shen2020interpreting,harkonen2020ganspace} and recently even discriminative tasks such as classification and regression \cite{xu2021generative,nitzan2021large}.

Typically, the scope of these models is restricted to domains where one can collect large sets of images. This requirement severely constrains their applicability. 
Indeed, in many cases (paintings by specific artists, rare medical conditions, imaginary scenes), there may not exist sufficient data to train a GAN, or even any data at all.

\begin{figure}
    \centering
    \setlength{\belowcaptionskip}{-3pt}
    \setlength{\tabcolsep}{1.5pt}
    \renewcommand{\arraystretch}{0.5}
    {\scriptsize
    \begin{tabular}{c c}

        \raisebox{0.1\textwidth}{\rotatebox[origin=t]{90}{\begin{tabular}{c@{}c@{}}Photo $\rightarrow$  \\ Fernando Botero Painting\end{tabular}}} &
        \includegraphics[width=0.95\linewidth]{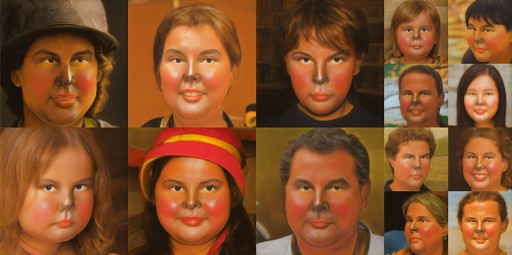} \\
        
        \raisebox{0.1\textwidth}{\rotatebox[origin=t]{90}{\begin{tabular}{c@{}c@{}}Photo of a Church $\rightarrow$  \\  Cryengine render of New York City\end{tabular}}} &
        \includegraphics[width=0.95\linewidth]{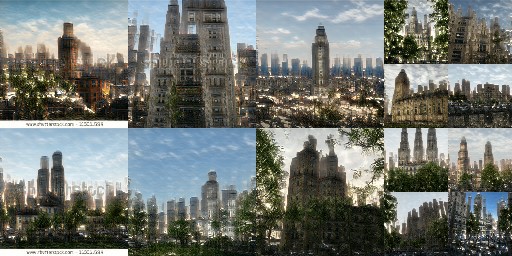} \\
        
        \raisebox{0.07\textwidth}{\rotatebox[origin=t]{90}{\begin{tabular}{c@{}c@{}}Photo $\rightarrow$ Painting \\ by Salvador Dali\end{tabular}}} &
        \includegraphics[width=0.95\linewidth]{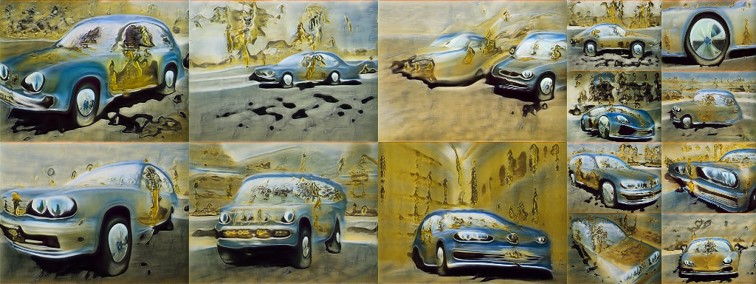} \\

    \end{tabular}
    }
    \caption{Examples of text-driven, out-of-domain generator adaptations induced by our method. The textual directions driving the change appear next to each set of generated images.}
    \label{fig:teaser}
    \vspace{-4pt}
\end{figure}

Recently, it has been shown that Vision-Language models (CLIP,~\cite{radford2021learning}) encapsulate generic information that can bypass the need for collecting data. Moreover, these models can be paired with generative models to provide a simple and intuitive text-driven interface for image generation and manipulation \cite{patashnik2021styleclip}.
However, such works are built on pre-trained generative models with fixed domains, limiting the user to \emph{in-domain} generation and manipulation.

In this work, we present a text-driven method that enables \emph{out-of-domain} generation. We introduce a training scheme that shifts the domain of a pre-trained model towards a new domain, using nothing more than a textual prompt. 
\cref{fig:teaser} demonstrates out-of-domain generation for three examples. 
All three models were trained \emph{blindly}, without seeing any image of the target domain. %

Leveraging CLIP for text-guided training is challenging. The \naive approach -- requiring generated images to maximize some CLIP-based classification score, often leads to adversarial solutions \cite{goodfellow2014explaining} (see \cref{sec:experiments}). Instead, we encode the difference between domains as a textually-prescribed direction in CLIP's embedding space. We propose a novel loss and a two-generator training architecture. One generator is kept frozen, providing samples from the source domain. The other is optimized to produce images that differ from this source only along a textually-described, cross-domain direction in CLIP-space.

To increase training stability for more drastic domain changes, we introduce an adaptive training approach that utilizes CLIP to identify and restrict training only to the most relevant network layers at each training iteration.

We apply our approach to StyleGAN2\cite{karras2020analyzing}, and demonstrate its effectiveness for a wide range of source and target domains. These include artistic styles, cross-species identity transfer and significant shape changes, like converting dogs to bears.
We compare our method to existing editing techniques and alternative few-shot approaches and show that it enables modifications which are beyond their scope -- and it does so without direct access to \textit{any} training data.

Finally, we verify that our method maintains the appealing structure of the latent space. Our shifted generator not only inherits the editing capabilities of the original, but it can even re-use any off-the-shelf editing directions and models trained for the original domain.

\vspace{-0pt}
\section{Related work}
\label{sec:related}
\paragraph{Text-guided synthesis.}
Vision-language tasks include language-based image retrieval, image captioning, visual question answering, and text-guided synthesis among others. Typically, to solve these tasks, a cross-modal vision and language representation is learned~\cite{Desai2020VirTexLV, sariyildiz2020learning, Tan2019LXMERTLC, Lu2019ViLBERTPT, Li2019VisualBERTAS, Su2020VL-BERT:, Li2020UnicoderVLAU, Chen2020UNITERUI, Li2020OscarOA}, often by training a transformer~\cite{NIPS2017_3f5ee243}. 

The recently introduced CLIP~\cite{radford2021learning} is a powerful model for joint vision-language representation. It is trained on 400 million text-image pairs. 
Using a contrastive learning goal, both image and text are mapped
into a joint, multi-modal embedding space. The representations learned by CLIP have been used in guiding a variety of tasks, including image synthesis \cite{murdock2021bigsleep,katherine2021vqganclip} and manipulation \cite{patashnik2021styleclip,bau2021paint}.
These methods utilize CLIP to guide the optimization of a latent code, generating or manipulating a specific image. In contrast, we present a novel approach in which a text prompt guides the \emph{training} of the image generator itself.

\vspace{-4pt}
\paragraph{Training generators with limited data.}
The goal of few-shot generative models is to mimic a rich and diverse data distribution using only a few images. Methods used to tackle such a task can be divided into two broad categories: training from-scratch, and fine-tuning --- which leverages the diversity of a pre-trained generator.

Among those that train a new generator, `few' often denotes several thousand images (rather than tens of thousands \cite{karras2020analyzing} or millions \cite{brock2018large}). 
Such works typically employ data augmentations \cite{tran2021data,zhao2020differentiable,zhao2020image,Karras2020ada} or empower the discriminator to better learn from existing data using auxiliary tasks \cite{yang2021data,liu2020towards}.

In the transfer-learning scenario, `few' typically refers to 
numbers
ranging from several hundred to as few as five~\cite{ojha2021few}. %
When training with extremely limited data, a primary concern is staving off mode-collapse or overfitting, to successfully transfer the diversity of the source generator to the target domain. Multiple methods have been proposed to tackle these challenges. Some place restrictions on the space of modified weights \cite{mo2020freeze,Robb2020FewShotAO,pinkney2020resolution}. Others introduce new parameters to control channel-wise statistics \cite{noguchi2019image}, steer sampling towards suitable regions of the latent space \cite{Wang_2020_CVPR}, add regularization terms \cite{Tseng2021RegularizingGA,li2020few} or enforce cross-domain consistency while adapting to a target style \cite{ojha2021few}.

While prior methods adapt generators with \textit{limited} data,  we do so \textit{without} direct access to \textit{any} training images. Additionally, prior methods constrained the space of trainable weights to fixed, hand-picked subsets. Our method introduces adaptive layer selection - accounting for both the state of the network at each training step, and for the target class.

\section{Preliminaries}
\label{sec:preliminary}

At the core of our approach are two components - StyleGAN2 \cite{karras2020analyzing} and CLIP \cite{radford2021learning}. In the following, we discuss relevant features in StyleGAN's architecture, and how CLIP has been employed in this context in the past.

\subsection{StyleGAN}

In recent years, StyleGAN and its variants \cite{karras2019style, karras2020analyzing, Karras2020ada,karras2021aliasfree} have established themselves as the state-of-the-art unconditional image generators. 
The StyleGAN generator consists of two main components. A mapping network converts a latent code $z$, sampled from a Gaussian prior, to a vector $w$ in a learned latent space \w. These latent vectors are then fed into 
the synthesis network, to control feature (or convolutional kernel) statistics.
By traversing 
\w , or by mixing different $w$ codes at different network layers, prior work demonstrated fine-grained control over semantic properties in generated images \cite{shen2020interpreting,harkonen2020ganspace,10.1145/3447648,patashnik2021styleclip}. These latent-space editing operations, however, are typically limited to modifications that align with the domain of the initial training set.

\subsection{StyleCLIP}

In a recent work, Patashnik \etal \cite{patashnik2021styleclip} combine the generative power of StyleGAN with the semantic knowledge of CLIP to discover novel editing directions, using only a textual description of the desired change. 
They outline three approaches for leveraging the semantic power of CLIP. The first two aim to minimize the CLIP-space distance between a generated image and some target text. They use direct latent code optimization, or train an encoder (or \textit{mapper}) to modify an input latent code. The third approach, which we adapt, uses CLIP to discover global directions of disentangled change in the latent space. They modify individual latent-code entries, and determine which ones induce an image-space change that is co-linear with the direction between two textual descriptors (denoting the source and desired target) in CLIP-space.

However, these approaches all share the limitation common to latent space editing methods - the modifications that they can apply are largely constrained to the domain of the pre-trained generator. As such, they can allow for changes in hairstyle, expressions, or even convert a wolf to a lion \textit{if the generator has seen both} - but they cannot convert a photo to a painting, or produce cats when trained on dogs.

\vspace{-2pt}
\begin{figure}[t]
\setlength{\tabcolsep}{1pt}
    \centering
    \includegraphics[width=0.99\linewidth]{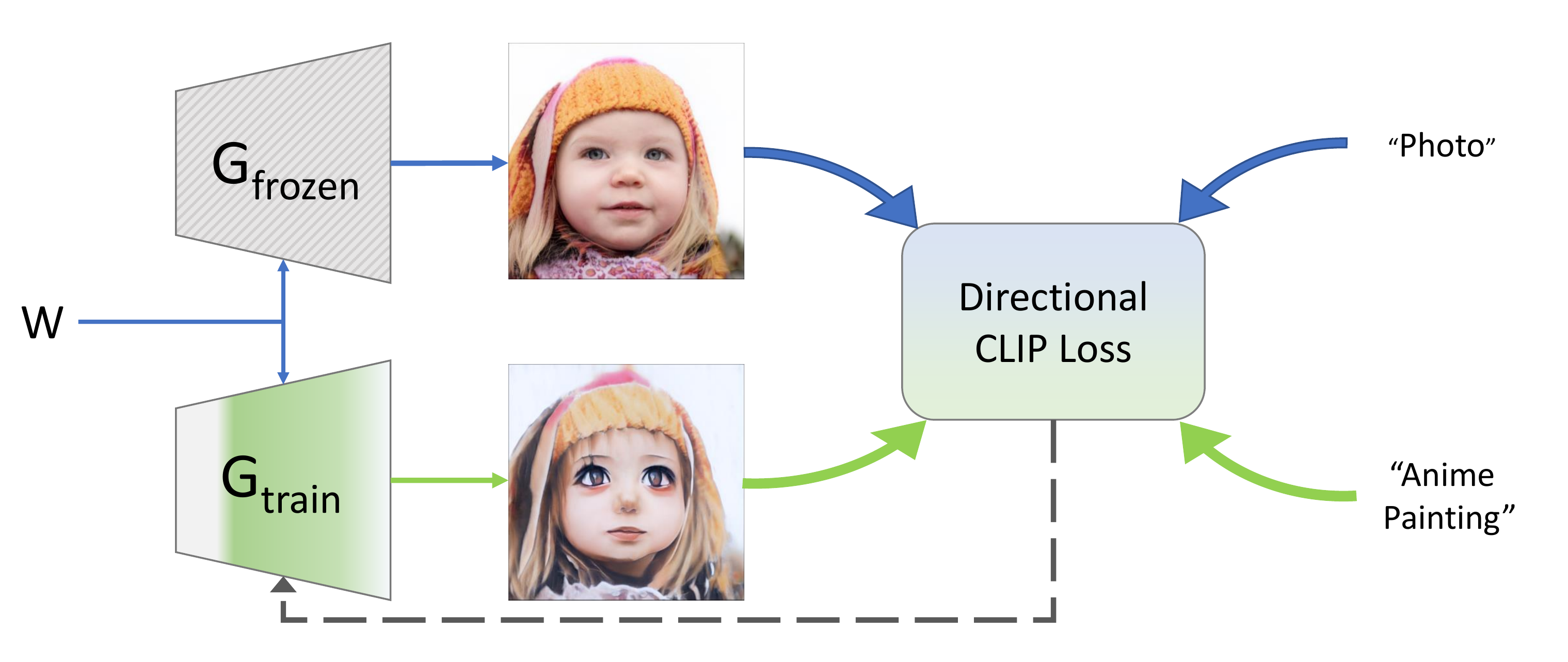}\vspace{-0pt}
    \caption{Overview of our training setup. We initialize two %
    generators - $G_{frozen}$ and $G_{train}$ using the weights of a generator pre-trained on images from a source domain (\eg FFHQ \cite{karras2019style}). $G_{frozen}$ remains fixed throughout the process. $G_{train}$ is modified through optimization and an iterative layer-freezing scheme. The process shifts the domain of $G_{train}$ according to a user-provided textual direction while maintaining a shared latent space.}
    \label{fig:arch}
    \vspace{-5pt}
\end{figure}

\section{Method}
\label{sec:method}

Our goal is to shift a pre-trained generator from a given source domain to a new target domain, described only through textual prompts, with no images. As a source of supervision, we use only a pre-trained CLIP model.

We approach the task through two key questions: (1) How can we best distill the semantic information encapsulated in CLIP? and (2) How should we regularize the optimization process to avoid adversarial solutions and mode collapse? In the following section we outline a training scheme and a loss that seek to answer both questions.

\subsection{CLIP-based guidance}
\paragraph{Global loss.}
We rely on a pre-trained CLIP model to serve as the sole source of supervision for our target domain. {\Naive}ly, one could use StyleCLIP's direct loss:\vspace{-2pt}
\begin{equation}\label{eq:global_loss}
    \mathcal{L}_{global} = D_{CLIP}\left(G\left(w\right), t_{target}\right)~,
\end{equation}\vspace{-2pt}%
where $G\left(w\right)$ is the image generated by the latent code $w$ fed to the generator $G$, $t_{target}$ is the textual description of the target class, and $D_{CLIP}$ is the CLIP-space cosine distance. We name this loss `global', as it 
does not depend on the initial image or domain.

We observe that in practice, this loss leads to adversarial solutions. In the absence of a fixed generator that favors solutions on the real-image manifold, optimization overcomes the classifier (CLIP) by adding pixel-level perturbations to the image. Moreover, this loss sees no benefit from maintaining diversity. Indeed, a mode-collapsed generator producing only one image
may be the best minimizer for the distance to a given textual prompt. 
See \cref{sec:clip_analysis} for an analysis of CLIP's embedding space, demonstrating this issue. These shortcomings make this loss unsuitable for training the generator. However, we do leverage it for adaptive layer selection %
(see \cref{subsec:layer_freezing}).

\vspace{-7pt}
\paragraph{Directional CLIP loss.} To address these issues, we draw inspiration from StyleCLIP's global direction approach. Ideally, we want to identify the CLIP-space direction between our source and target domains. Then, we'll fine-tune the generator so that the images it produces differ from the source \textit{only} along this direction. 

To do so, we first identify a cross-domain direction in CLIP-space by embedding a pair of textual prompts describing the source and target domains (\eg ``Dog" and ``Cat"). Then, we must determine the CLIP-space direction between images produced by the generator before and after fine-tuning. We do so using a two-generator setup. 
We begin with a generator pre-trained on a single source domain (\eg faces, dogs, churches or cars), and clone it. One copy is kept frozen throughout the training process. Its role is to provide an image in the source domain for every latent code. The second copy is trained. 
It is fine-tuned to produce images that, for any sampled code, differ from the source generator's only along the textually described direction. We name these generators $G_{frozen}$ and $G_{train}$ respectively.

In order to maintain latent space alignment, the two generators share a single mapping network which is kept frozen throughout the process. The full training setup is shown in \cref{fig:arch}. An illustration of the CLIP-space directions is provided in \cref{fig:direction}.
The direction loss is given by:
\begin{figure}[t]
\setlength{\tabcolsep}{1pt}
    \centering
    \includegraphics[width=0.8\linewidth]{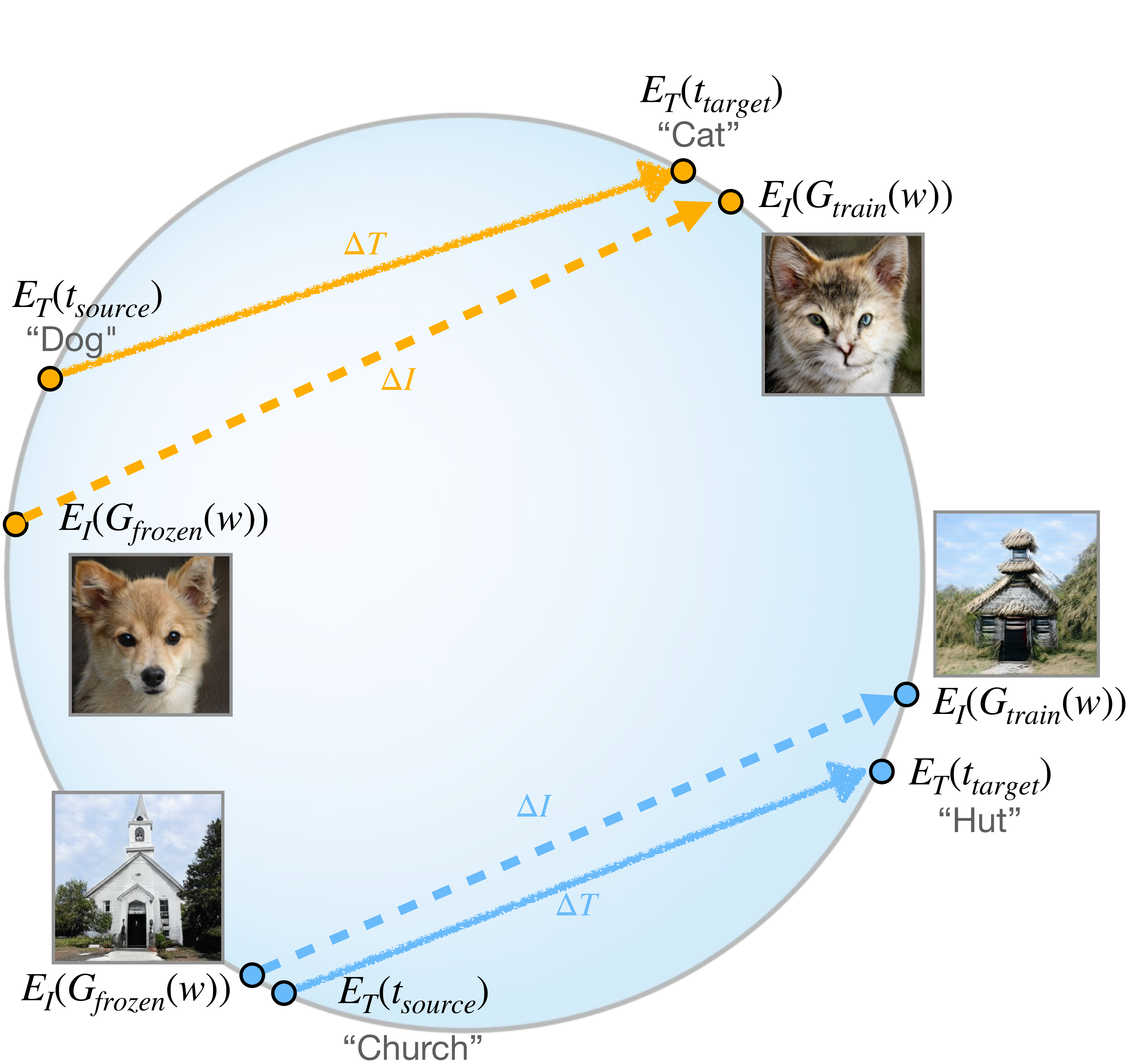}
    \caption{Illustration of CLIP-space directions in our directional loss (\cref{eq:directional_loss}). We embed images generated by both generators into CLIP-space and demand that the vector connecting them, $\Delta I$, is parallel to the vector connecting a source and target text, $\Delta T$.}
    \label{fig:direction}
    \vspace{-8pt}
\end{figure}

\vspace{-1pt}
\begin{equation}\label{eq:directional_loss}
\begin{gathered}
\Delta T = E_{T}\left(t_{target}\right) - E_{T}\left(t_{source}\right)~, \\
\Delta I = E_{I}\left(G_{train}\left(w\right)\right) - E_{I}\left(G_{frozen}\left(w\right)\right)~, \\ 
\mathcal{L}_{direction} = 1 - \frac{\Delta I \cdot \Delta T}{\left|\Delta I\right|\left|\Delta T\right|}~.
\end{gathered}
\end{equation}%
\vspace{-2pt}%
Here, $E_{I}$ and $E_{T}$ are CLIP's image and text encoders, %
and $t_{target}$, $t_{source}$ are the source and target class texts.

Such a loss overcomes the global loss' shortcomings: First, it is adversely affected by mode-collapsed examples. If the target generator only creates a single image, the CLIP-space directions from all sources to this target image will be different. As such, they can't all align with the textual direction. Second, it is harder for the network to converge to adversarial solutions, because it has to engineer perturbations that fool CLIP across an infinite set of different instances.

\subsection{Layer-Freezing}\label{subsec:layer_freezing}

For domain shifts which are predominantly texture-based, such as converting a photo to a sketch, the training scheme described above quickly converges before mode collapse or overfitting occurs. However, more extensive shape modifications require longer training, which in turn destabilizes the network and leads to poor results.

Prior works on few-shot domain adaptation observed that the quality of synthesized results can be significantly improved by restricting training to a subset of network weights \cite{mo2020freeze,Robb2020FewShotAO}. 
The intuition is that some layers of the source generator are useful for generating aspects of the target domain, so we want to preserve them. Furthermore, optimizing fewer parameters reduces the model complexity and the risk of overfitting. Following these approaches, we regularize the training process by limiting the number of weights that can be modified at each training iteration. 

Ideally, we would like to restrict training to those model weights that are most relevant to a given change. To identify these weights, we turn back to latent-space editing techniques, and specifically to StyleCLIP. %

In the case of StyleGAN, it has been shown that codes provided to different network layers, influence different semantic attributes. Thus, by considering editing directions in the \wplus space \cite{abdal2019image2stylegan} -- the latent space where each layer of StyleGAN can be provided with a different code $w_{i} \in \mathcal{W}$ -- we can identify which layers are most strongly tied to a given change. %
Building on this intuition, we suggest a training scheme, where at each iteration we (i) select the $k$-most relevant layers, and (ii) perform a single training step where we optimize only these layers, while freezing all others. 

\begin{figure}[t]
\setlength{\tabcolsep}{1pt}
    \centering
    \includegraphics[width=\linewidth]{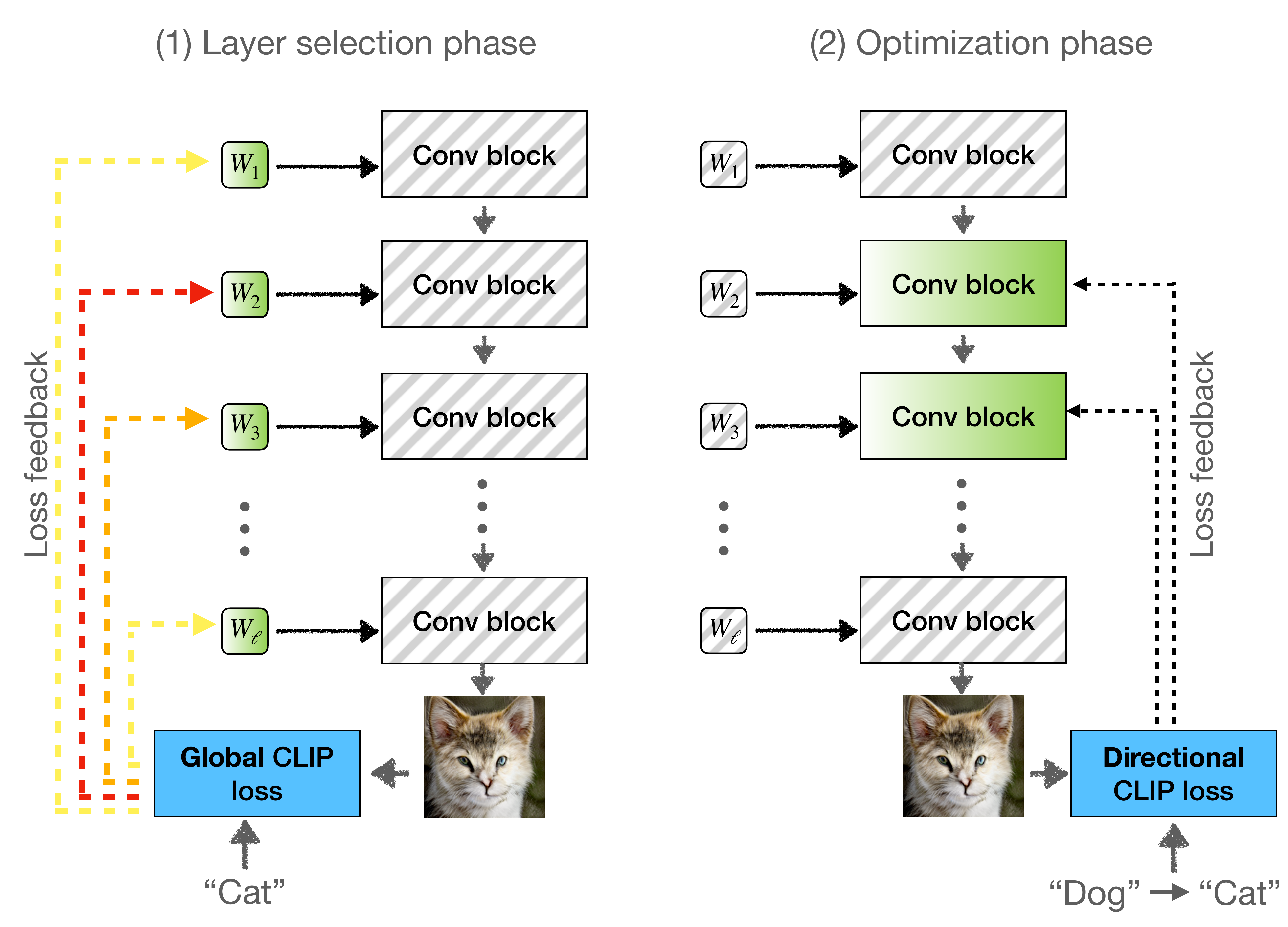}\vspace{-2pt}
    \caption{The adaptive layer-freezing mechanism has two phases. In the first phase (left), we optimize a set of latent codes in \wplus (turquoise), leaving all network weights fixed. This optimization is conducted using the Global CLIP loss (\cref{eq:global_loss}) . We select the layers whose corresponding $w$ entry changed most significantly (darker colors, left). In the second phase (right), we unfreeze the weights of the selected layers. We then optimize these layers using the directional CLIP loss (\cref{eq:directional_loss}).}
    \label{fig:layer_focus}
    \vspace{-8pt}
\end{figure}

To select the $k$ layers,  
we randomly sample $N_{w}$ latent codes $\in \mathcal{W}$ and convert them to \wplus by replicating the same code for each layer. We then perform $N_{i}$ iterations of the StyleCLIP \emph{latent-code} optimization method, %
using the global loss (\cref{eq:global_loss}). We select the $k$ layers for which the latent-code changed most significantly. 
The two-step process is illustrated in \cref{fig:layer_focus}. In all cases we additionally freeze StyleGAN's mapping network, affine code transformations, and all toRGB layers.

Note that this process is inherently different from selecting layers according to gradient magnitudes during direct training. Latent-code optimization using a \textit{frozen} generator tends to favor solutions which remain on the real-image manifold. By using it to select layers, we bias training towards similarly \textit{realistic} changes. In contrast, direct training allows the model more easily drift towards unrealistic or adversarial solutions.

\begin{figure*}
    \centering
    \setlength{\belowcaptionskip}{-3pt}
    \setlength{\tabcolsep}{1pt}
    {
    \begin{tabular}{c c}
        
        \includegraphics[width=0.5\textwidth]{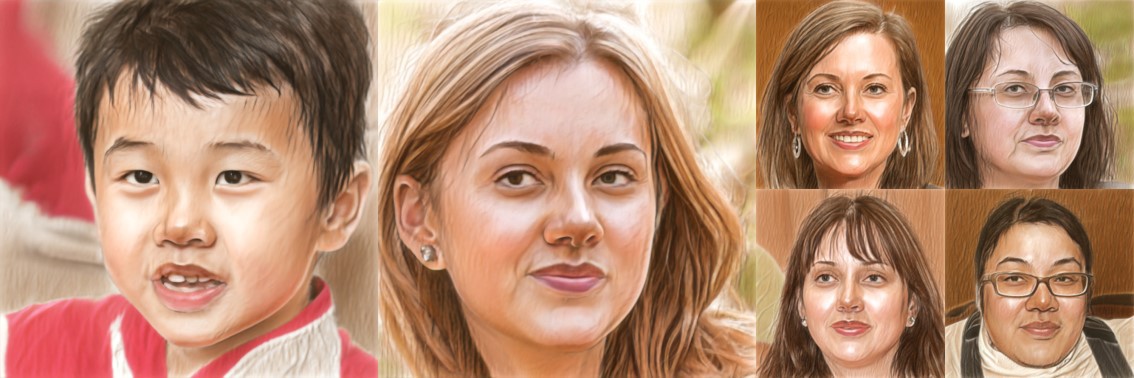} &
        
        \includegraphics[width=0.5\textwidth]{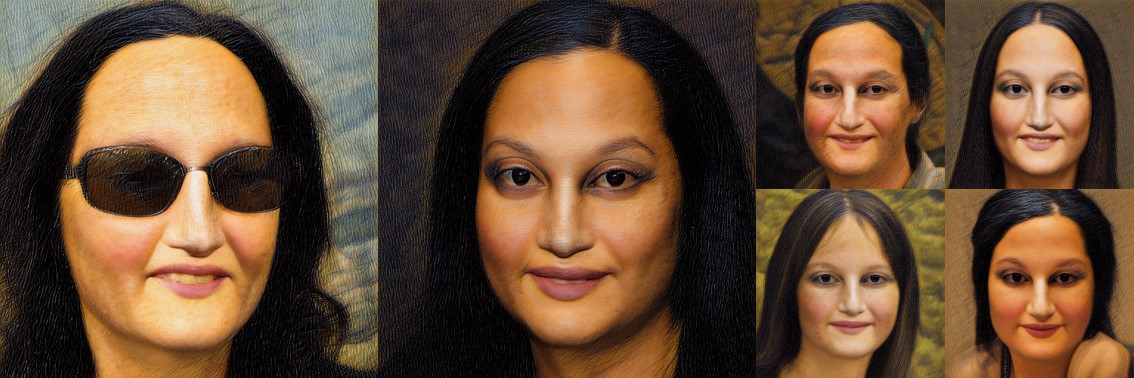} \\
        Photo $\rightarrow$  Sketch & Photo $\rightarrow$  Mona Lisa Painting \\
        \includegraphics[width=0.5\textwidth]{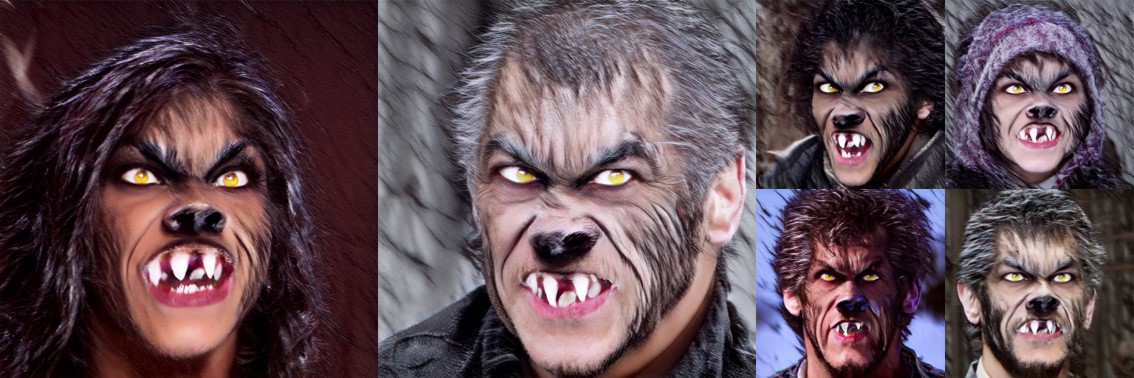} &
        
        \includegraphics[width=0.5\textwidth]{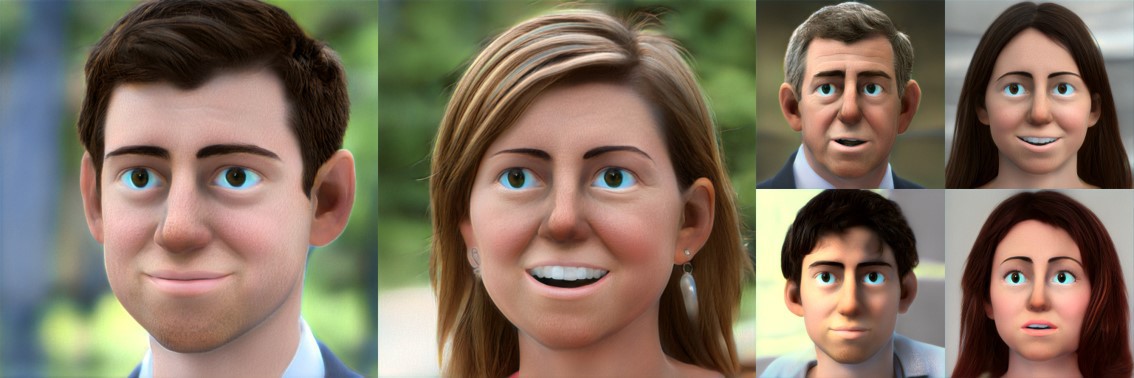} \\
        Human $\rightarrow$  Werewolf & Photo $\rightarrow$  3D Render in the Style of Pixar
    \end{tabular}
    }
    \vspace{-3pt}
    \caption{Image synthesis using models adapted from StyleGAN2-FFHQ \cite{karras2020analyzing} to a set of textually-prescriped target domains. All images were sampled randomly, using truncation with $\psi = 0.7$. The driving texts appear below each generated set.}
    \label{fig:generated_ffhq_small}
\end{figure*}

\subsection{Latent-Mapper `mining'}\label{subsec:mapper}

For some shape changes, we found that the generator does not complete a full transformation. For example, when transforming dogs to cats, the fine-tuning process results in a new generator that outputs both cats, dogs, and an assortment of images that lie in between. 
To alleviate this, we note that the shifted generator now includes \textit{both} cats and dogs within its domain. We thus turn to in-domain latent-editing techniques, specifically StyleCLIP's latent mapper, to map all latent codes into the cat-like region of the latent space. Training details for the mapper are provided \cref{sec:train_details}. %

\section{Experiments}
\label{sec:experiments}

\subsection{Results}
We begin by showcasing the wide range of out-of-domain adaptations enabled by our method. These range from style and texture changes to shape modifications, and from realistic to fantastical, including zero-shot prompts for which real data does not even exist (\eg Nicolas Cage dogs). All these are achieved through a simple text-based interface, and are typically trained within a few minutes.

\begin{figure*}[!hbt]
    \centering
    \setlength{\belowcaptionskip}{-3pt}
    \setlength{\tabcolsep}{1pt}
    {\footnotesize
    \begin{tabular}{c c c}
        \includegraphics[height=0.147\linewidth]{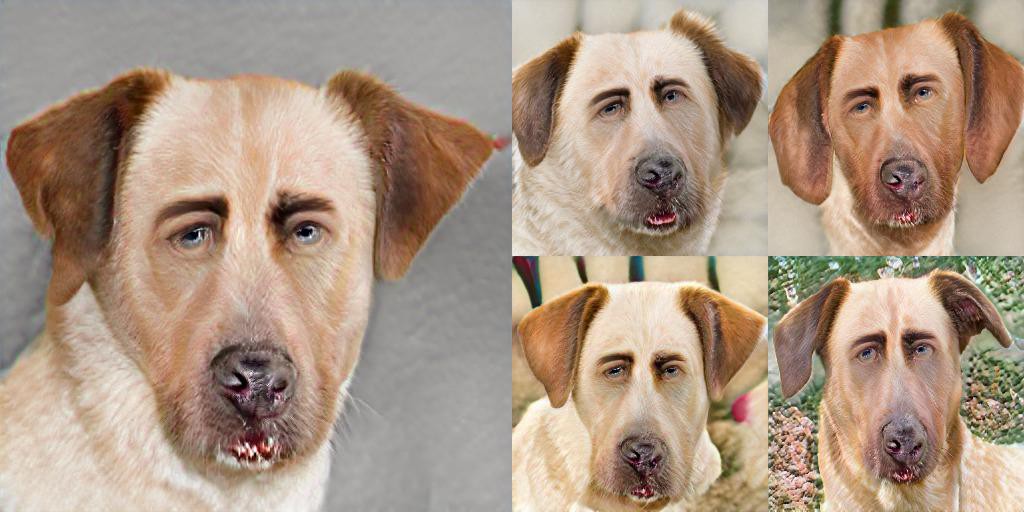} &
        \includegraphics[height=0.147\linewidth]{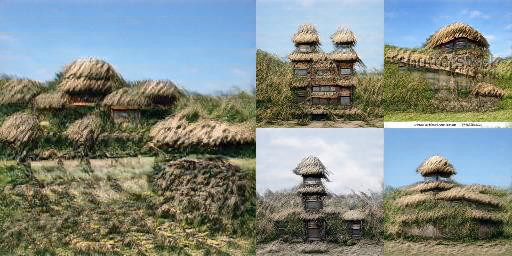} &
        \includegraphics[height=0.147\linewidth]{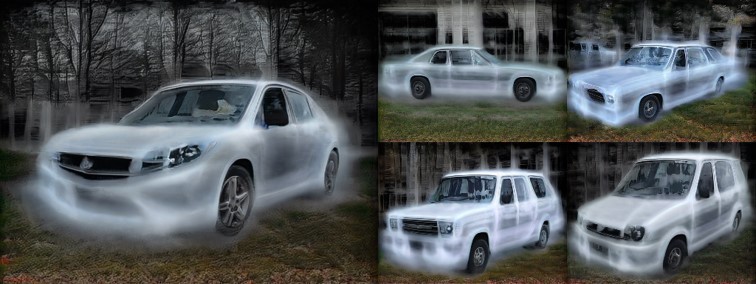} \\
        
        Dog $\rightarrow$ Nicolas Cage & Church $\rightarrow$ Hut & Car $\rightarrow$ Ghost car \\
        
        \includegraphics[height=0.147\linewidth]{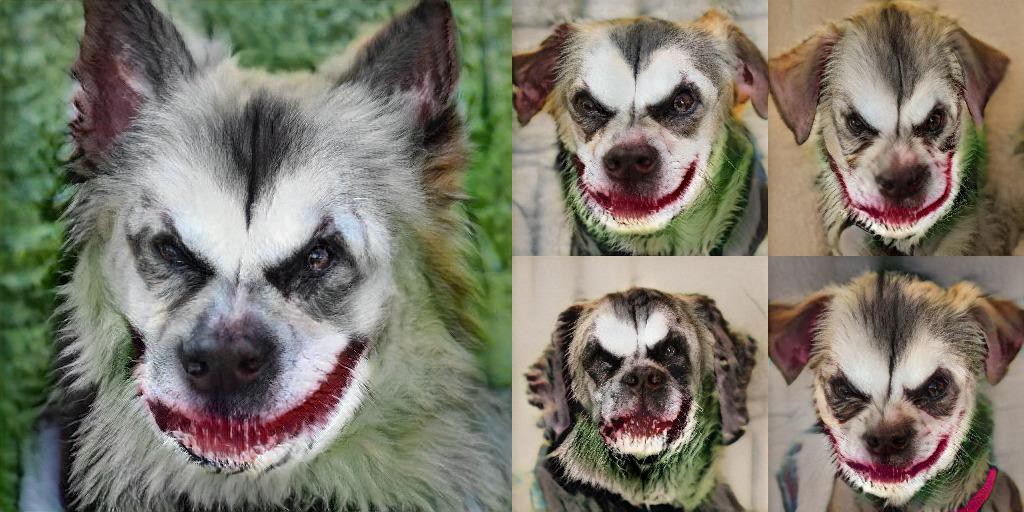} &
        \includegraphics[height=0.147\linewidth]{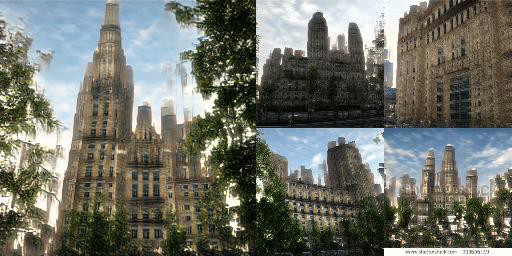} &        
        \includegraphics[height=0.147\linewidth]{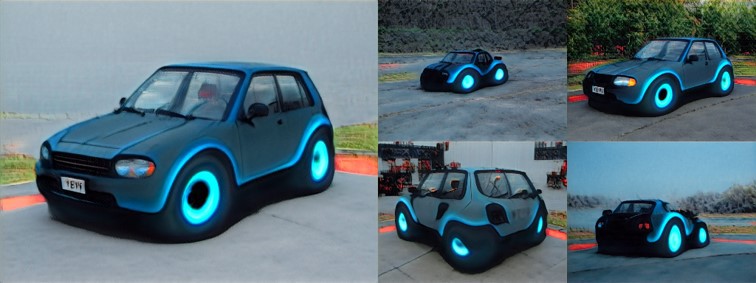} \\

        Dog $\rightarrow$ The Joker & {\begin{tabular}{c@{}c@{}} Photo of a church  $\rightarrow$ \\ Cryengine render of New York city \end{tabular}} & Chrome wheels $\rightarrow$ TRON wheels

    \end{tabular}
    }
    \vspace{-3pt}
    \caption{Image synthesis using models adapted from StyleGAN2's \cite{karras2020analyzing} LSUN Church, LSUN Car \cite{yu2015lsun} models and StyleGAN-ADA \cite{Karras2020ada} AFHQ-Dog \cite{choi2020starganv2}. All images were sampled randomly, using truncation with $\psi = 0.7$. The driving texts appear below each generated set.}
    \label{fig:generated_church_small}
\end{figure*}

In \cref{fig:generated_ffhq_small,fig:generated_church_small}, we show a series of randomly sampled images synthesized by generators converted from faces, churches, dogs and cars to various target domains.  Additional large scale galleries portraying a wide set of target domains are shown in \cref{sec:more_samples}.

\begin{figure*}[!hbt]
    \centering
    \setlength{\belowcaptionskip}{-1pt}
    \setlength{\tabcolsep}{1pt}
    {\footnotesize
    \begin{tabular}{c c c}
    
        \includegraphics[width=0.33\textwidth]{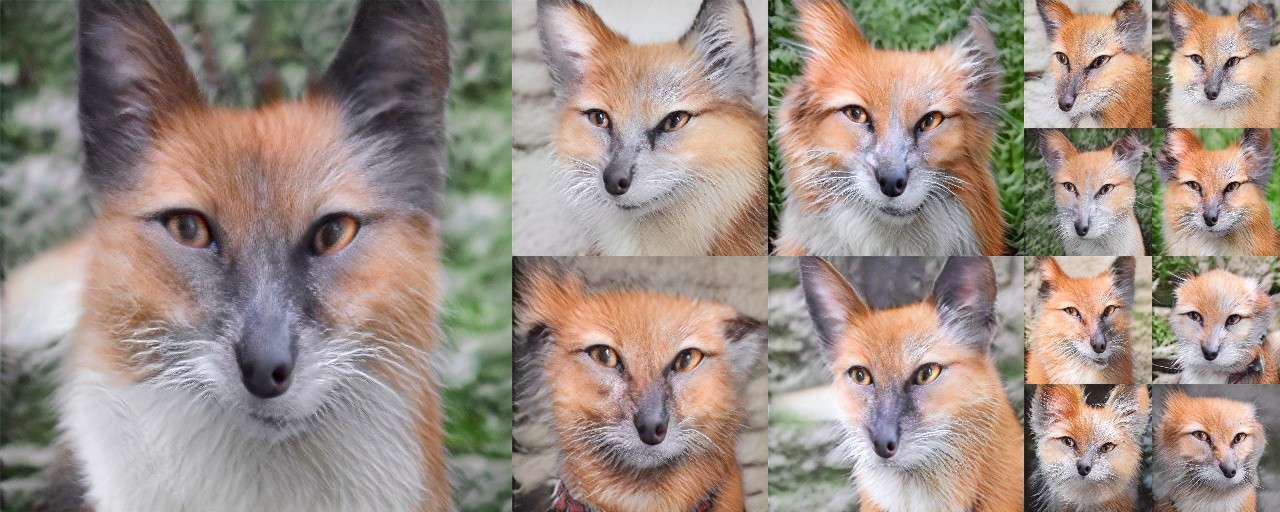} &
        \includegraphics[width=0.33\textwidth]{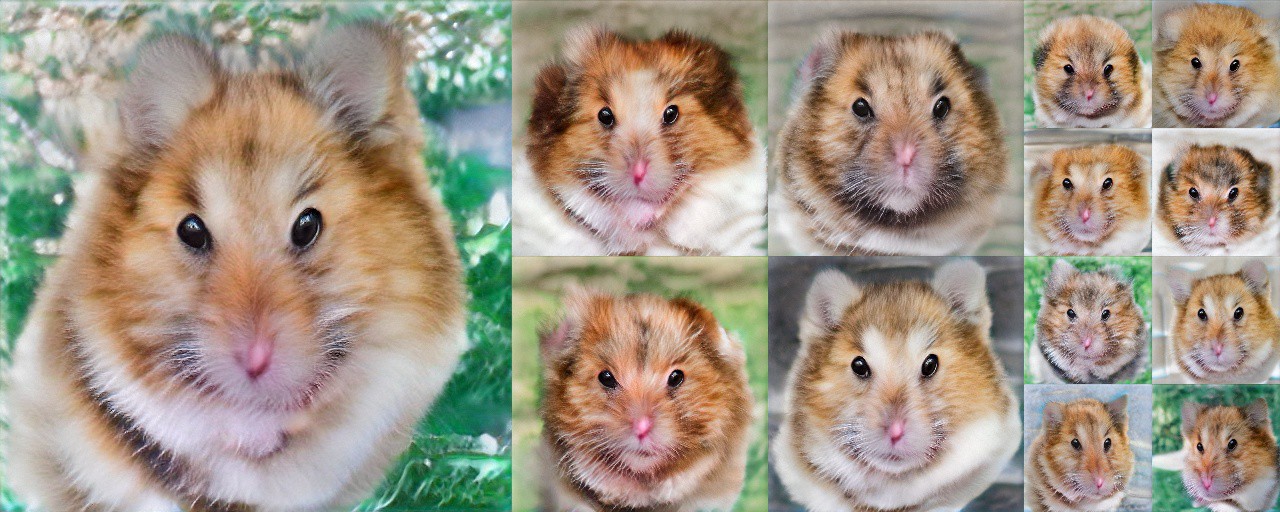} &  
        \includegraphics[width=0.33\textwidth]{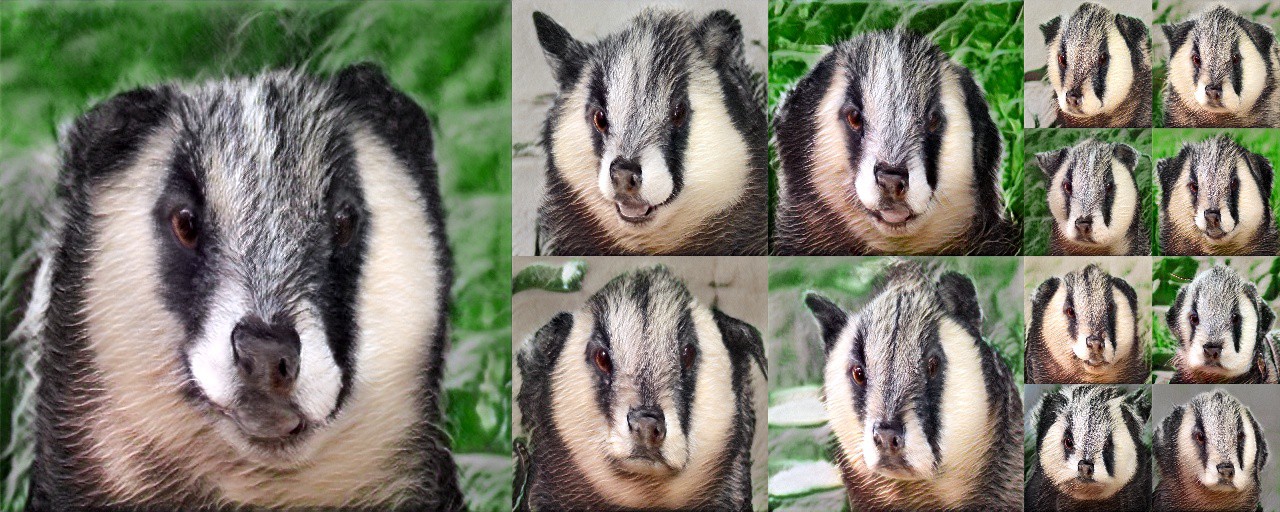} \\
        "Fox" & "Hamster" & "Badger" \\
        
        \includegraphics[width=0.33\textwidth]{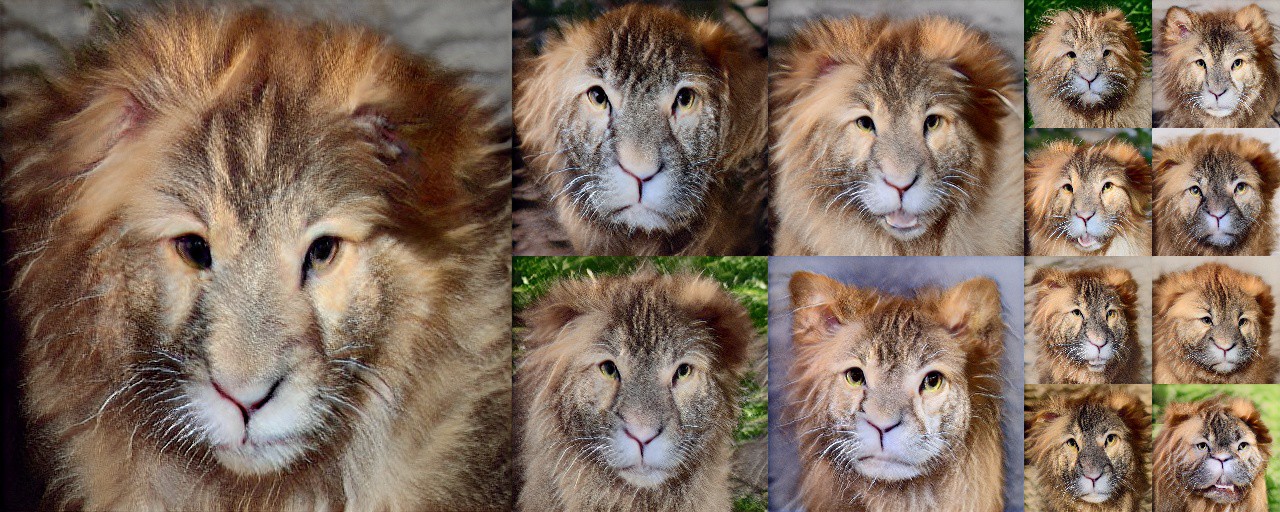} &
        \includegraphics[width=0.33\textwidth]{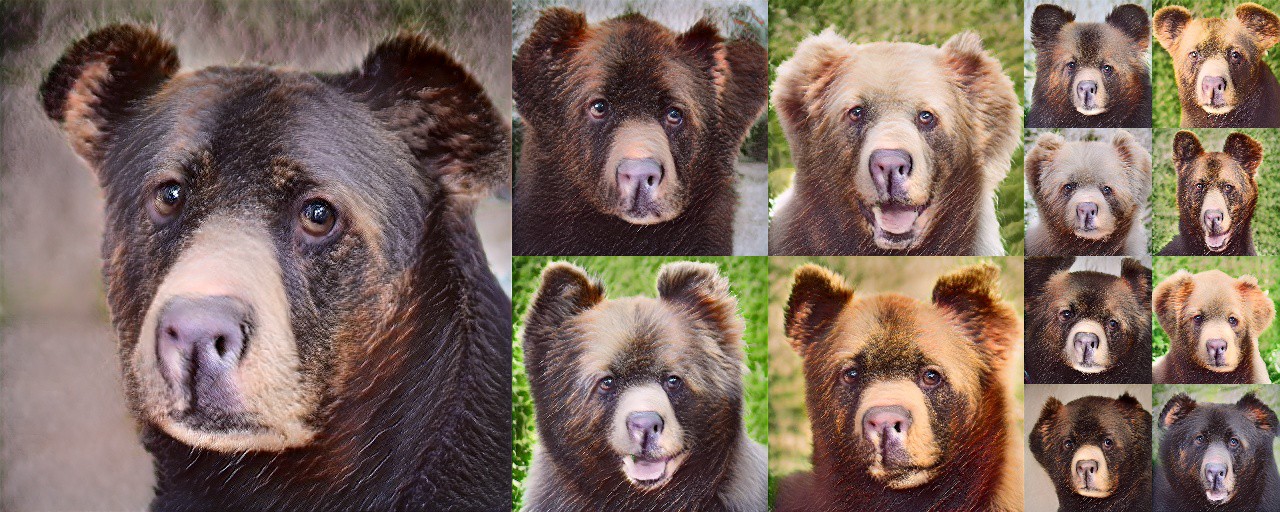} &
        \includegraphics[width=0.33\textwidth]{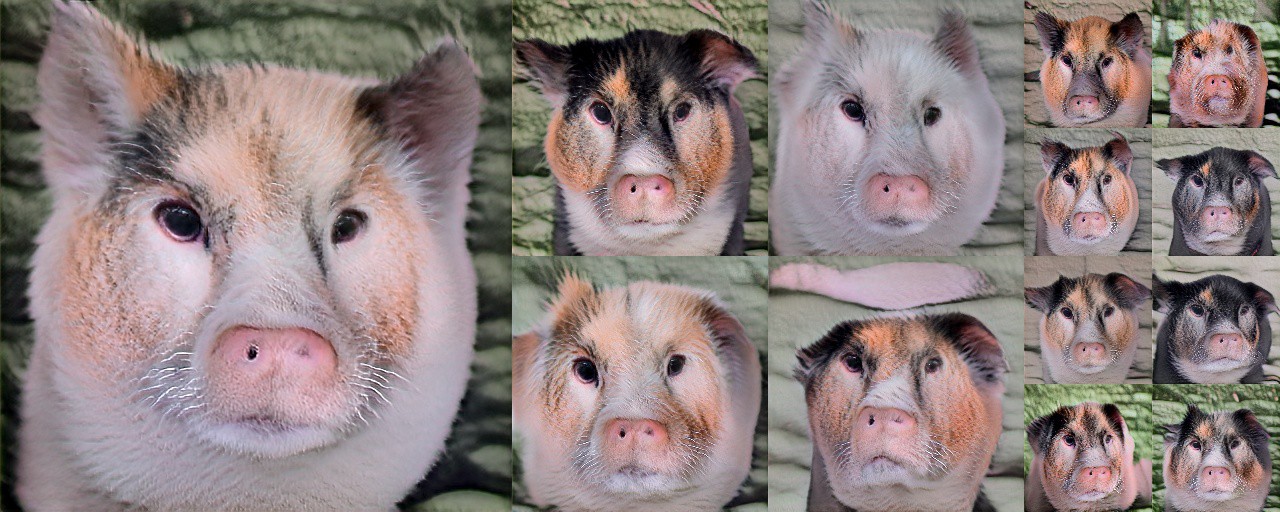} \\
        "Lion" & "Bear" & "Pig" \\

    \end{tabular}
    }
    \vspace{-6pt}
    \caption{Generator translation to multiple animal domains. In all cases we begin with a StyleGAN-ADA~\cite{Karras2020ada} AFHQ-Dog~\cite{choi2020starganv2} model. All generators are adapted using our method and a StyleCLIP \cite{patashnik2021styleclip} latent mapper. For all experiments, the source domain text was `Dog'. The target domain text is shown below each image. }
    \label{fig:animals}\vspace{-3pt}
\end{figure*}

\cref{fig:animals} shows domain adaptation from dogs to a wide range of animals.  While Figs.~\ref{fig:generated_ffhq_small},~\ref{fig:generated_church_small}
focused on style and minor shape adjustments, here the model performs significant shape modifications. For example, many animals sport upright ears, while most AFHQ-Dog~\cite{choi2020starganv2} breeds do not.
Training details for all scenarios are provided in \cref{sec:train_details}.

\vspace{-8pt}
\subsection{Latent space exploration}
 Modern image generators (and StyleGAN in particular), are known to have a well-behaved latent space. Such a latent space is conductive for tasks such as image editing and image-to-image translation \cite{shen2020interpreting, harkonen2020ganspace, patashnik2021styleclip, richardson2020encoding, alaluf2021matter}. The ability to manipulate real images is of particular interest, leading to an ever-increasing list of methods for GAN inversion \cite{abdal2019image2stylegan,tov2021designing,richardson2020encoding, alaluf2021restyle,xia2021gan}. We show that our transformed generators can still support such manipulations, using the same techniques and inversion methods. Indeed, as outlined below, our model can even reuse off-the-shelf models pre-trained on the source generator's domain, with no need for additional fine-tuning.

\begin{figure}[!hbt]\vspace{-0pt}
    \centering
    \setlength{\belowcaptionskip}{-2.5pt}
    \setlength{\tabcolsep}{1pt}
    {\scriptsize
    \begin{tabular}{c c c c c}
    
        \includegraphics[width=0.19\linewidth]{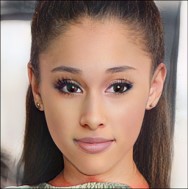} &
        \includegraphics[width=0.19\linewidth]{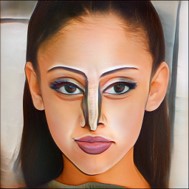} &
        \includegraphics[width=0.19\linewidth]{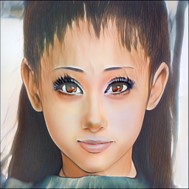} &
        \includegraphics[width=0.19\linewidth]{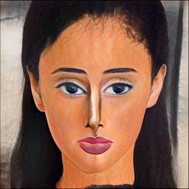} &
        \includegraphics[width=0.19\linewidth]{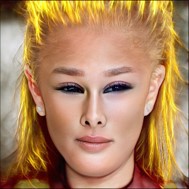}
        \\
        
        \includegraphics[width=0.19\linewidth]{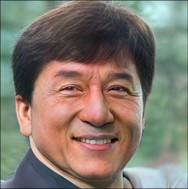} &
        \includegraphics[width=0.19\linewidth]{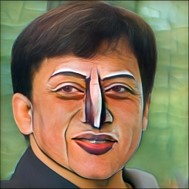} &
        \includegraphics[width=0.19\linewidth]{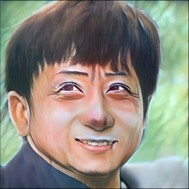} &
        \includegraphics[width=0.19\linewidth]{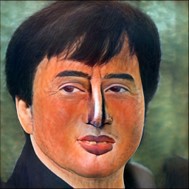} &
        \includegraphics[width=0.19\linewidth]{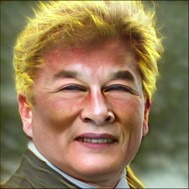}
        \\
        
        \includegraphics[width=0.19\linewidth]{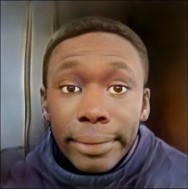} &
        \includegraphics[width=0.19\linewidth]{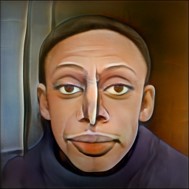} &
        \includegraphics[width=0.19\linewidth]{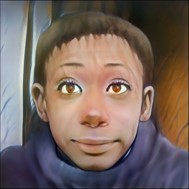} &
        \includegraphics[width=0.19\linewidth]{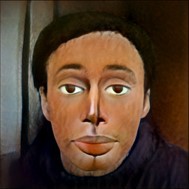} &
        \includegraphics[width=0.19\linewidth]{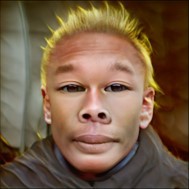}\vspace{-6pt}
        \\
        \\
        \vspace{-6pt}
        Inverted & Cubism & Anime & Modigliani & Super Saiyan

    \end{tabular}}
    \vspace{-2pt}
    \caption{Out-of-domain editing through latent-code equivalence between generators. We invert an image into the latent space of a StyleGAN2 FFHQ model \cite{karras2020analyzing}, using a pre-trained ReStyle encoder \cite{alaluf2021restyle}. We then feed the same latent code into the transformed generators in order to map the same identity to a novel domain. }
    \label{fig:real_edits}
\end{figure}

\vspace{-0.15cm}
\paragraph{GAN Inversion.}
We begin by pairing existing inversion methods with our transformed generator.  Given a real image, we first invert it using a ReStyle encoder~\cite{alaluf2021restyle}, pre-trained on the human face domain. We then insert the inverted latent-code $w\in$ \wplus into our transformed generators. 
\cref{fig:real_edits} shows results obtained in such a manner, using generators adapted across multiple domains. Our generators successfully preserve the identity tied to the latent code, even for codes obtained from the inversion of real images. %

\vspace{-0.12cm}
\paragraph{Latent traversal editing.}
The inversion results suggest that the latent space of the adapted generator is aligned with that of the source generator. This is not entirely surprising. First, due to our \textit{intertwined} generator architecture and the nature of the directional loss. Second, because prior and concurrent methods successfully employed the natural alignment of fine-tuned generators for downstream applications \cite{pinkney2020resolution,10.1145/3450626.3459771,wang2021crossdomain,wu2021stylealign}.
However, our fine-tuning approach is non-adversarial and thus differs from these prior methods. Consequently, verifying that latent-space alignment remains unbroken is of great interest.
We use existing editing techniques and show that latent-space directions do indeed maintain their semantic meaning. As a result, rather than finding new paths in the latent space of the modified generator, we can simply reuse the same paths and editing models found for the original, source generator.

\begin{figure}[!hbt]
    \centering
    \setlength{\belowcaptionskip}{-2.5pt}
    \setlength{\tabcolsep}{1pt}
    {\tiny
    \begin{tabular}{c c c c c c}
    
        \raisebox{0.038\textwidth}{\rotatebox[origin=t]{90}{\scalebox{0.9}{ $G_{frozen}\left(w\right)$ }}} & 
        \includegraphics[width=0.18\linewidth]{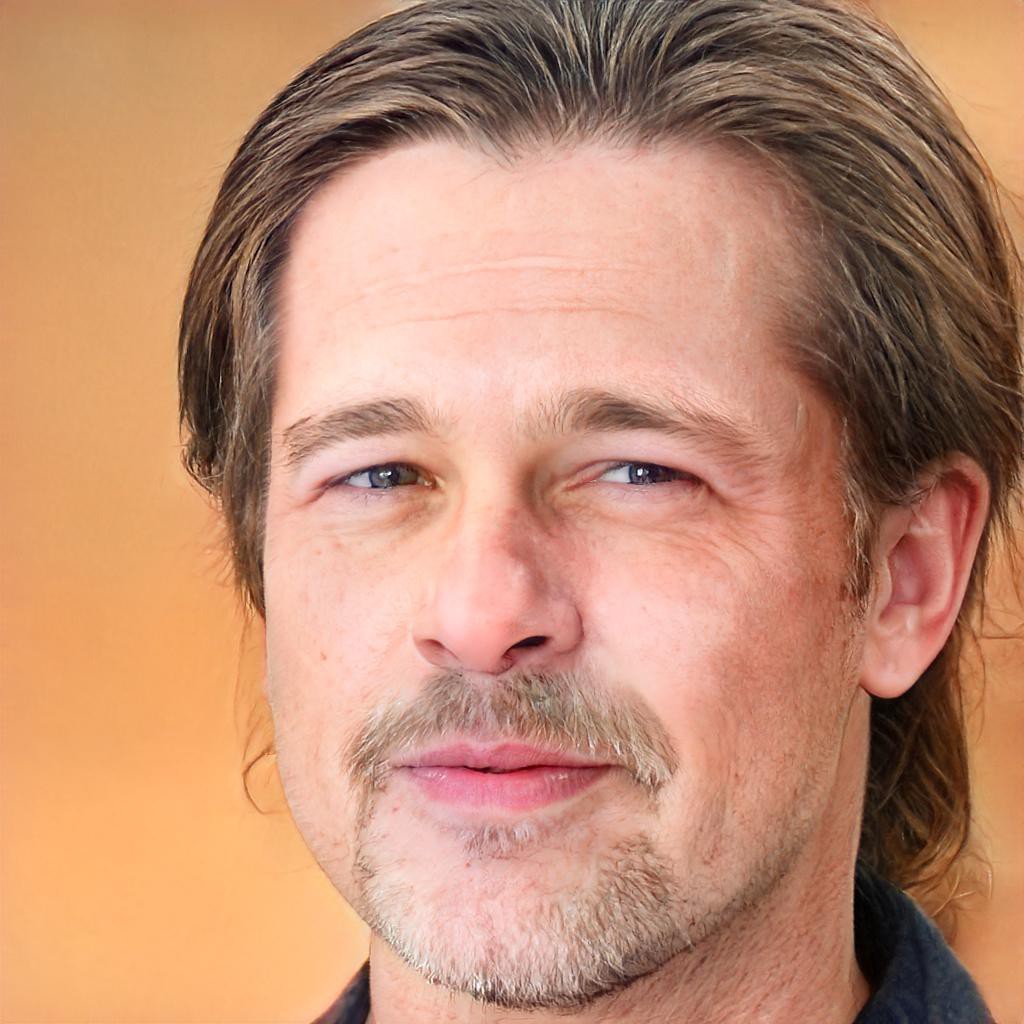} & \includegraphics[width=0.18\linewidth]{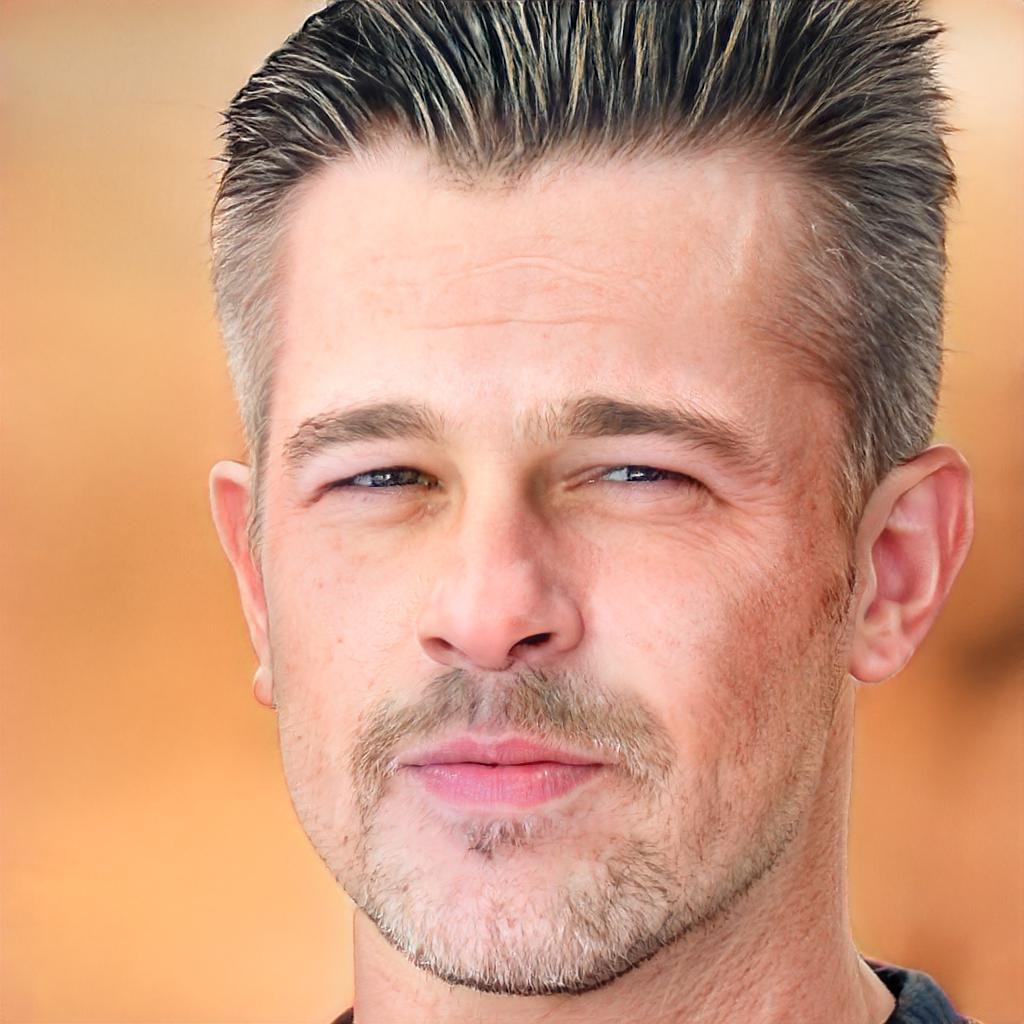} &
        \includegraphics[width=0.18\linewidth]{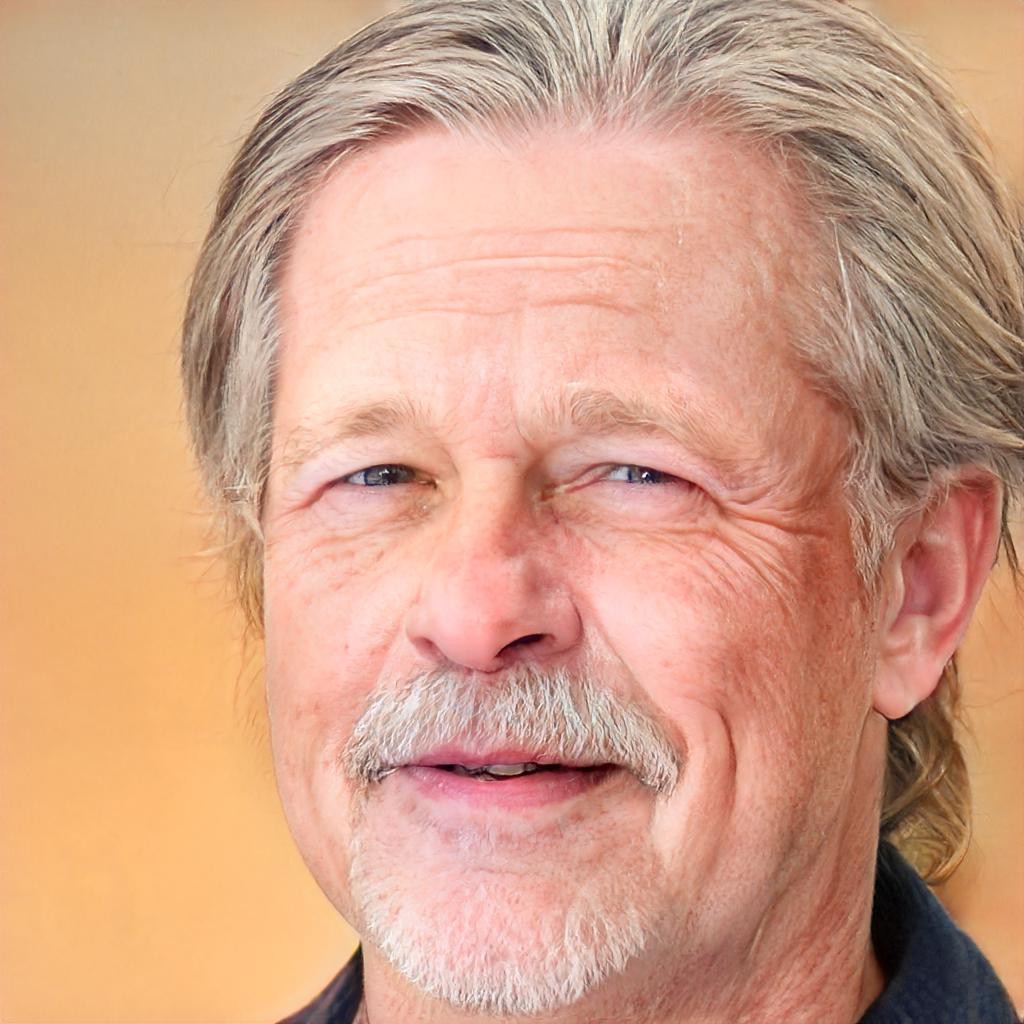} &
        \includegraphics[width=0.18\linewidth]{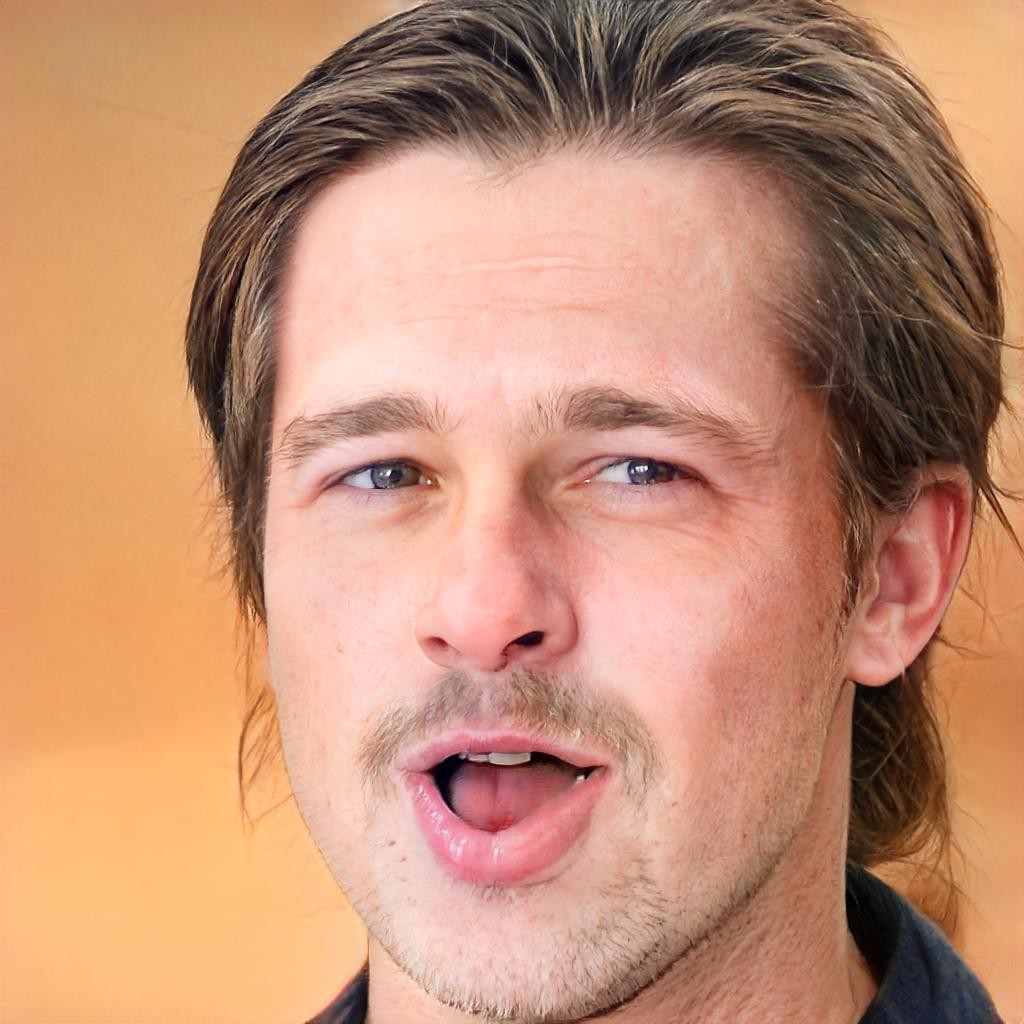} &
        \includegraphics[width=0.18\linewidth]{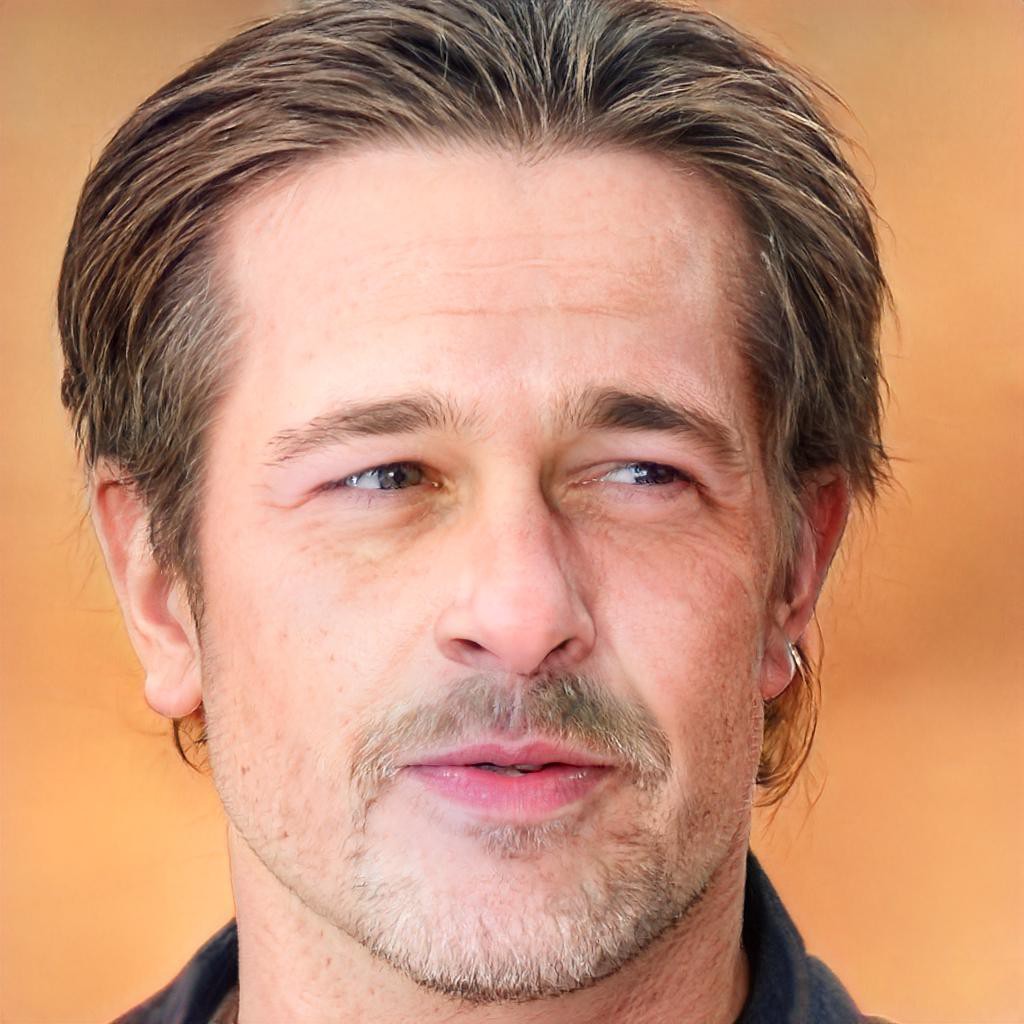} %
        \\
        
        \raisebox{0.022\textwidth}{\rotatebox[origin=t]{90}{\scalebox{0.9}{\begin{tabular}{c@{}c@{}c@{}} Photo $\rightarrow$  \\ Fernando Botero \\ Painting\end{tabular}}}} & 
        \includegraphics[width=0.18\linewidth]{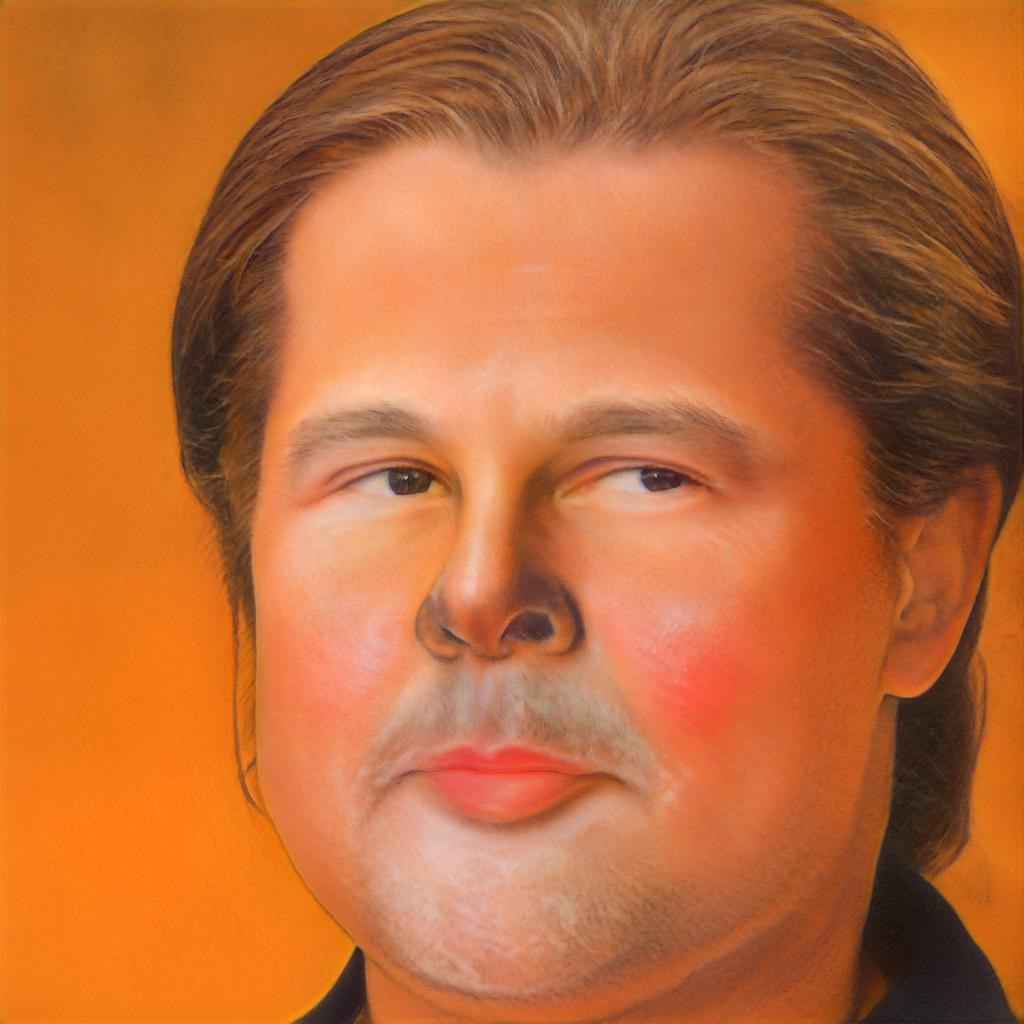} &
        \includegraphics[width=0.18\linewidth]{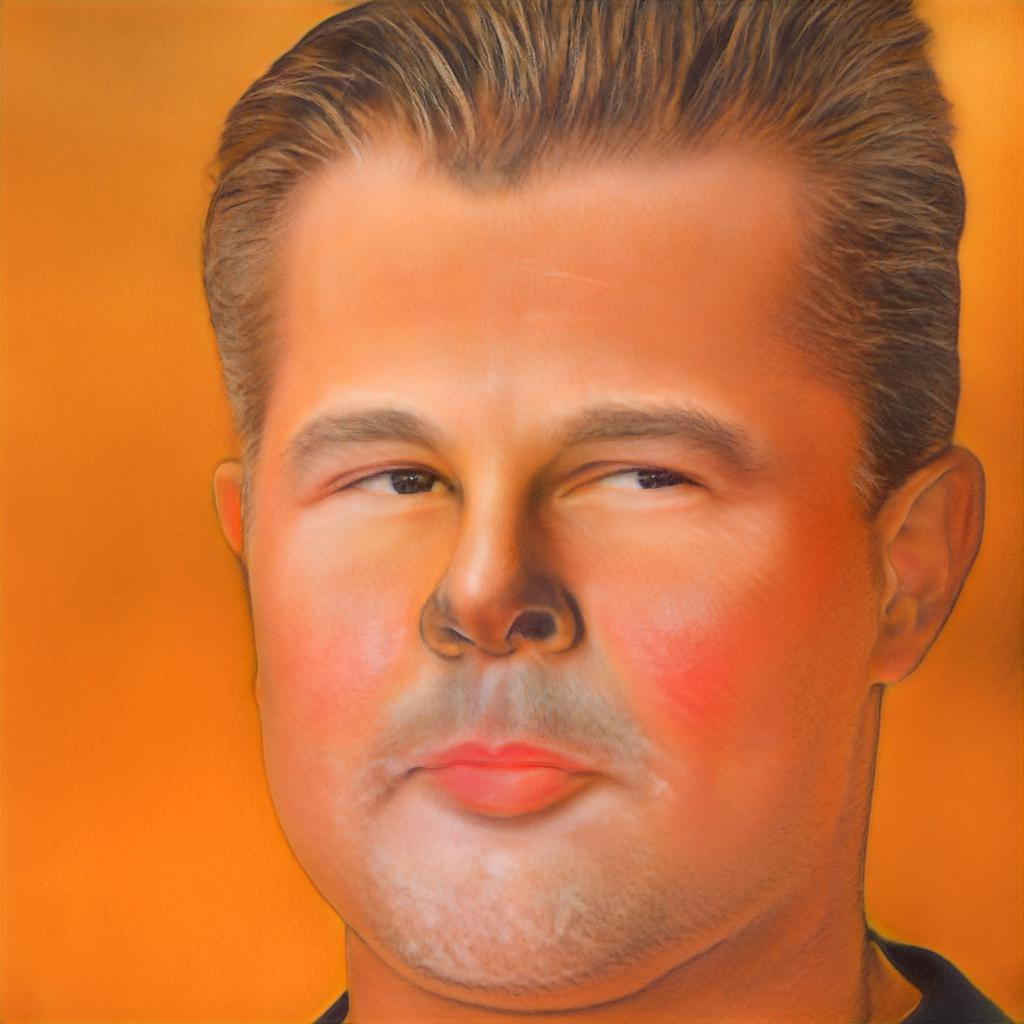} &
        \includegraphics[width=0.18\linewidth]{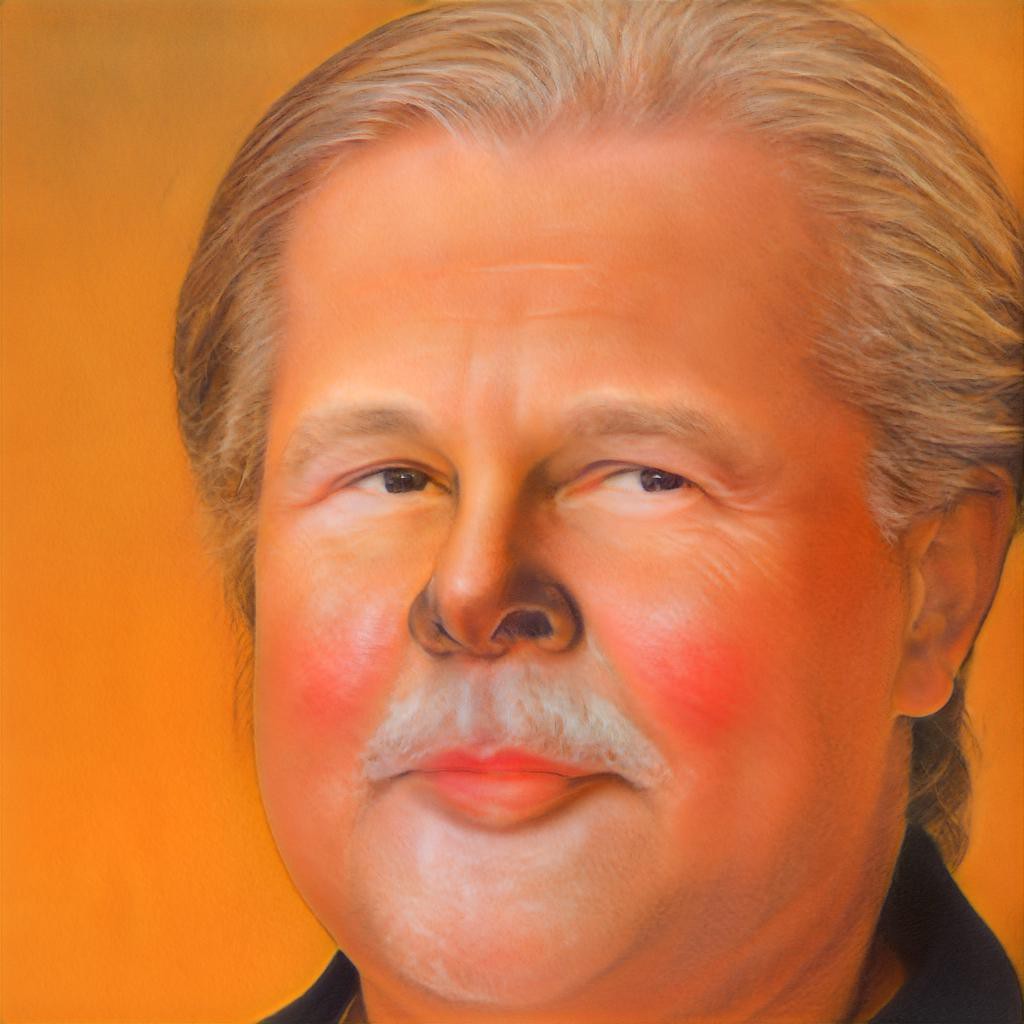} &
        \includegraphics[width=0.18\linewidth]{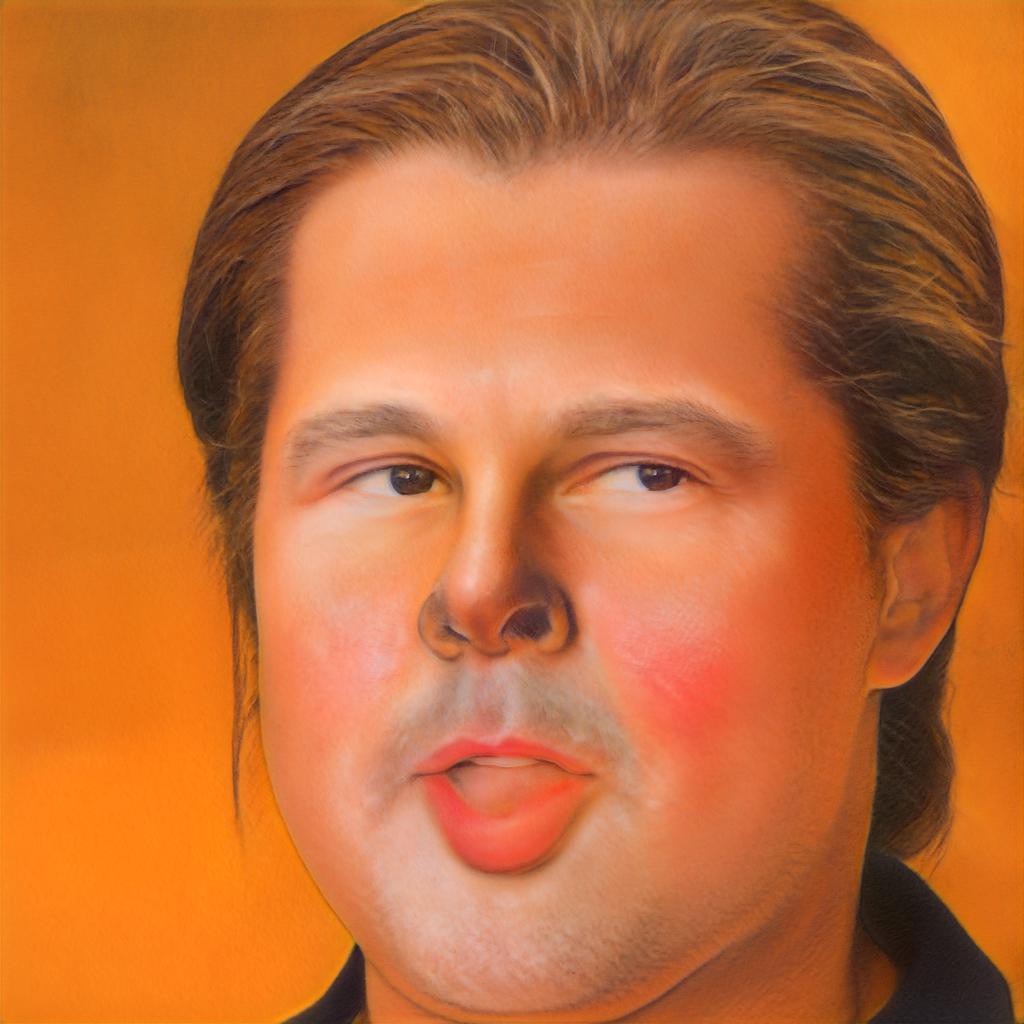} &
        \includegraphics[width=0.18\linewidth]{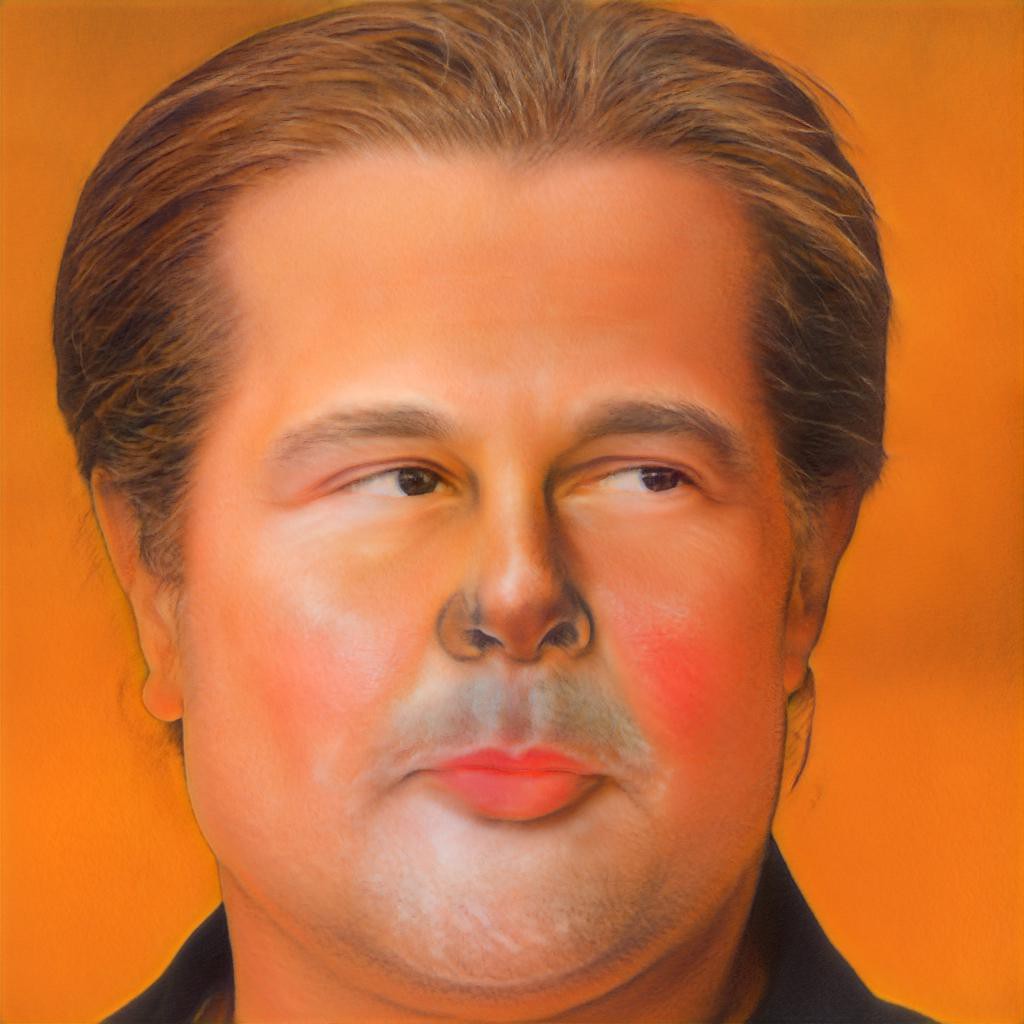}
        \\
        
        \raisebox{0.025\textwidth}{\rotatebox[origin=t]{90}{\scalebox{0.9}{\begin{tabular}{c@{}c@{}c@{}} Photo $\rightarrow$   \\ Rendered 3D in  \\ the Style of Pixar \end{tabular}}}} &
        \includegraphics[width=0.18\linewidth]{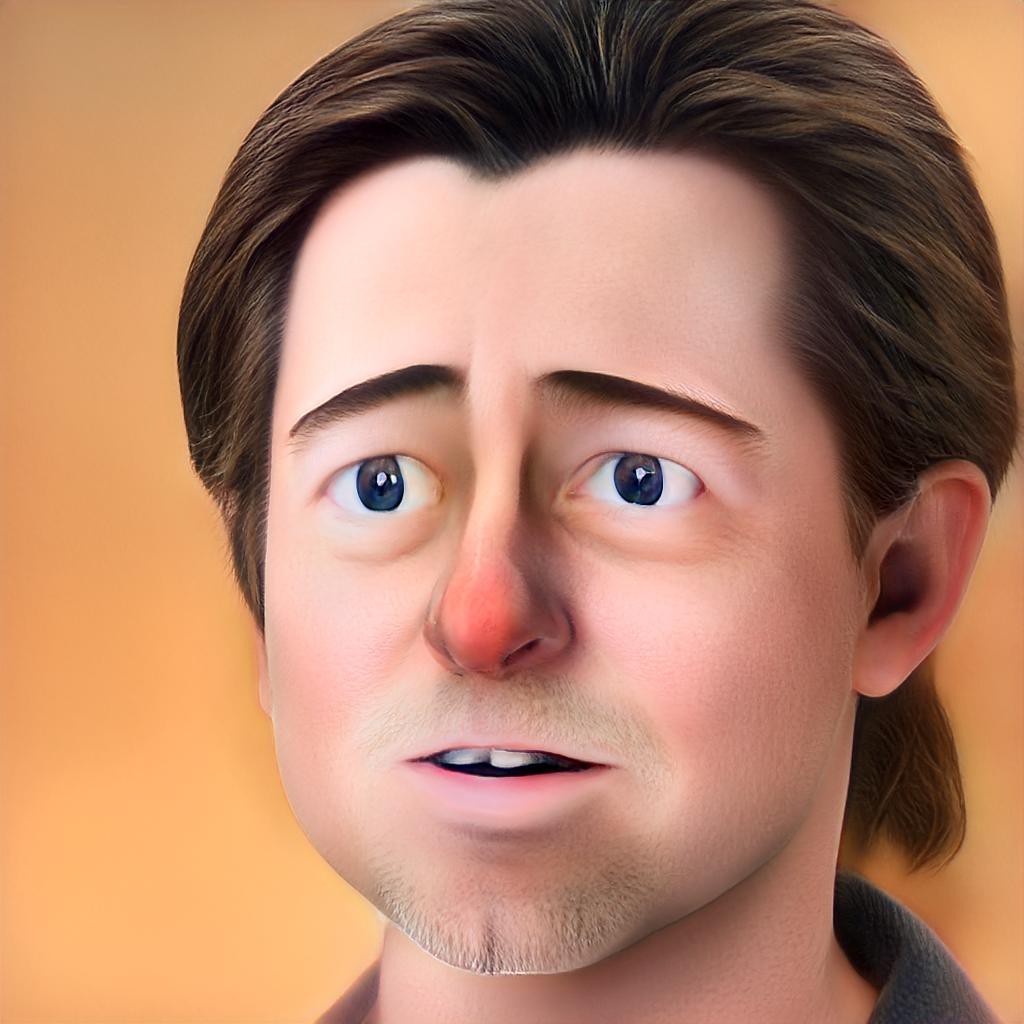} &
        \includegraphics[width=0.18\linewidth]{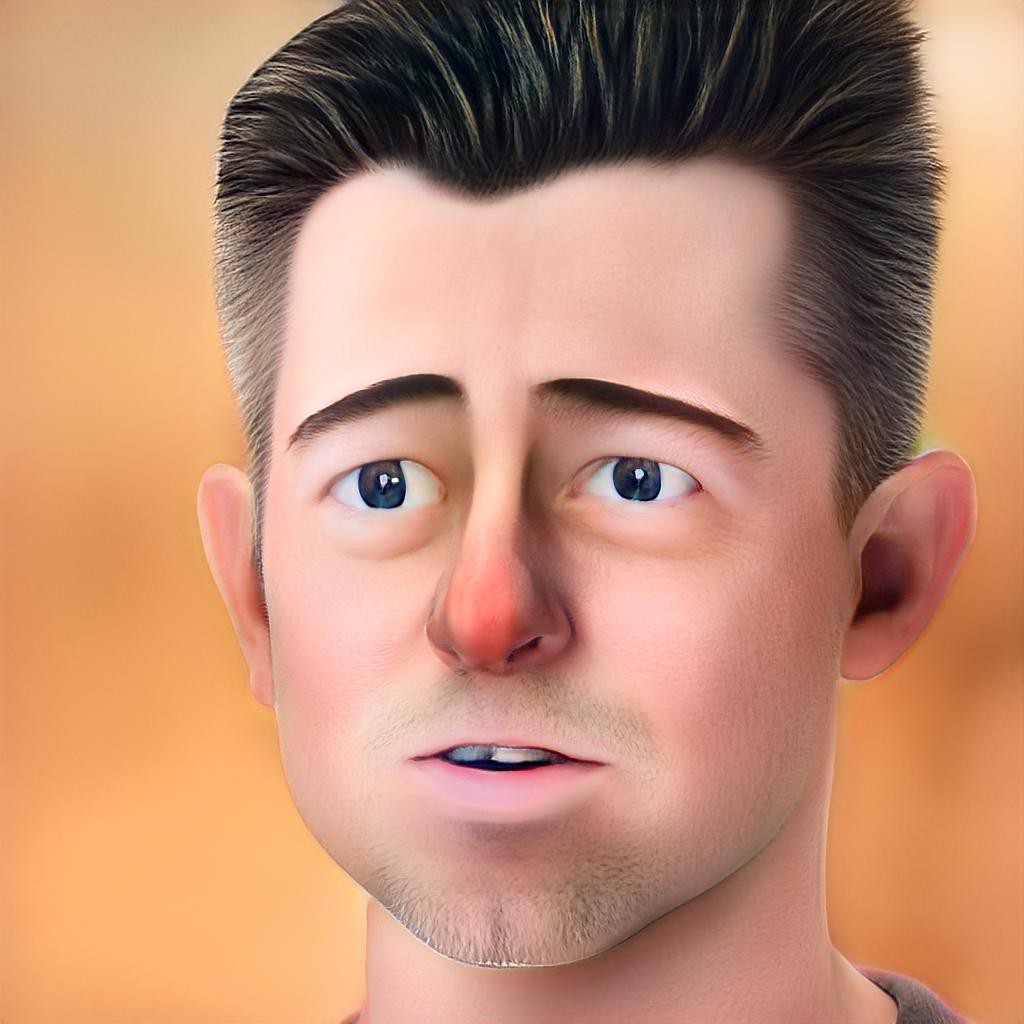} &
        \includegraphics[width=0.18\linewidth]{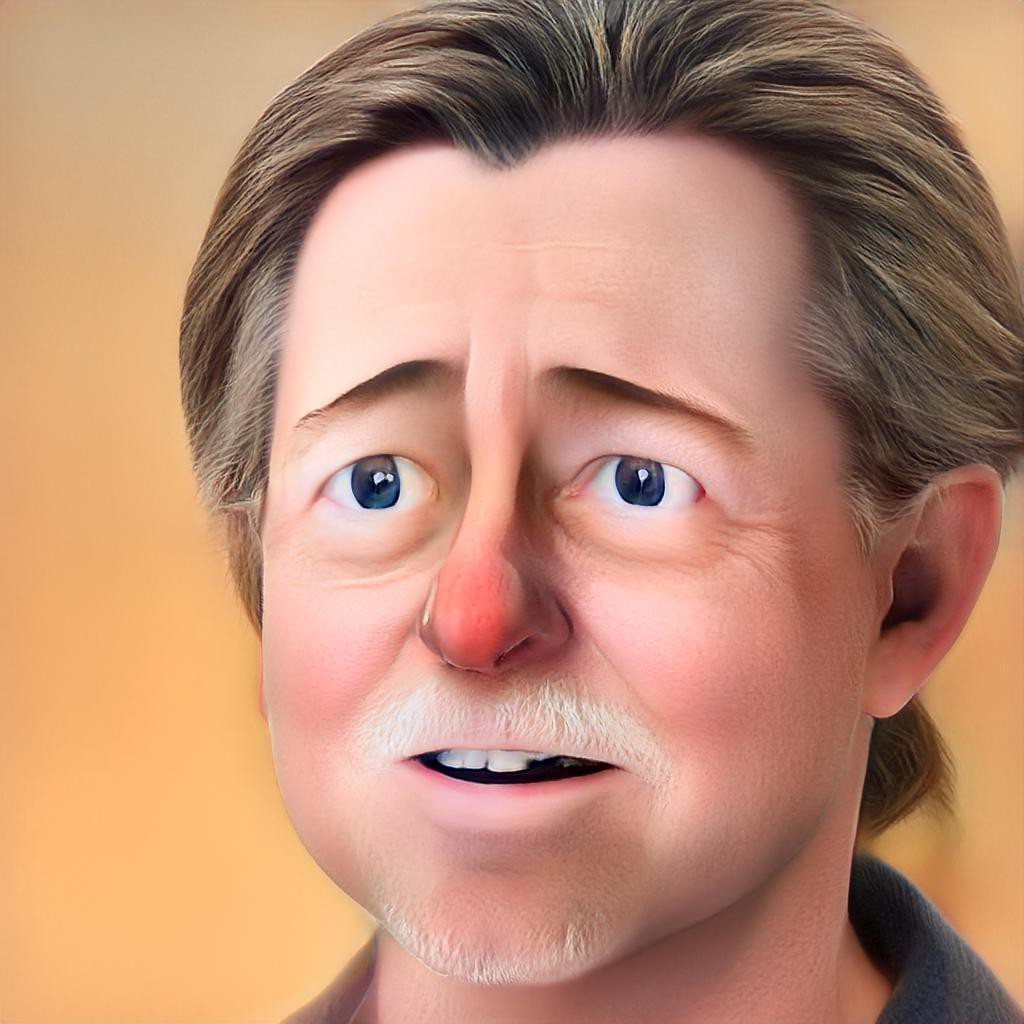} &
        \includegraphics[width=0.18\linewidth]{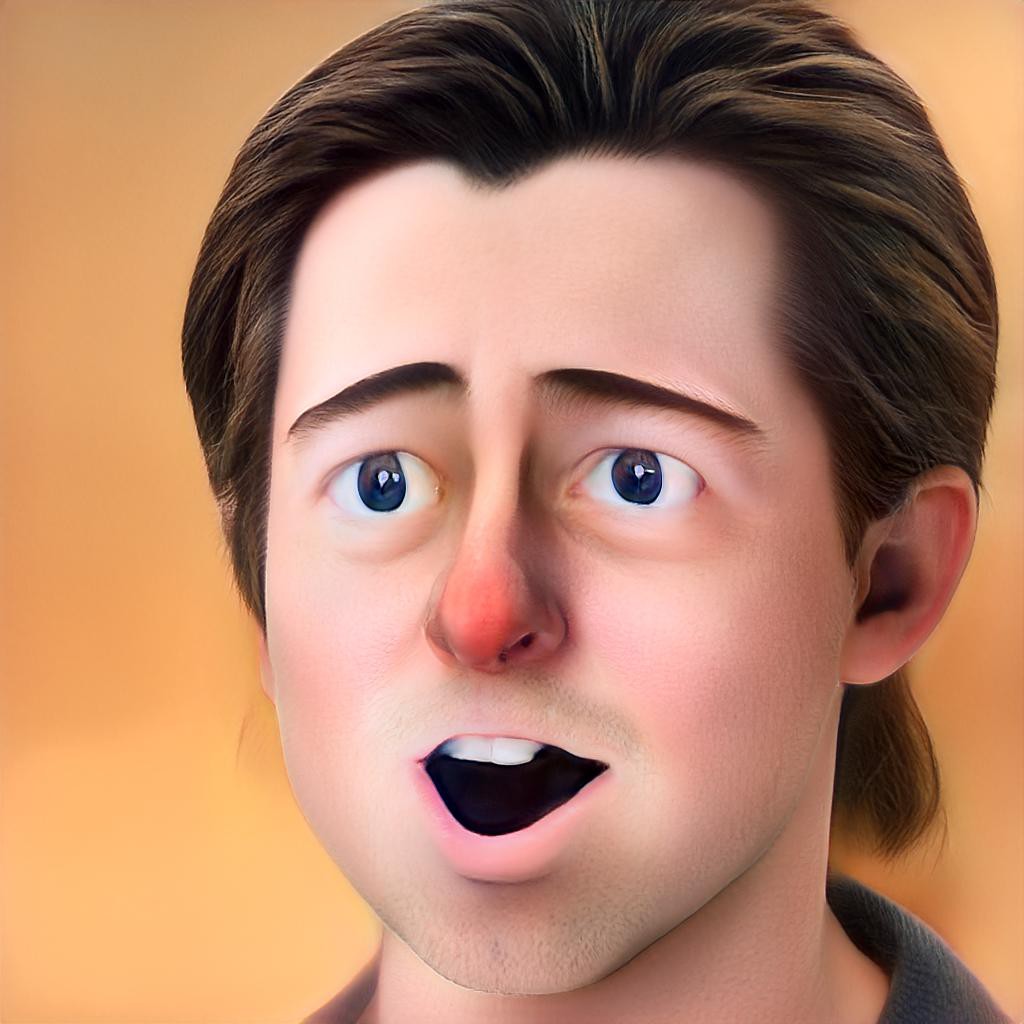} &
        \includegraphics[width=0.18\linewidth]{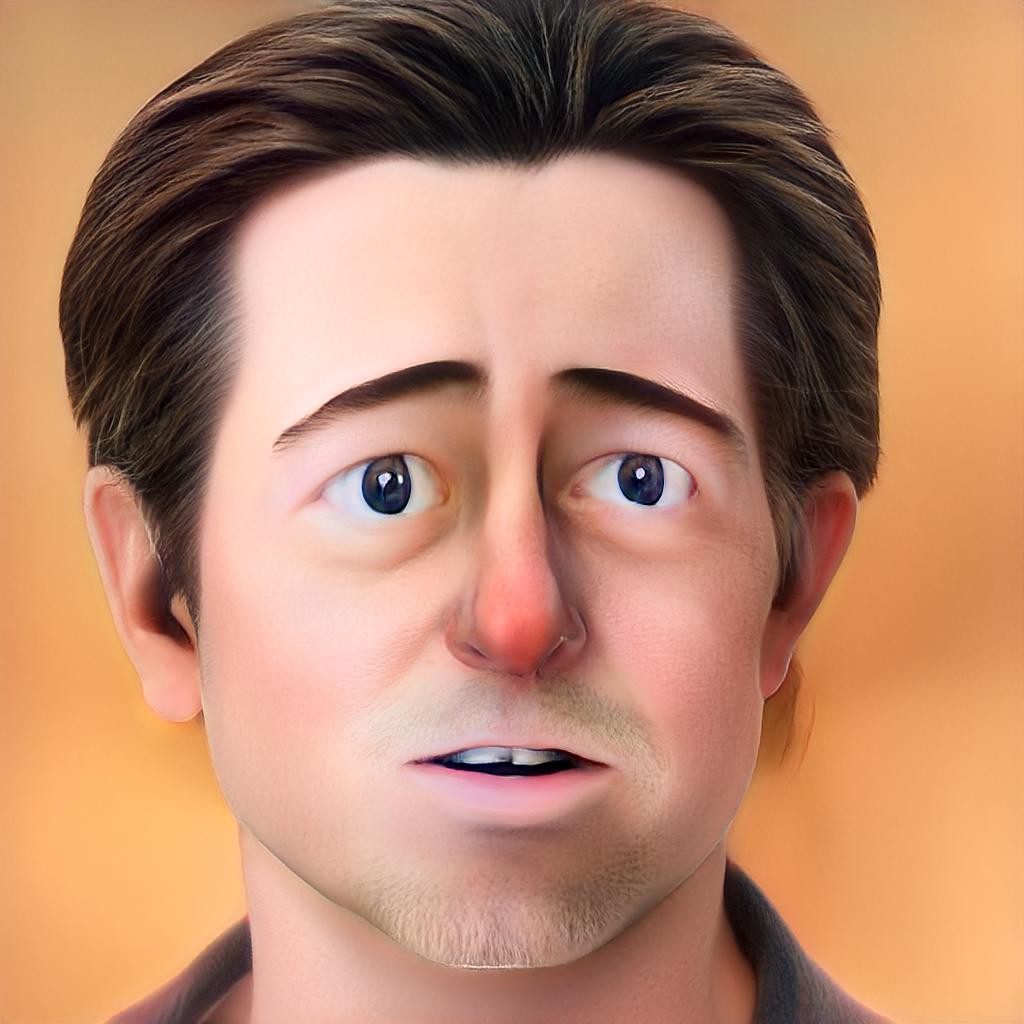} 
        \\

        \vspace{-7pt}
        & \scriptsize Unmodified  & \scriptsize Mohawk & \scriptsize Age & \scriptsize Surprised & \scriptsize Pose

    \end{tabular}}
    \vspace{0.1cm}
    \caption{ Multi-domain editing of a real, inverted image using StyleCLIP \cite{patashnik2021styleclip} (mohawk and surprised), InterFaceGAN \cite{shen2020interpreting} (age) and StyleFlow \cite{10.1145/3447648} (pose). The top row portrays editing in the source domain. All rows below show the same editing operations in our translated domains.  }
    \label{fig:editing_styleflow} \vspace{-5pt}
\end{figure}

In Figure~\ref{fig:editing_styleflow}, we edit a real image mapped into novel domains, using several off-the-shelf methods. We use StyleCLIP \cite{patashnik2021styleclip} to edit expression and hairstyle, StyleFlow \cite{10.1145/3447648} to edit pose, and InterFaceGAN \cite{shen2020interpreting} to edit age. We use the original implementations, pre-trained on the \textit{source} domain. %

\begin{figure}[!hbt]
    \centering
    \setlength{\belowcaptionskip}{-2.5pt}
    \setlength{\tabcolsep}{0.5pt}
    \renewcommand{\arraystretch}{0.5}
    {\scriptsize
    \vspace{-0pt}
    \begin{tabular}{c c c c c}
        
        Source &
        {\begin{tabular}{c@{}c@{}} Human $\rightarrow$ \\  White Walker\end{tabular}} & {\begin{tabular}{c@{}c@{}} Photo $\rightarrow$ \\ Sketch \end{tabular}} & {\begin{tabular}{c@{}c@{}} Human $\rightarrow$ \\ Disney Princess \end{tabular}} & {\begin{tabular}{c@{}c@{}c@{}c@{}} Photo $\rightarrow$ \\ Painting in \\ the Style of \\ Edvard Munch \end{tabular}} \\

        \includegraphics[width=0.195\linewidth]{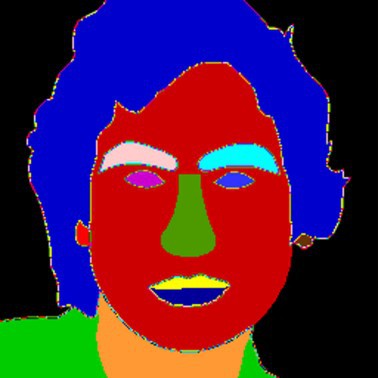} &
        \includegraphics[width=0.195\linewidth]{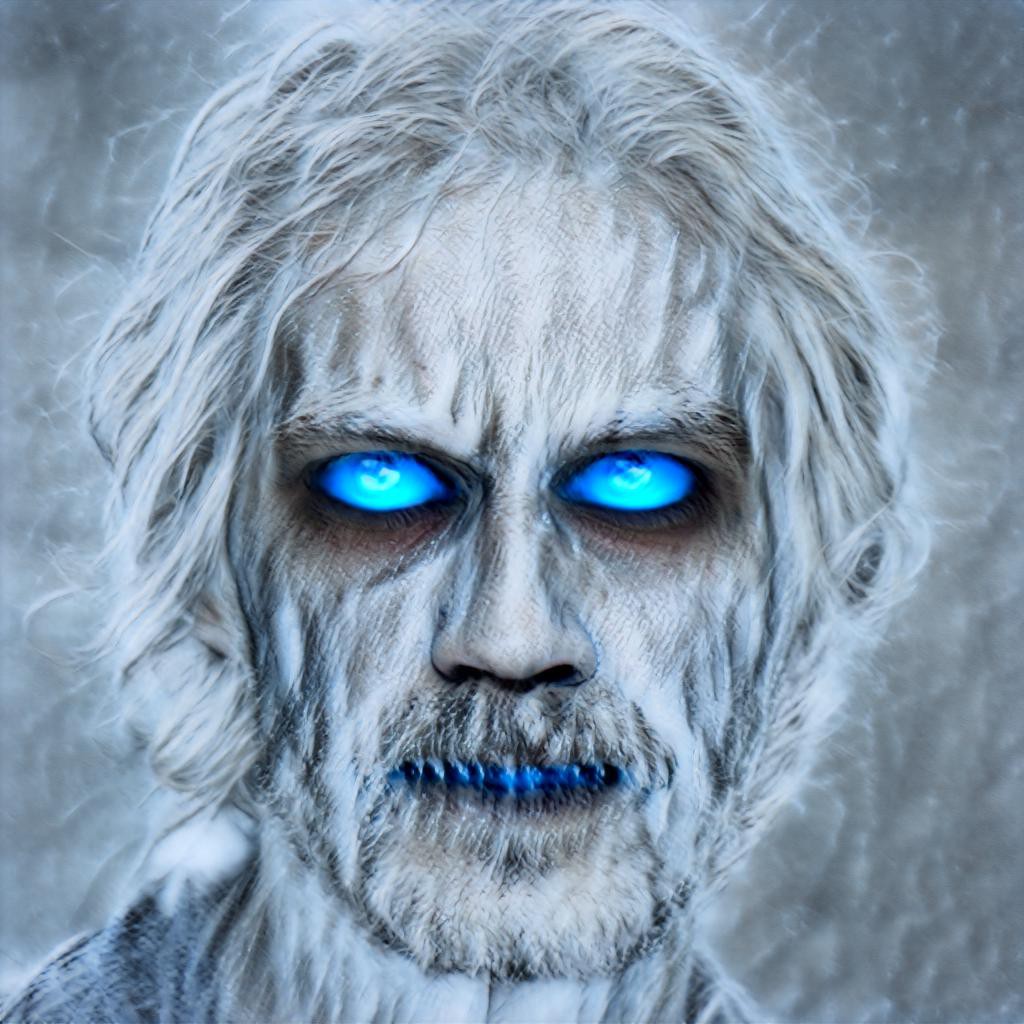} &
        \includegraphics[width=0.195\linewidth]{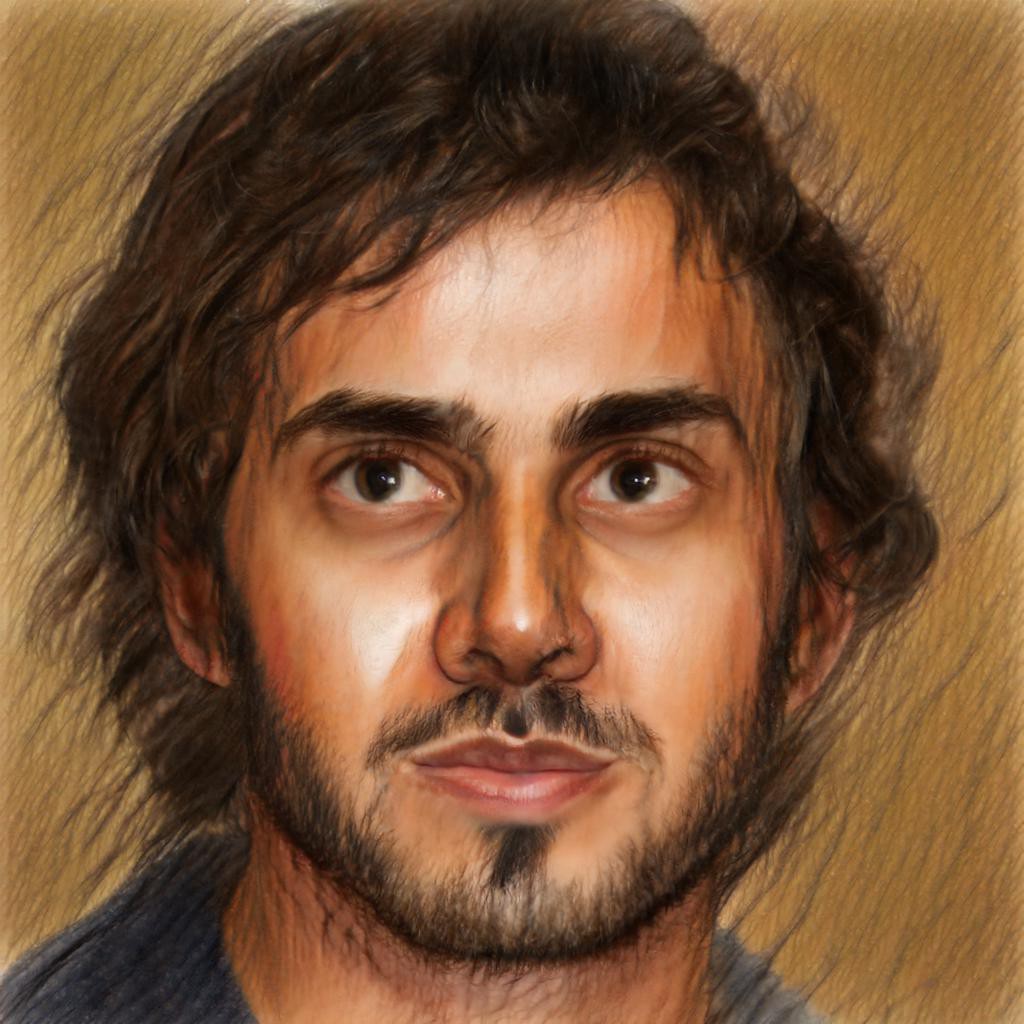} &
        \includegraphics[width=0.195\linewidth]{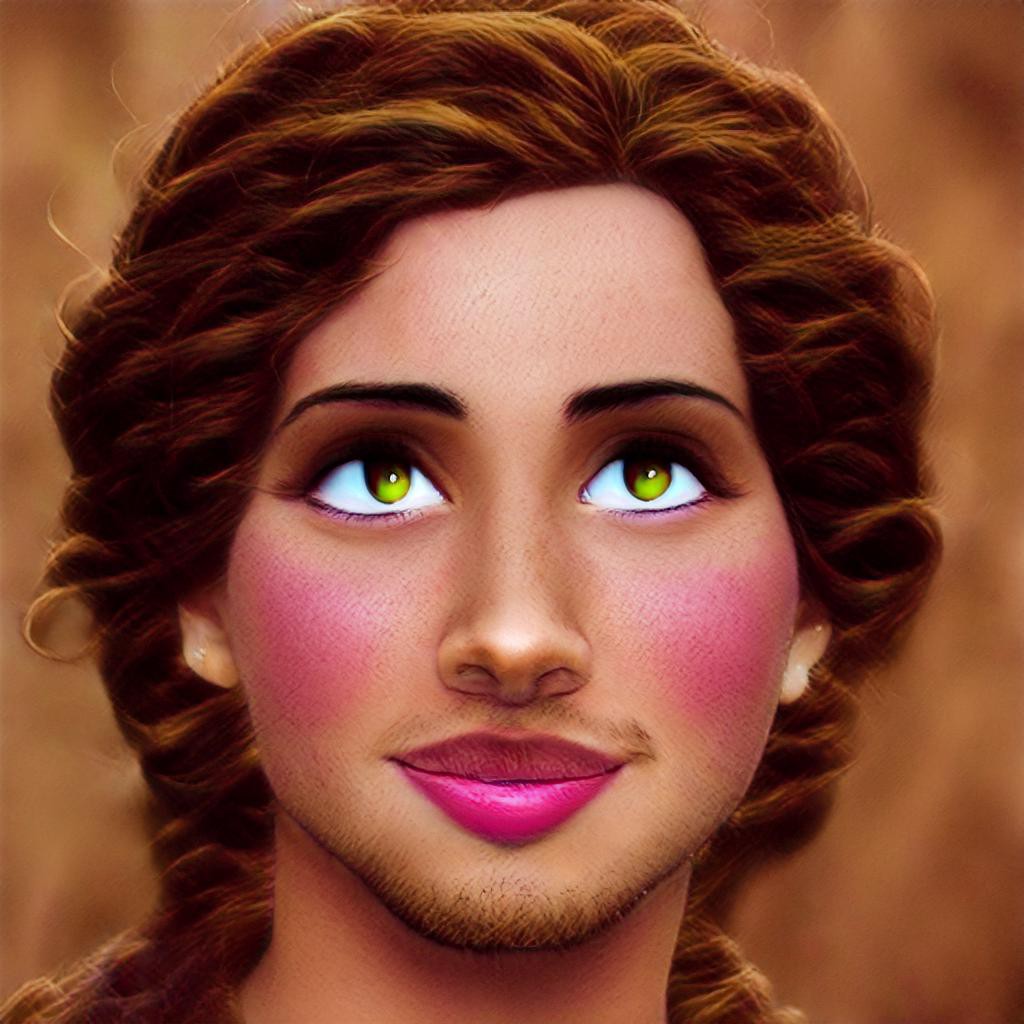} &
        \includegraphics[width=0.195\linewidth]{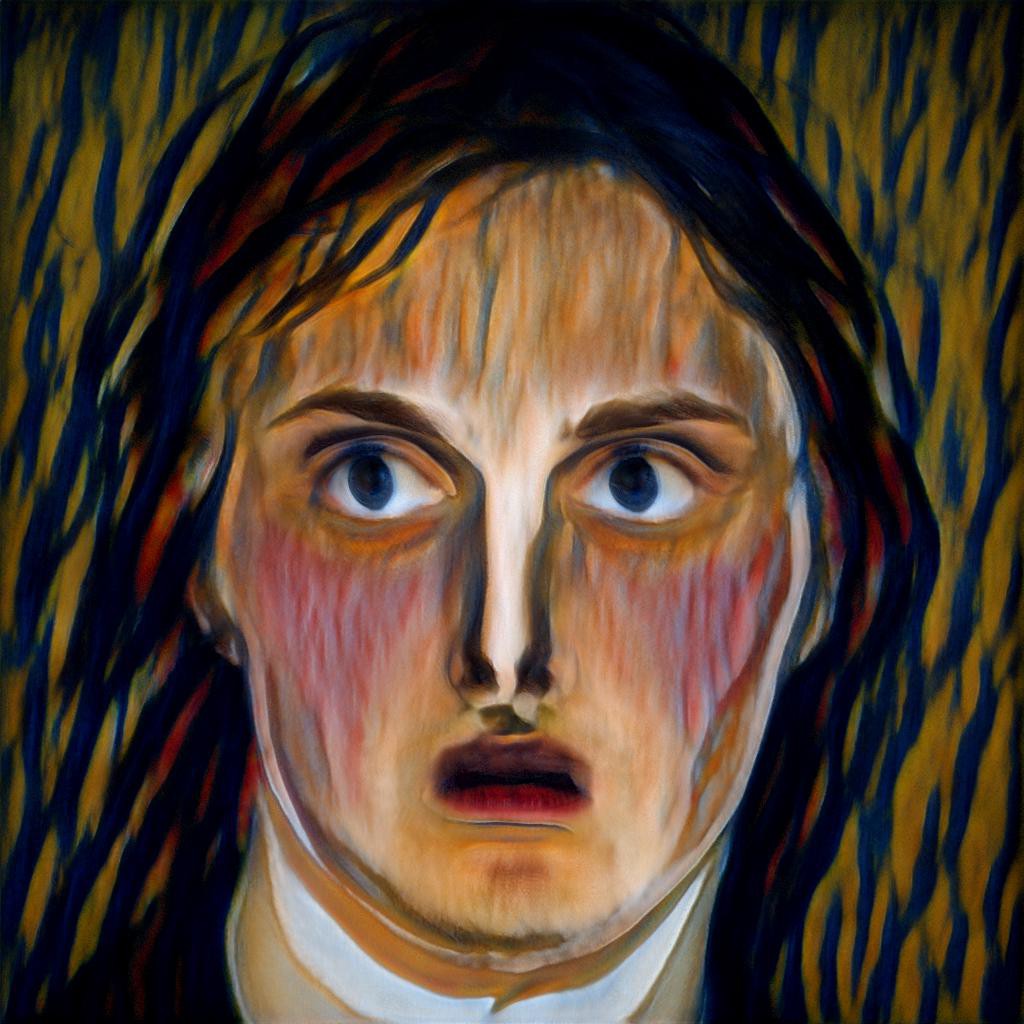} \\
        
        \includegraphics[width=0.195\linewidth]{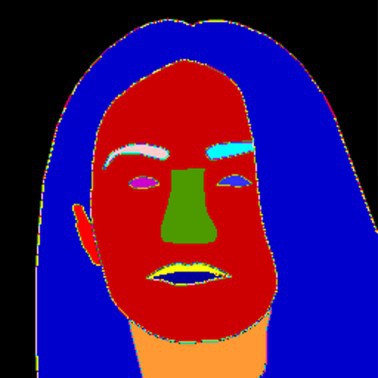} &
        \includegraphics[width=0.195\linewidth]{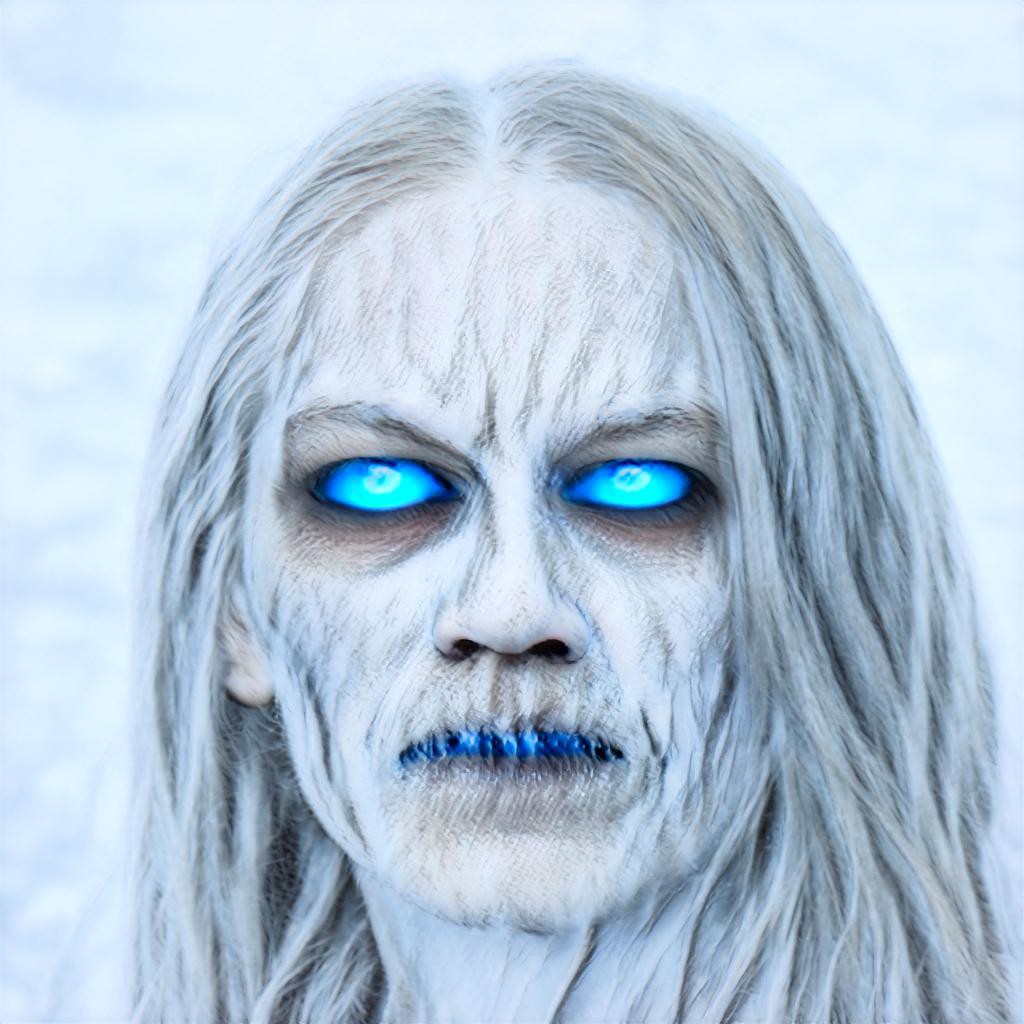} &
        \includegraphics[width=0.195\linewidth]{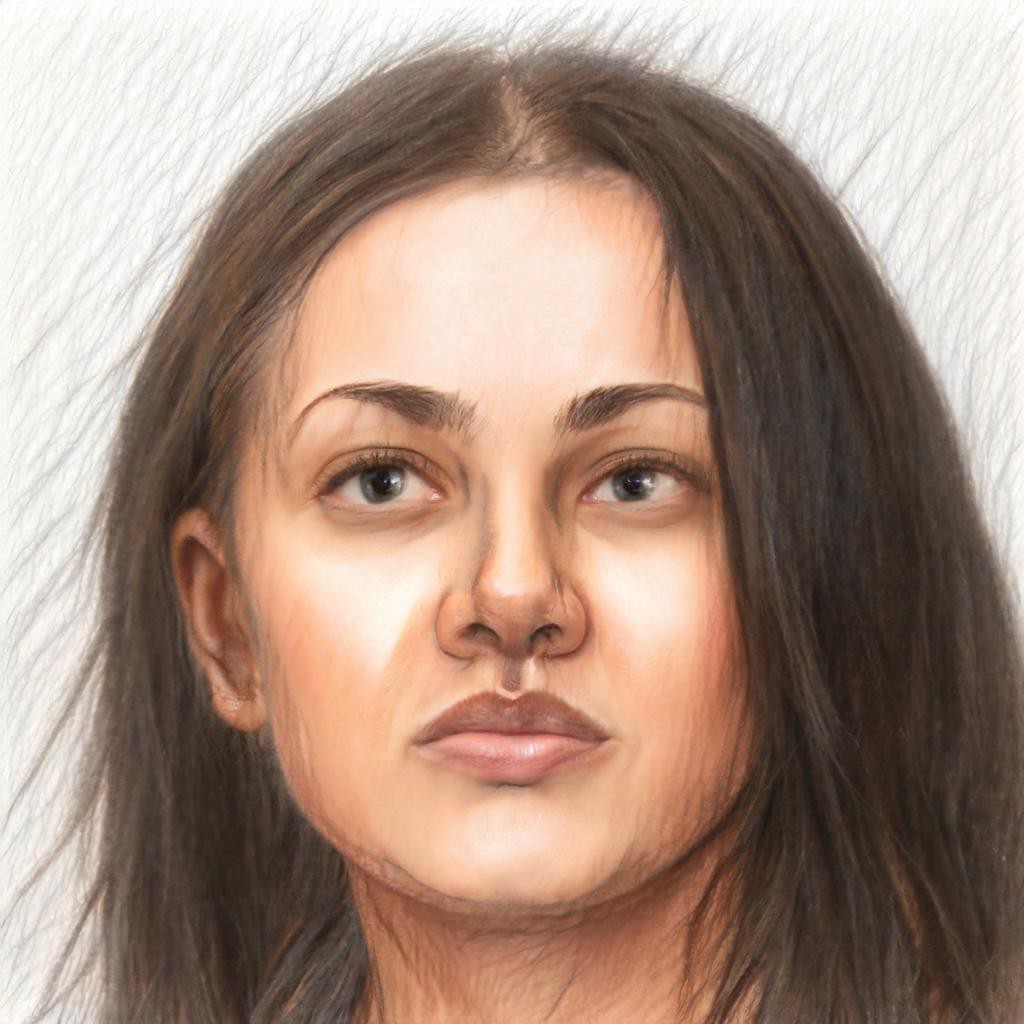} &
        \includegraphics[width=0.195\linewidth]{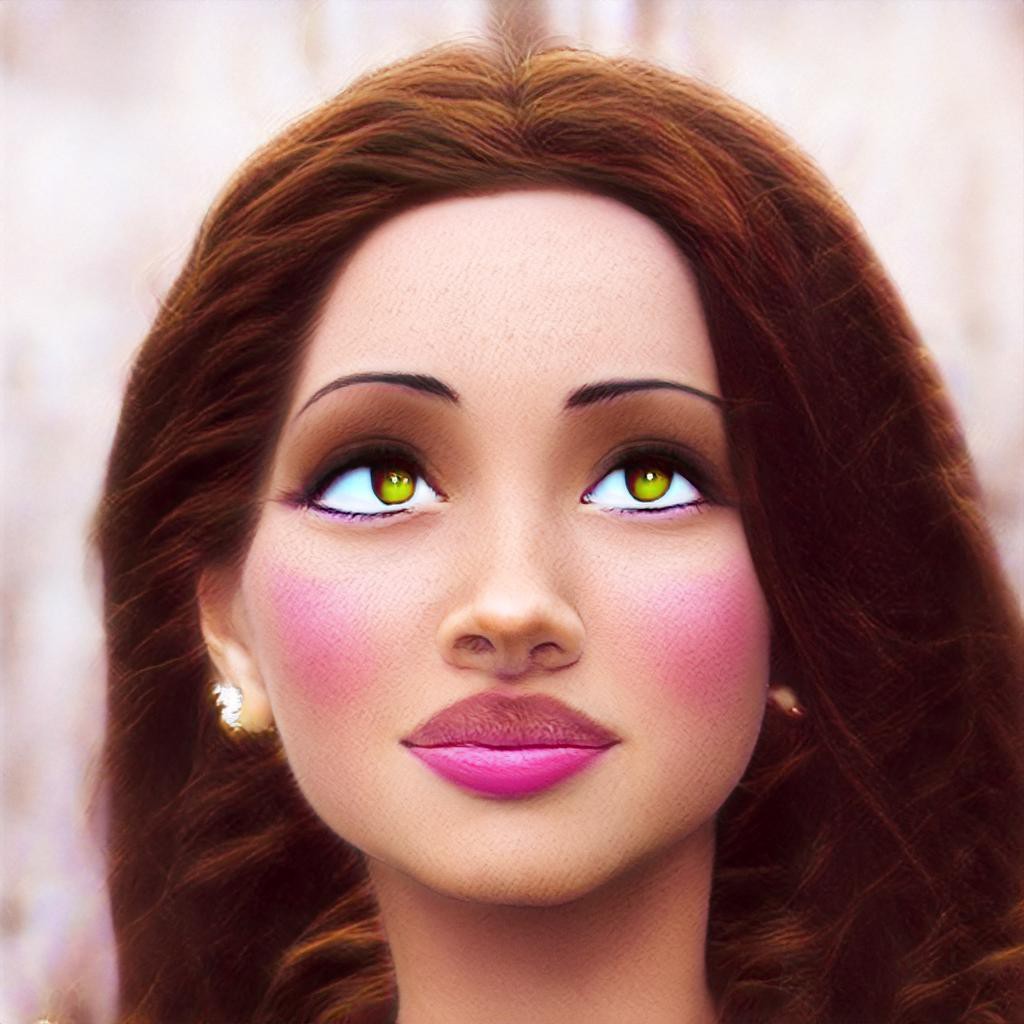} &
        \includegraphics[width=0.195\linewidth]{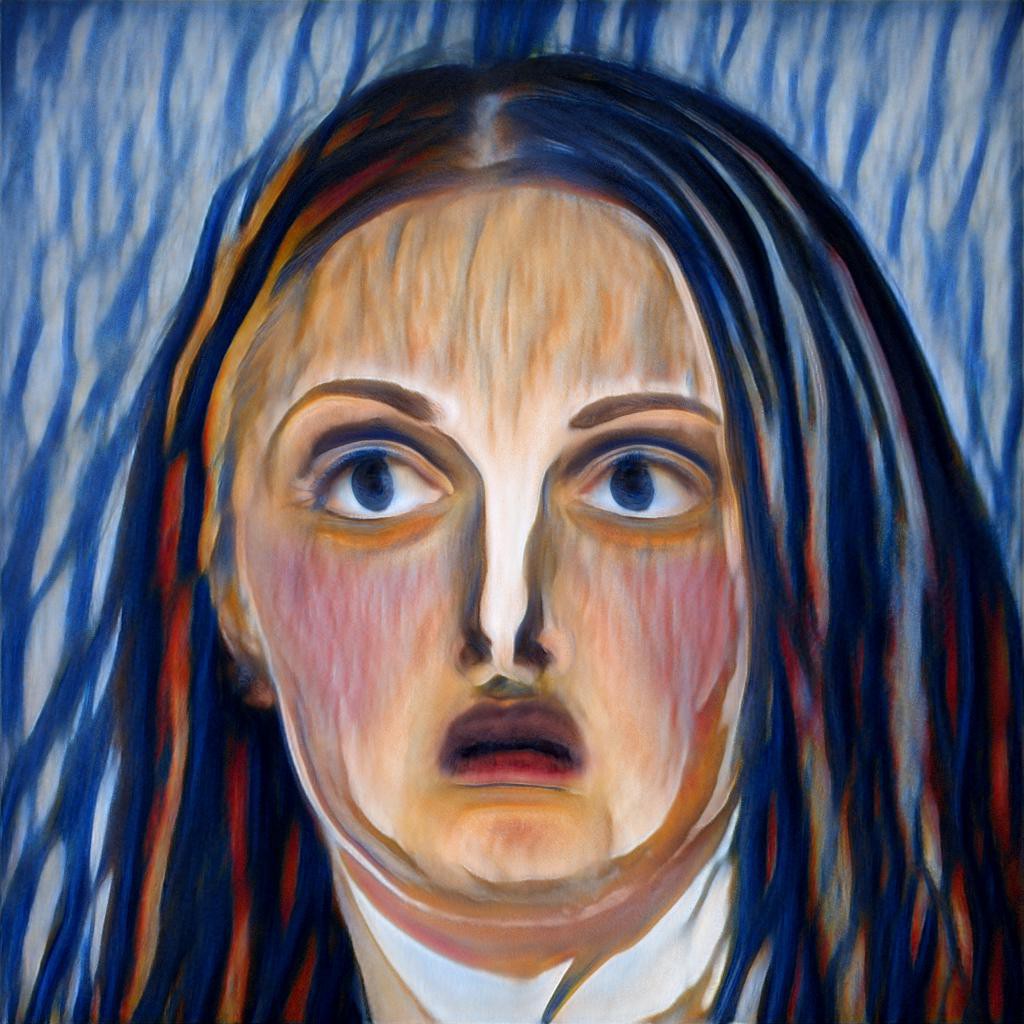} \\
        
        \includegraphics[width=0.195\linewidth]{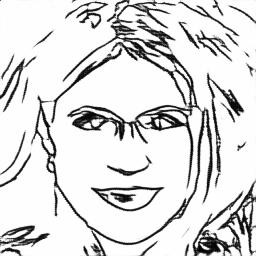} &
        \includegraphics[width=0.195\linewidth]{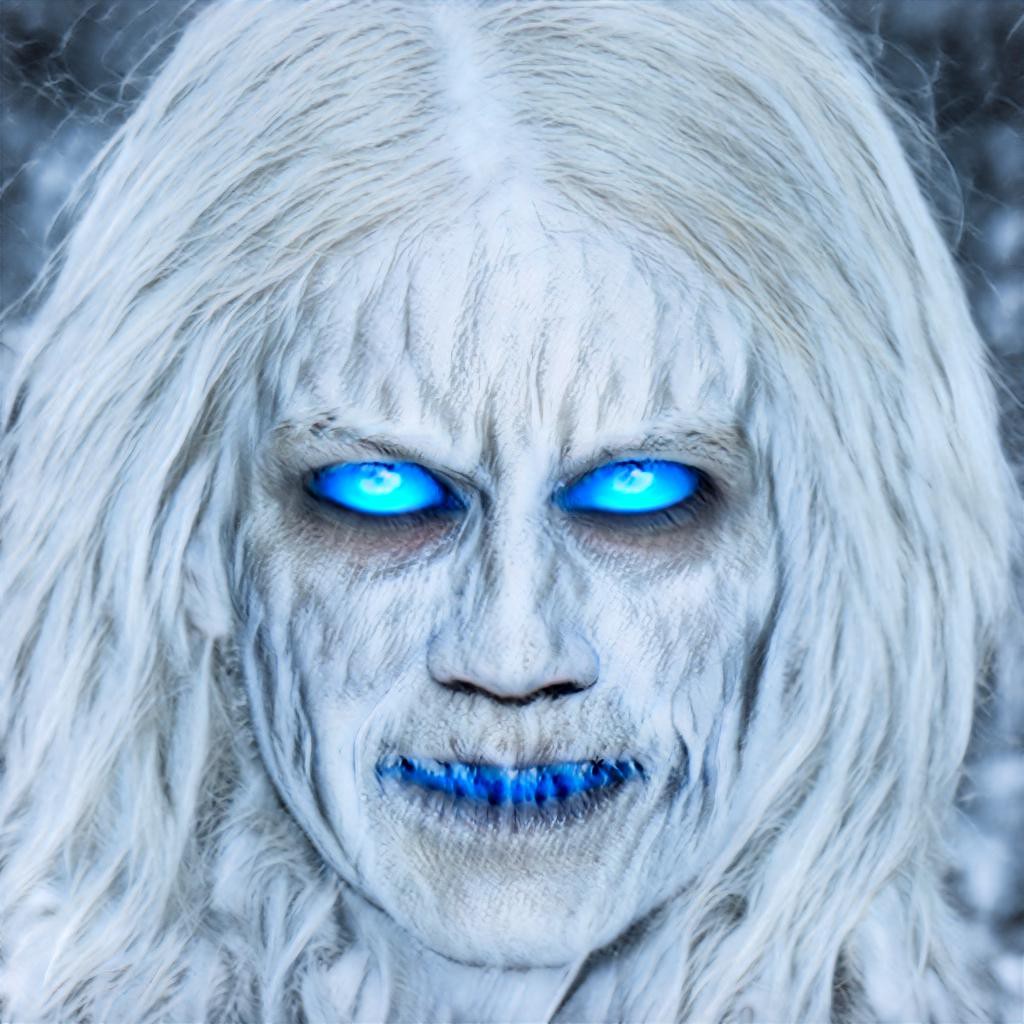} &
        \includegraphics[width=0.195\linewidth]{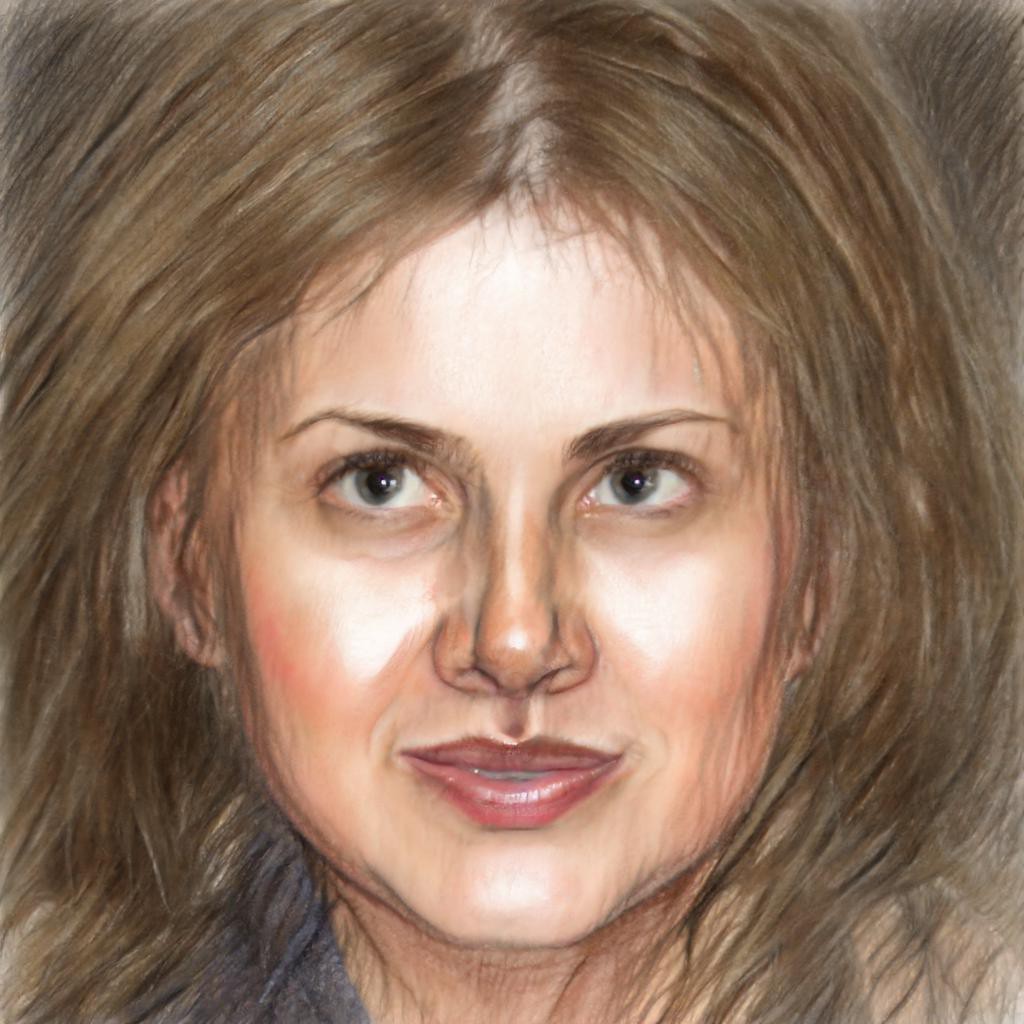} &
        \includegraphics[width=0.195\linewidth]{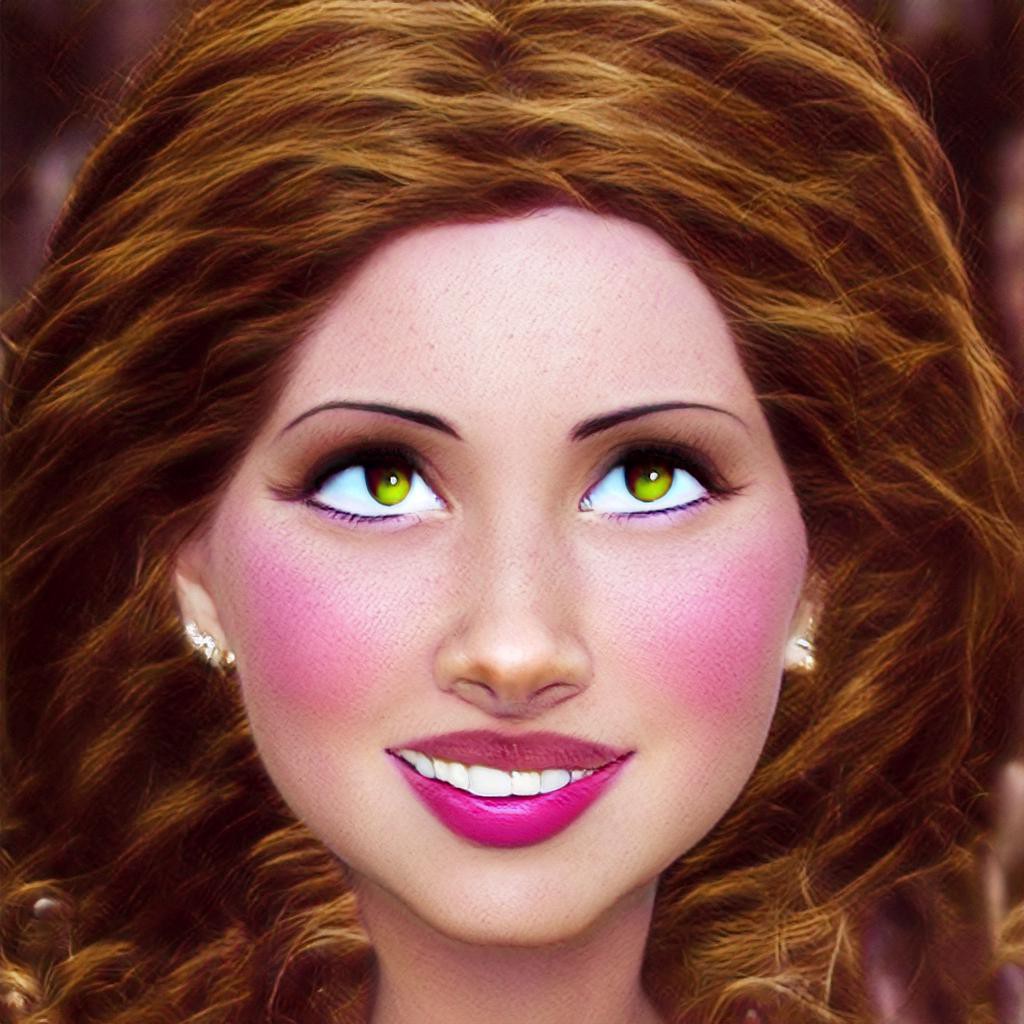} &
        \includegraphics[width=0.195\linewidth]{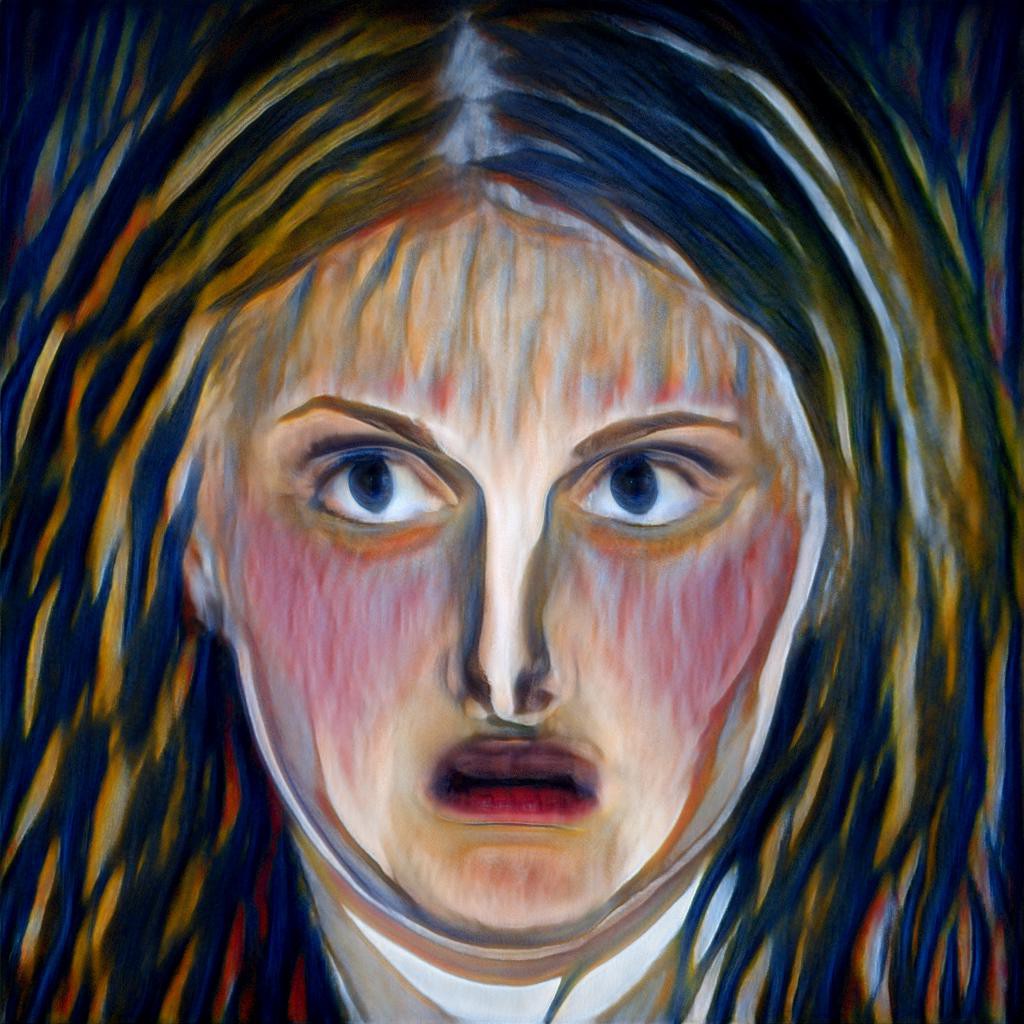} \\

    \end{tabular}}
    \vspace{-0.1cm}
    \caption{Conditional synthesis in multiple domains. We use the official pSp \cite{richardson2020encoding} FFHQ encoder to invert segmentation masks and simple  sketches (left) into the latent space of the GAN. The inverted-codes work seamlessly with our adapted models.}
    \label{fig:conditional_gen} \vspace{-25pt}
\end{figure}

\vspace{-5pt}
\paragraph{Image-to-image translation.}
Generative objectives extend beyond image-editing. 
Richardson \etal~\cite{richardson2020encoding} demonstrate a wide range of image-to-image translation applications approached by training encoders. These encoders learn a mapping from images in arbitrary domains to the latent space of a pre-trained generator. The generator is then used to re-synthesize an image in its own domain. They demonstrate this approach for conditional synthesis tasks, image restoration and super resolution. However, a significant limitation of their method is that the target domain of the generated images is restricted to domains for which a StyleGAN generator can be trained.
We show that these pre-trained encoders can also be paired with our adapted generators, enabling more generic image-to-image translation.
Specifically, \cref{fig:conditional_gen} shows conditional image synthesis in multiple domains, using segmentation masks and sketch-based guidance, without re-training the encoder.

\subsection{Comparison to other methods}
We compare two aspects of our method to alternative approaches. First, we show that the text-driven out-of-domain capabilities of our method cannot be replicated by current latent-editing based techniques. Then, we demonstrate that StyleGAN-NADA can affect large shape changes better than current few-shot training approaches.

\begin{figure}[t]
    \centering
    \setlength{\belowcaptionskip}{-1pt}
    \setlength{\tabcolsep}{1pt}
    {\tiny
    \begin{tabular}{c c c c c c}

        \raisebox{0.025\textwidth}{\rotatebox[origin=t]{90}{\begin{tabular}{c@{}c@{}}Photo $\rightarrow$ \\ Raphael Painting\end{tabular}}} &
        \includegraphics[width=0.18\linewidth]{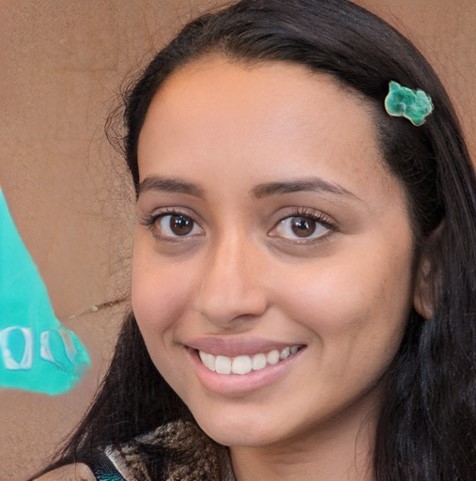} &
        \includegraphics[width=0.18\linewidth]{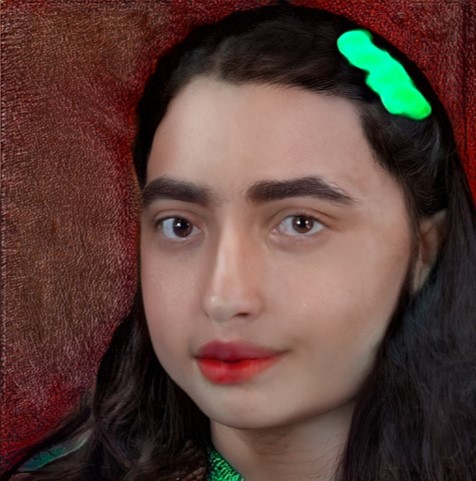} &
        \includegraphics[width=0.18\linewidth]{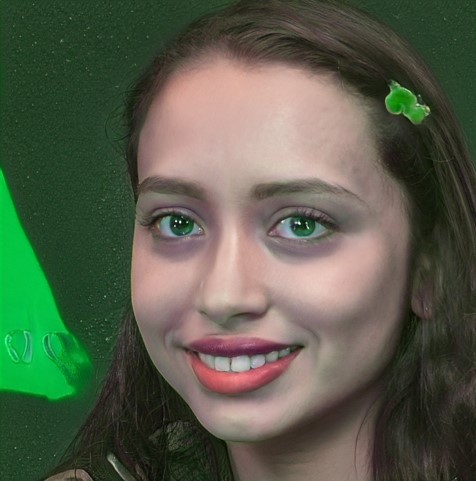} &
        \includegraphics[width=0.18\linewidth]{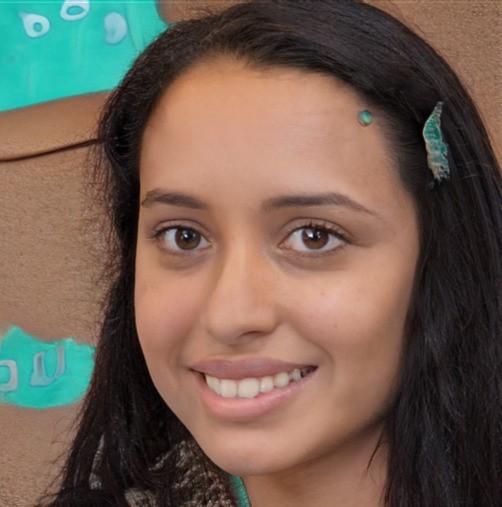} &
        \includegraphics[width=0.18\linewidth]{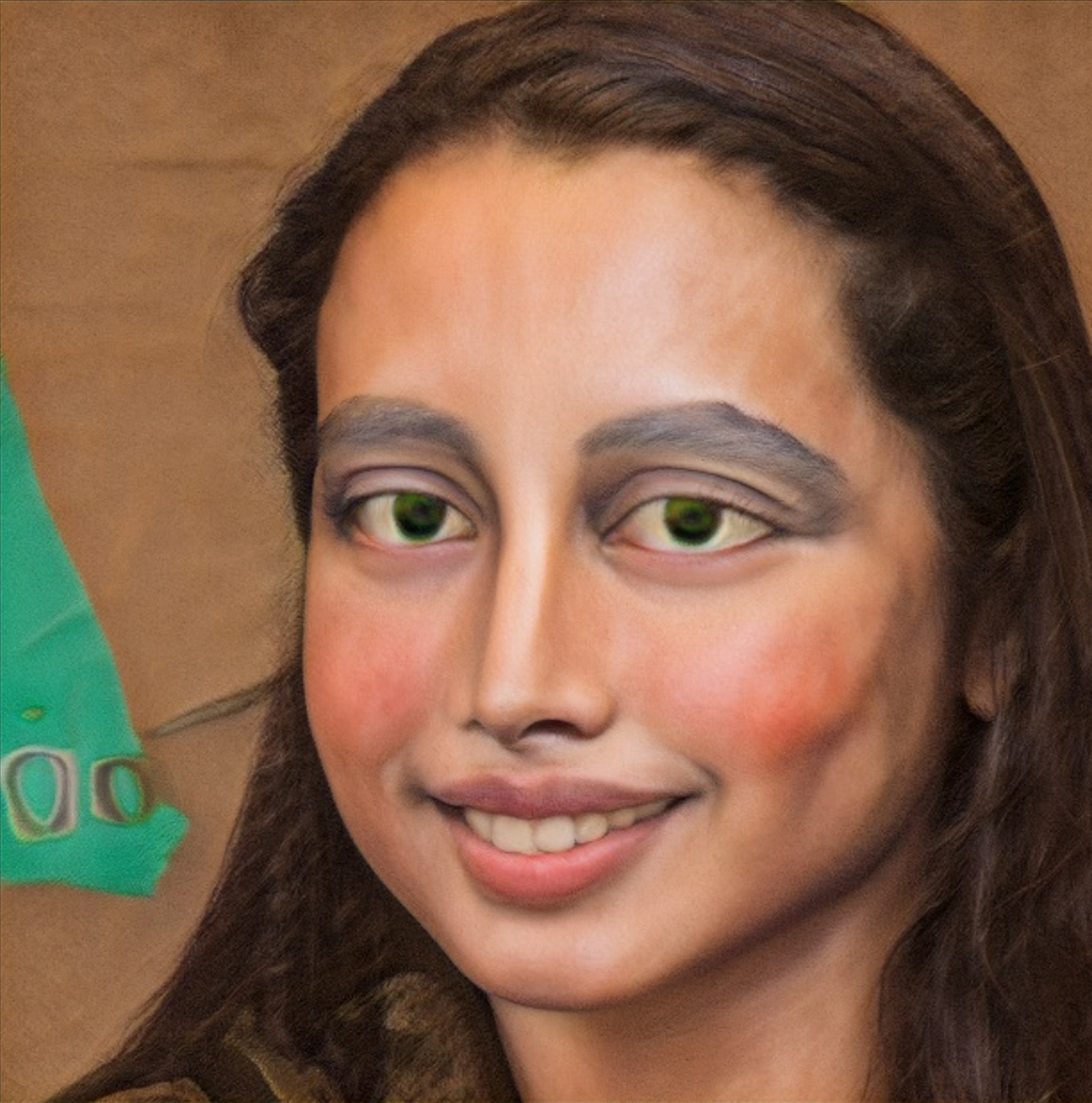} \\
        
        \raisebox{0.025\textwidth}{\rotatebox[origin=t]{90}{\begin{tabular}{c@{}c@{}}Dog $\rightarrow$ \\ The Joker\end{tabular}}} &
        \includegraphics[width=0.18\linewidth]{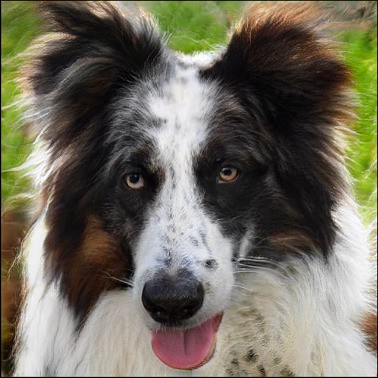} &
        \includegraphics[width=0.18\linewidth]{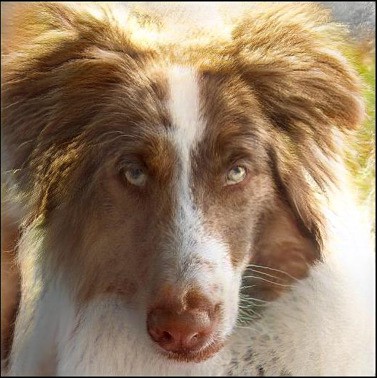} &
        \includegraphics[width=0.18\linewidth]{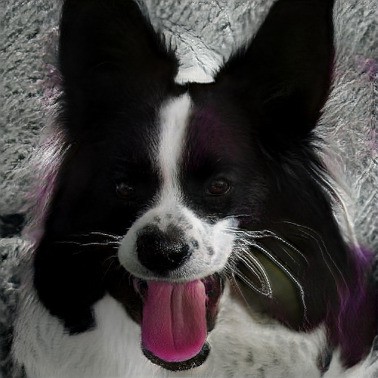} &
        \includegraphics[width=0.18\linewidth]{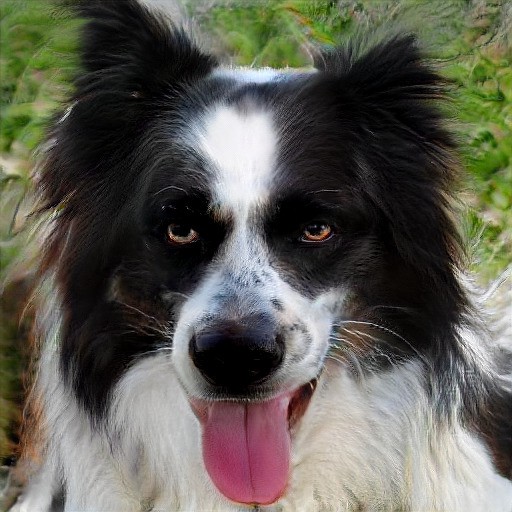} &
        \includegraphics[width=0.18\linewidth]{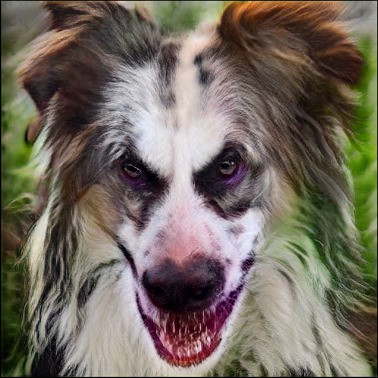} \\

        \raisebox{0.025\textwidth}{\rotatebox[origin=t]{90}{\begin{tabular}{c@{}c@{}}Dog $\rightarrow$ \\ Nicolas Cage\end{tabular}}} &
        \includegraphics[width=0.18\linewidth]{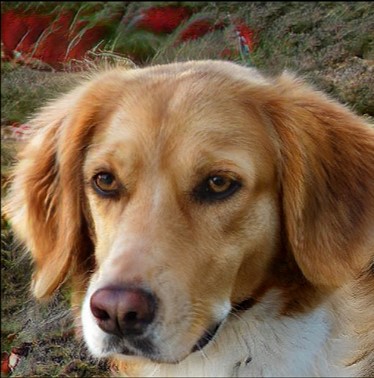} &
        \includegraphics[width=0.18\linewidth]{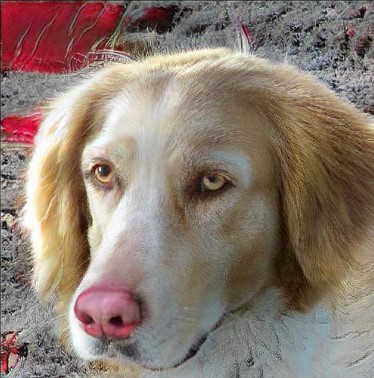} &
        \includegraphics[width=0.18\linewidth]{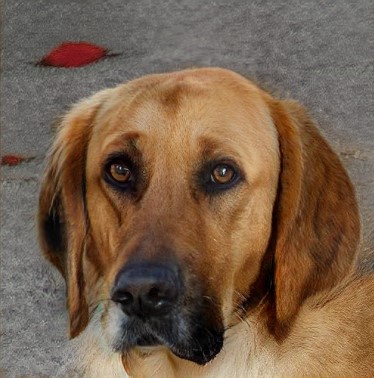} &
        \includegraphics[width=0.18\linewidth]{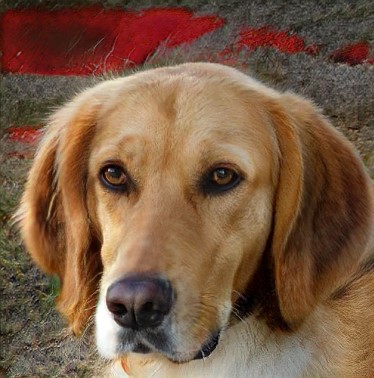} &
        \includegraphics[width=0.18\linewidth]{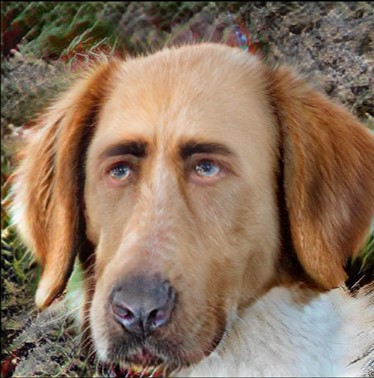} \\

        \raisebox{0.025\textwidth}{\rotatebox[origin=t]{90}{\begin{tabular}{c@{}c@{}}Church $\rightarrow$ \\ The Shire\end{tabular}}} &
        \includegraphics[width=0.18\linewidth]{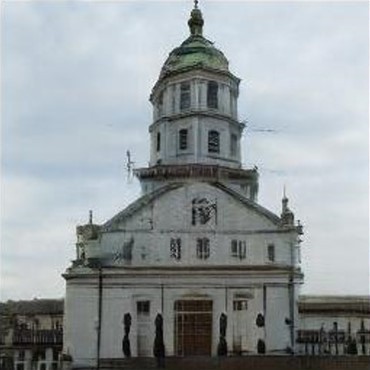} &
        \includegraphics[width=0.18\linewidth]{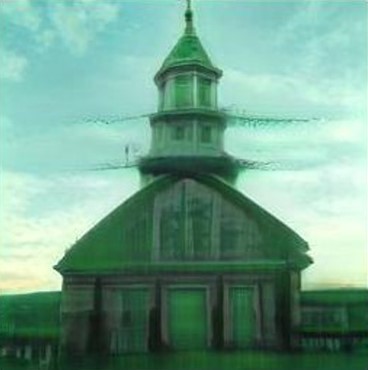} &
        \includegraphics[width=0.18\linewidth]{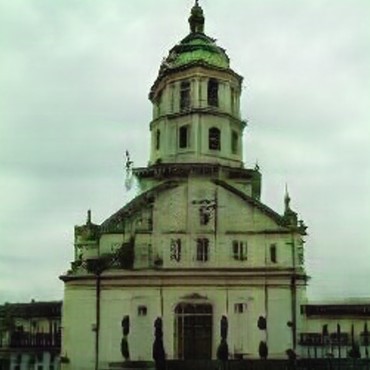} &
        \includegraphics[width=0.18\linewidth]{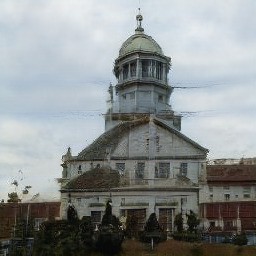} &
        \includegraphics[width=0.18\linewidth]{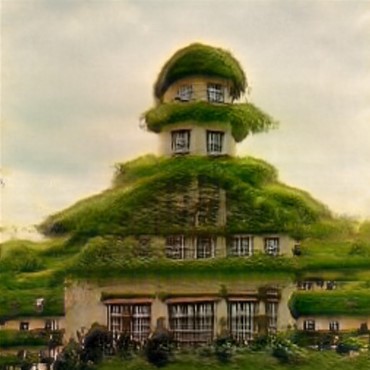} \\
        
        &
        \begin{tabular}{@{}c@{}}$G_{frozen}\left(w\right)$\end{tabular} &
        \begin{tabular}{@{}c@{}}StyleCLIP \\ Latent Optimization\end{tabular} &
        \begin{tabular}{@{}c@{}}StyleCLIP \\ Latent Mapper\end{tabular} &
        \begin{tabular}{@{}c@{}}StyleCLIP \\ Global Directions\end{tabular} &
        \begin{tabular}{@{}c@{}}Ours\end{tabular}

    \end{tabular}
    }
    \caption{Out-of-domain manipulation comparisons to StyleCLIP \cite{patashnik2021styleclip}. In each row we show: an image synthesized with a randomly sampled code (left), the results of editing the same code towards an out-of-domain textual direction using all three StyleCLIP \cite{patashnik2021styleclip} methods, and the image produced for the same code with a generator converted using our method (right). }
    \label{fig:styleclip_comparisons}
    \vspace{-2pt}
\end{figure}

\vspace{-5pt}
\paragraph{Text-guided editing.}
We show that existing editing techniques that operate \emph{within} the domain of a pre-trained generator, fail to shift images beyond said domain. In particular, we are interested in text-driven methods that operate without additional data. A natural comparison is then StyleCLIP and its three CLIP-guided editing approaches. \cref{fig:styleclip_comparisons} shows the results of such a comparison. As can be seen, none of StyleCLIP's approaches succeed in performing out-of-domain manipulations, even when only minor changes are needed (such as inducing celebrity identities on dogs).

\vspace{-2pt}
\paragraph{Few-shot generators.}
We compare StyleGAN-NADA with several few-shot alternatives: Ojha \etal \cite{ojha2021few}, MineGAN \cite{Wang_2020_CVPR}, TGAN~\cite{Wang2018TransferringGG} and TGAN + ADA~\cite{Karras2020ada}. In all cases, we convert the official StyleGAN-ADA AFHQ-Dog model~\cite{Karras2020ada, choi2020starganv2} to a cat model. Our method operates in a zero-shot manner. Other methods were trained using samples of $5$, $10$ and $100$ images from AFHQ-Cat\cite{choi2020starganv2}. We evaluate two aspects of these models --- quality and diversity. Quality is measured using a user study with a two-alternative forced choice setup. Users were presented with one of our generated images, and one from a competing method. They were asked to pick the image portraying a higher-quality cat. We gathered 1200 responses from 218 unique users. We report the percentage of users which preferred each method to our own. For diversity, we follow Ojah \etal~\cite{ojha2021few} by clustering the data and computing an average LPIPS~\cite{zhang2018perceptual} distance within the clusters. However, their method clusters according to LPIPS distances from the training set. As our own method does not use a training set, we cluster around generated images using K-Medoids\cite{pam1990medoids}. %

\begin{figure}[t]\vspace{-2pt}
    \centering
    \includegraphics[width=0.49\textwidth]{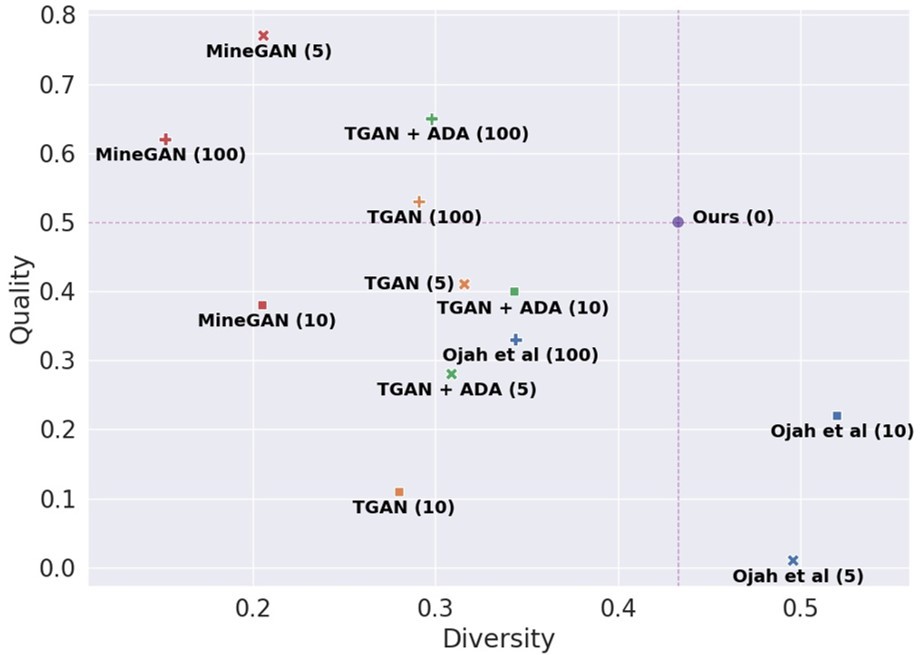} 
    \vspace{-12pt}
    \caption{Quality $\left(\uparrow\right)$ and diversity  $\left(\uparrow\right)$ comparison for our method and selected few-shot approaches. Numbers in parenthesis denote the number of training images. Our model pushes the Pareto front, and it does so without using a single image.}
    \label{fig:few_shot_quantitative}
    \vspace{-0pt}
\end{figure}
\begin{figure}[!hbt]
    \centering
    \includegraphics[width=0.49\textwidth]{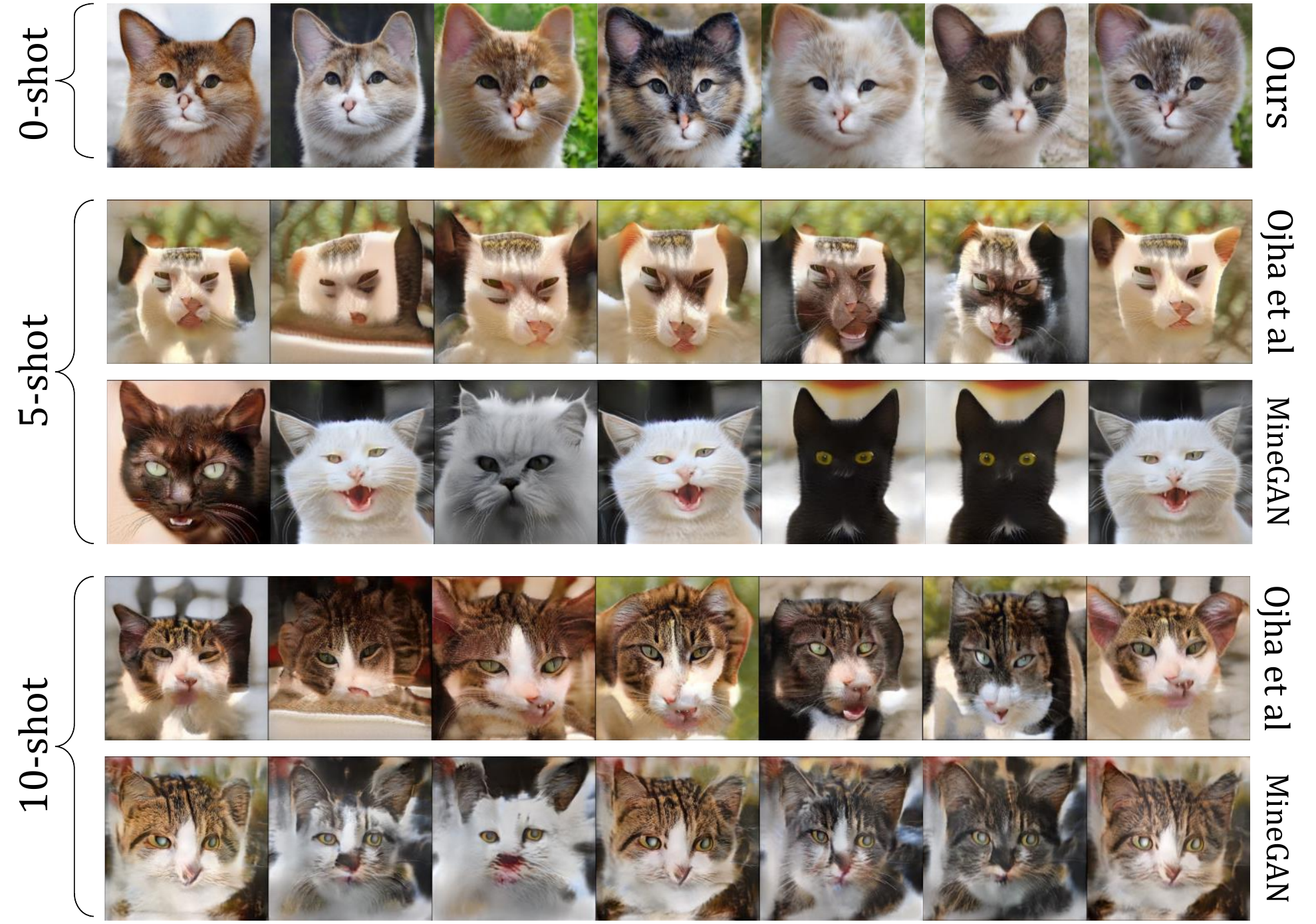} 
    \vspace{-10pt}
    \caption{Image synthesis using models adapted from StyleGAN-ADA \cite{Karras2020ada} AFHQ-Dog \cite{choi2020starganv2} to the cat domain. We compare to two few shot models - Ojha \etal \cite{ojha2021few} and MineGAN \cite{Wang_2020_CVPR}. Next to each method we list the number of training images used during training. With 5 examples, MineGAN memorizes the training set. }
    \label{fig:few_shot}
    \vspace{-8pt}
\end{figure}
Results are shown in \cref{fig:few_shot_quantitative}. Our model consistently outperforms most $5$ and $10$-shot methods in terms of quality, and displays improved diversity even when compared to methods trained with a hundred images. With $5$ images, MineGAN~\cite{Wang_2020_CVPR} memorizes the data. \cref{fig:few_shot} shows qualitative results for selected methods. See \cref{sec:few_shot_qual} for more.

In addition to these comparisons, we find that in many cases, using our method as a pre-training step before employing a few-shot method improves synthesis performance. See \cref{sec:few_shot_clip}.

\subsection{Ablation study}
\begin{figure}[!hbt]
    \centering
    \setlength{\belowcaptionskip}{-0.0pt}
    \setlength{\tabcolsep}{1pt}
    {\tiny
    \begin{tabular}{c c c c c c c}
    
        \raisebox{0.02\textwidth}{\rotatebox[origin=t]{90}{\begin{tabular}{c@{}c@{}} Dog $\rightarrow$ \\  Cat \end{tabular}}} & 
        \includegraphics[width=0.15\linewidth]{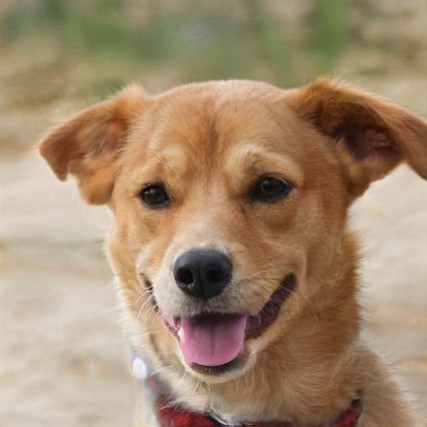} &
        \includegraphics[width=0.15\linewidth]{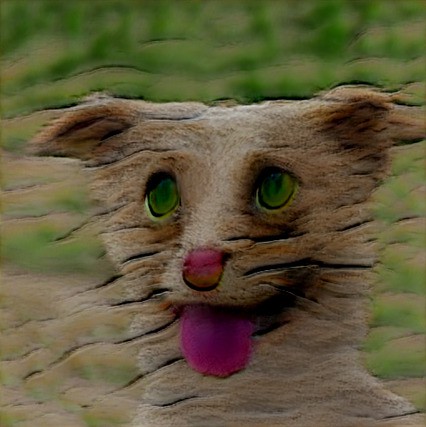} &
        \includegraphics[width=0.15\linewidth]{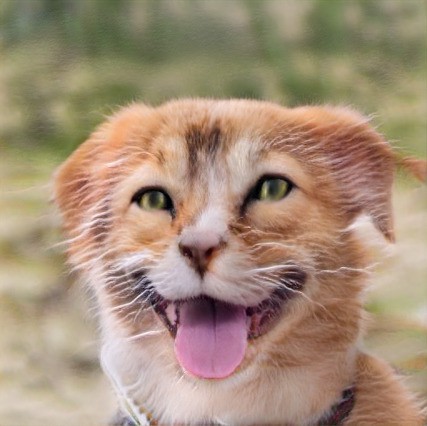} &
        \includegraphics[width=0.15\linewidth]{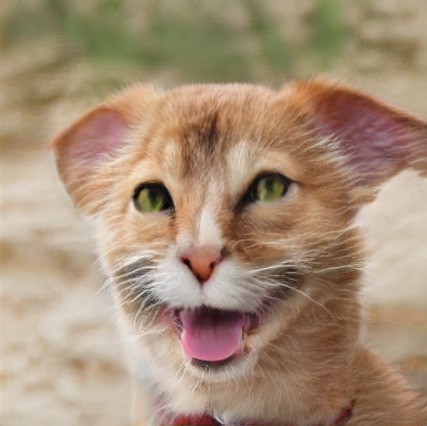} &
        \includegraphics[width=0.15\linewidth]{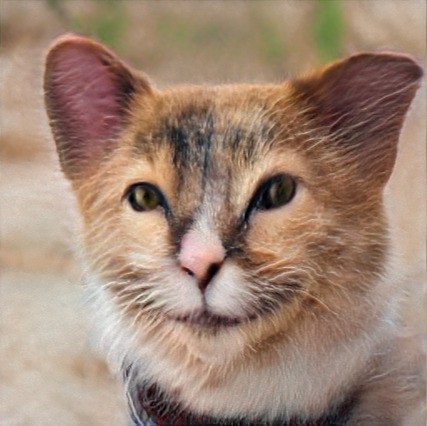} &
        \includegraphics[width=0.15\linewidth]{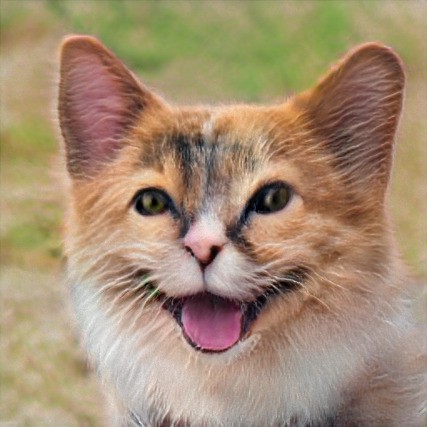}
        \\
        
        \raisebox{0.02\textwidth}{\rotatebox[origin=t]{90}{\begin{tabular}{c@{}c@{}} Dog $\rightarrow$ \\  Quokka\end{tabular}}} & 
        \includegraphics[width=0.15\linewidth]{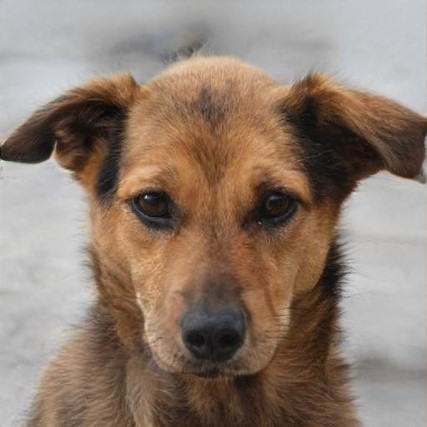} &
        \includegraphics[width=0.15\linewidth]{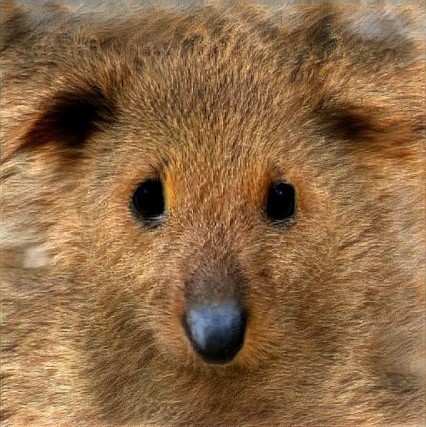} &
        \includegraphics[width=0.15\linewidth]{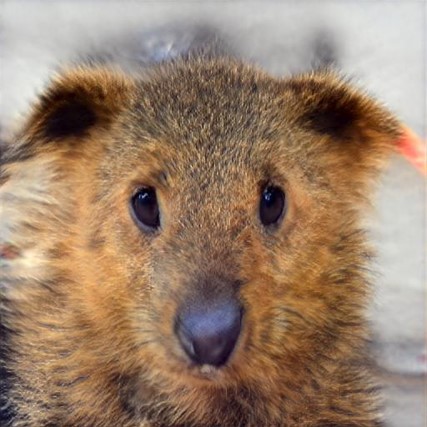} &
        \includegraphics[width=0.15\linewidth]{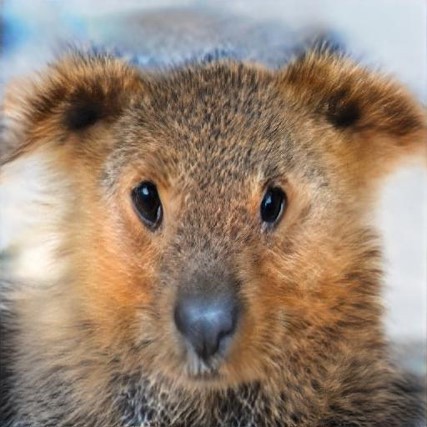} &
        \includegraphics[width=0.15\linewidth]{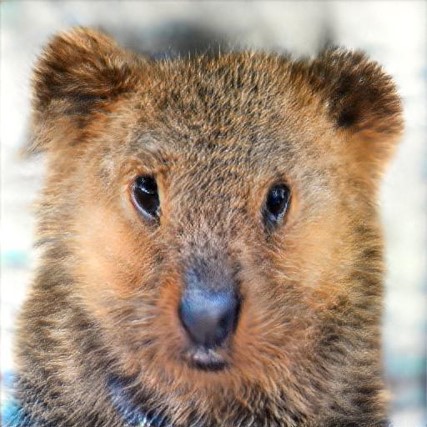} &
        \includegraphics[width=0.15\linewidth]{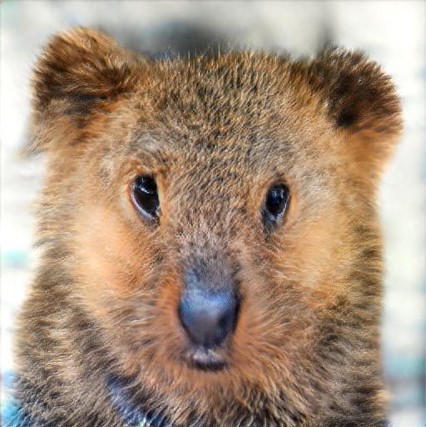}
        \\
    
        \raisebox{0.02\textwidth}{\rotatebox[origin=t]{90}{\begin{tabular}{c@{}c@{}} Human $\rightarrow$ \\  Plastic Puppet\end{tabular}}} & 
        \includegraphics[width=0.15\linewidth]{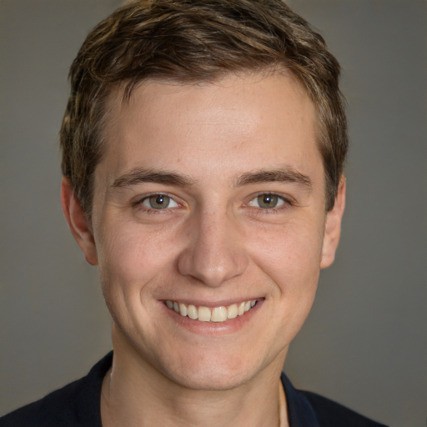} &
        \includegraphics[width=0.15\linewidth]{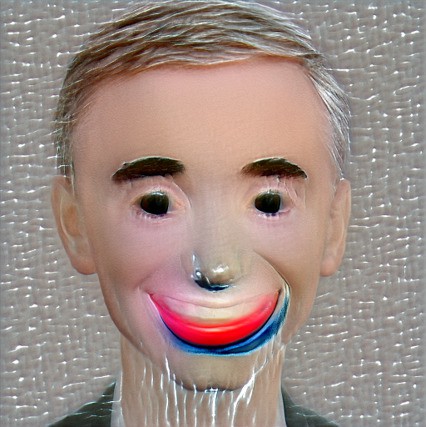} &
        \includegraphics[width=0.15\linewidth]{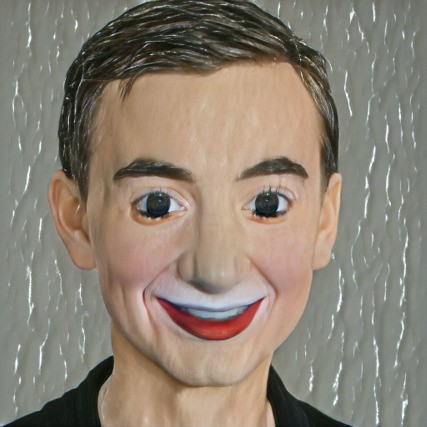} &
        \includegraphics[width=0.15\linewidth]{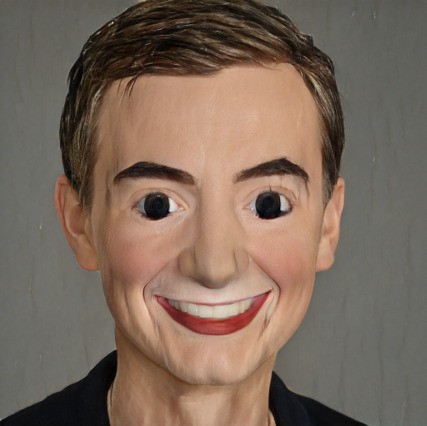} &
        \includegraphics[width=0.15\linewidth]{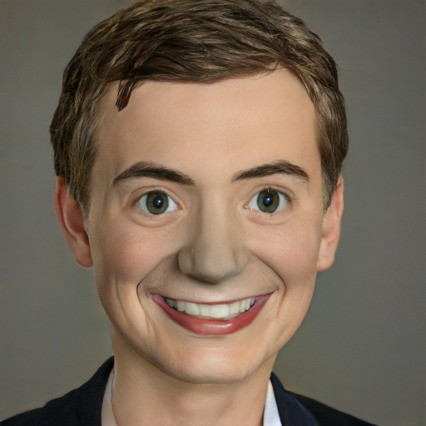} &
        \includegraphics[width=0.15\linewidth]{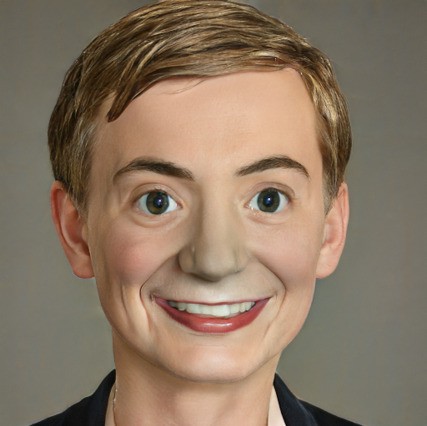}
        \\

        \raisebox{0.02\textwidth}{\rotatebox[origin=t]{90}{\begin{tabular}{c@{}c@{}} Photo $\rightarrow$ Sierra \\ Quest Graphics \end{tabular}}} & 
        \includegraphics[width=0.15\linewidth]{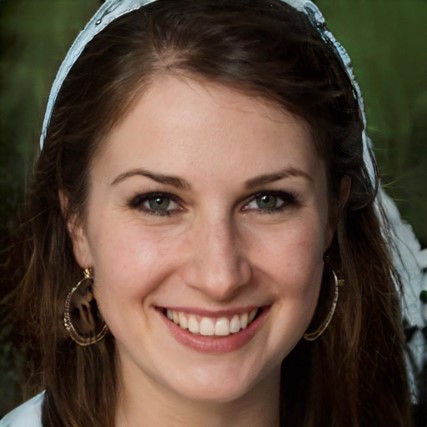} &
        \includegraphics[width=0.15\linewidth]{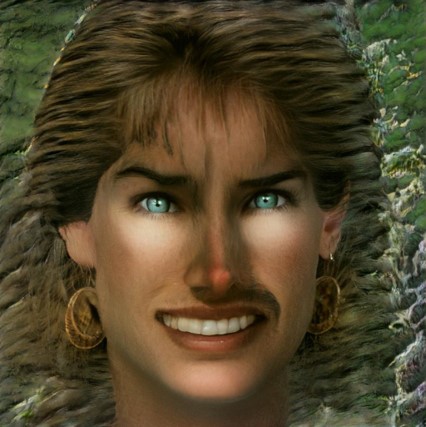} &
        \includegraphics[width=0.15\linewidth]{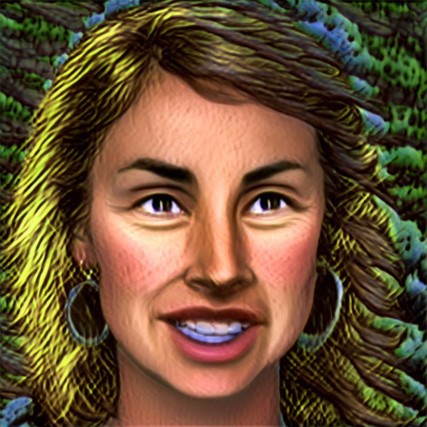} &
        \includegraphics[width=0.15\linewidth]{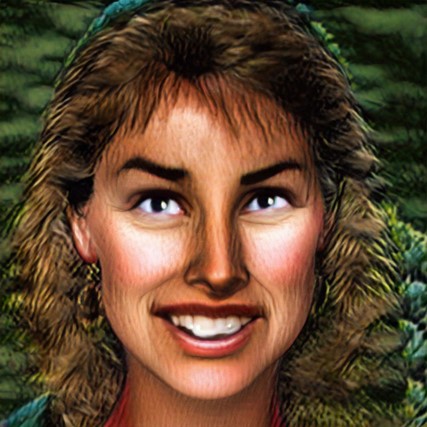} &
        \includegraphics[width=0.15\linewidth]{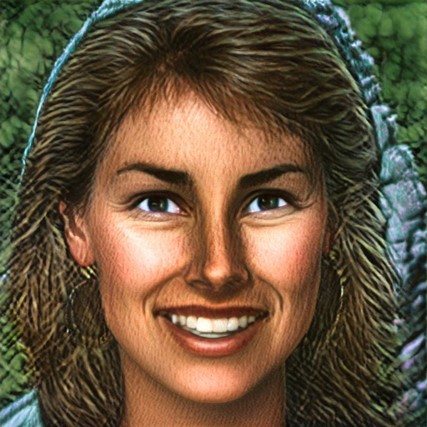} &
        \includegraphics[width=0.15\linewidth]{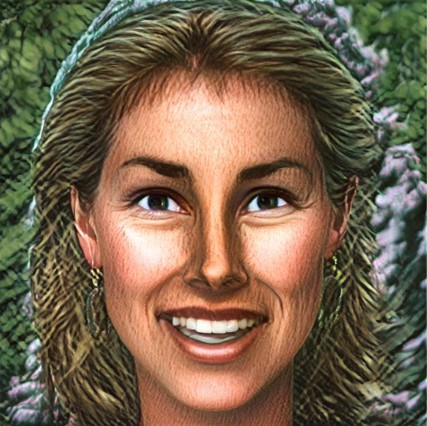}
        \\
        
        & { $G_{frzn}\left(w\right)$ } & \begin{tabular}{@{}c@{}}Global \\ Loss\end{tabular} & \begin{tabular}{@{}c@{}}All \\ Layers\end{tabular} & \begin{tabular}{@{}c@{}}Manual \\ Layers\end{tabular} & \begin{tabular}{@{}c@{}} Adaptive \\ Layers\end{tabular} & \begin{tabular}{@{}c@{}}With \\ Mapper\end{tabular}

    \end{tabular}}
    \vspace{-3pt}
    \caption{
     Images synthesized by a transformed generator when changing individual network components. The first column shows images created by the source generator. We then show images produced by generators trained using: the global CLIP loss (\cref{eq:global_loss}), training all layers, manually selecting layers, using the adaptive layer-freezing method, and when adding a StyleCLIP \cite{patashnik2021styleclip} mapper. The mapper is only needed for more extensive shape changes.} \vspace{-5pt}
    \label{fig:ablation_qual}
\end{figure}

We evaluate our suggested modifications through an ablation study. See results in \cref{fig:ablation_qual}. The global loss approach consistently fails to produce meaningful results, across all domains and modifications. Meanwhile, our directional-loss model with adaptive layer-freezing achieves the best visual results. In some cases, quality can be improved further by training a latent mapper.

\vspace{-1pt}
\section{Conclusions}
\label{sec:future}

We presented StyleGAN-NADA, a CLIP-guided zero-shot method for Non-Adversarial Domain Adaptation of image generators. By using CLIP to guide the training of a generator, rather than an exploration of its latent space, we are able to 
affect large changes in both style and shape, far beyond the generator's original domain.

The ability to train generators without data leads to exciting new possibilities - from editing images in ways that are constrained almost only by the user's creativity, to synthesizing paired cross-domain data for downstream applications such as image-to-image translation.

Our method, however, is not without limitations. By relying on CLIP, we are limited to concepts that CLIP has observed. Textual guidance is also inherently limited by the ambiguity of natural language prompts. When one describes a `Raphael Painting', for example, do they refer to the artistic style of the Renaissance painter, a portrait of his likeness, or an animated turtle bearing that name? 

Our method works particularly well for style and fine details, but it may struggle with large scale attributes or geometry changes. Such restrictions are common also in few-shot approaches. We find that a good translation often requires a fairly similar pre-trained generator as a starting point.

We focused on transforming existing generators. An intriguing question is whether one can do away with this requirement and train a generator from scratch, using only CLIP's guidance. While such an idea may seem outlandish, recent progress in inverting classifiers \cite{dong2021deep} and in generative art \cite{katherine2021vqganclip,murdock2021bigsleep} gives us hope that it is not beyond reach.

We hope our work can inspire others to continue exploring the world of textually-guided generation, and particularly the astounding ability of CLIP to guide visual transformations. Perhaps, not too long in the future, our day-to-day efforts would no longer be constrained by data requirements - but only by our creativity.

\vspace{-3pt}
\paragraph{Acknowledgments} We thank Yuval Alaluf, Ron Mokady and Ethan Fetaya for reviewing early drafts and helpful suggestions. Assaf Hallak for discussions, and Zonzge Wu for assistance with StyleCLIP comparisons.

\bibliographystyle{plain}
\bibliography{main}

\begin{thebibliography}{10}

\bibitem{pam1990medoids}
{\em Partitioning Around Medoids (Program PAM)}, chapter~2, pages 68--125.
\newblock John Wiley and Sons, Ltd, 1990.

\bibitem{abdal2019image2stylegan}
Rameen Abdal, Yipeng Qin, and Peter Wonka.
\newblock Image2stylegan: How to embed images into the stylegan latent space?
\newblock In {\em Proceedings of the IEEE/CVF International Conference on
  Computer Vision}, pages 4432--4441, 2019.

\bibitem{10.1145/3447648}
Rameen Abdal, Peihao Zhu, Niloy~J. Mitra, and Peter Wonka.
\newblock Styleflow: Attribute-conditioned exploration of stylegan-generated
  images using conditional continuous normalizing flows.
\newblock {\em ACM Trans. Graph.}, 40(3), May 2021.

\bibitem{alaluf2021matter}
Yuval Alaluf, Or~Patashnik, and Daniel Cohen-Or.
\newblock Only a matter of style: Age transformation using a style-based
  regression model, 2021.

\bibitem{alaluf2021restyle}
Yuval Alaluf, Or~Patashnik, and Daniel Cohen-Or.
\newblock Restyle: A residual-based stylegan encoder via iterative refinement.
\newblock {\em arXiv preprint arXiv:2104.02699}, 2021.

\bibitem{sklearn2019extra}
Christos Aridas, Jan-Oliver Joswig, Timothée Mathieu, and Roman Yurchak.
\newblock scikit-learn-extra module, 2019.
\newblock \url{https://scikit-learn-extra.readthedocs.io/en/stable/index.html}.

\bibitem{bau2021paint}
David Bau, Alex Andonian, Audrey Cui, YeonHwan Park, Ali Jahanian, Aude Oliva,
  and Antonio Torralba.
\newblock Paint by word, 2021.

\bibitem{brock2018large}
Andrew Brock, Jeff Donahue, and Karen Simonyan.
\newblock Large scale gan training for high fidelity natural image synthesis.
\newblock {\em arXiv preprint arXiv:1809.11096}, 2018.

\bibitem{caesar2018coco}
Holger Caesar, Jasper Uijlings, and Vittorio Ferrari.
\newblock Coco-stuff: Thing and stuff classes in context.
\newblock In {\em Computer vision and pattern recognition (CVPR), 2018 IEEE
  conference on}. IEEE, 2018.

\bibitem{Chen2020UNITERUI}
Yen-Chun Chen, Linjie Li, Licheng Yu, A.~E. Kholy, Faisal Ahmed, Zhe Gan,
  Y.~Cheng, and Jing jing Liu.
\newblock Uniter: Universal image-text representation learning.
\newblock In {\em ECCV}, 2020.

\bibitem{choi2020starganv2}
Yunjey Choi, Youngjung Uh, Jaejun Yoo, and Jung-Woo Ha.
\newblock Stargan v2: Diverse image synthesis for multiple domains.
\newblock In {\em Proceedings of the IEEE Conference on Computer Vision and
  Pattern Recognition}, 2020.

\bibitem{katherine2021vqganclip}
Katherine Crowson.
\newblock Vqgan + clip, 2021.
\newblock
  \url{https://colab.research.google.com/drive/1L8oL-vLJXVcRzCFbPwOoMkPKJ8-aYdPN}.

\bibitem{Desai2020VirTexLV}
Karan Desai and J.~Johnson.
\newblock {VirTex}: Learning visual representations from textual annotations.
\newblock {\em ArXiv}, abs/2006.06666, 2020.

\bibitem{dong2021deep}
Xin Dong, Hongxu Yin, Jose~M. Alvarez, Jan Kautz, and Pavlo Molchanov.
\newblock Deep neural networks are surprisingly reversible: A baseline for
  zero-shot inversion, 2021.

\bibitem{dosovitskiy2020vit}
Alexey Dosovitskiy, Lucas Beyer, Alexander Kolesnikov, Dirk Weissenborn,
  Xiaohua Zhai, Thomas Unterthiner, Mostafa Dehghani, Matthias Minderer, Georg
  Heigold, Sylvain Gelly, Jakob Uszkoreit, and Neil Houlsby.
\newblock An image is worth 16x16 words: Transformers for image recognition at
  scale.
\newblock {\em ICLR}, 2021.

\bibitem{pearson1901pca}
Karl~Pearson F.R.S.
\newblock On lines and planes of closest fit to systems of points in space.
\newblock {\em The London, Edinburgh, and Dublin Philosophical Magazine and
  Journal of Science}, 2(11):559--572, 1901.

\bibitem{goodfellow2014generative}
Ian Goodfellow, Jean Pouget-Abadie, Mehdi Mirza, Bing Xu, David Warde-Farley,
  Sherjil Ozair, Aaron Courville, and Yoshua Bengio.
\newblock Generative adversarial nets.
\newblock {\em Advances in neural information processing systems}, 27, 2014.

\bibitem{goodfellow2014explaining}
Ian~J Goodfellow, Jonathon Shlens, and Christian Szegedy.
\newblock Explaining and harnessing adversarial examples.
\newblock {\em arXiv preprint arXiv:1412.6572}, 2014.

\bibitem{harkonen2020ganspace}
Erik H{\"a}rk{\"o}nen, Aaron Hertzmann, Jaakko Lehtinen, and Sylvain Paris.
\newblock Ganspace: Discovering interpretable gan controls.
\newblock {\em arXiv preprint arXiv:2004.02546}, 2020.

\bibitem{Karras2020ada}
Tero Karras, Miika Aittala, Janne Hellsten, Samuli Laine, Jaakko Lehtinen, and
  Timo Aila.
\newblock Training generative adversarial networks with limited data.
\newblock In {\em Proc. NeurIPS}, 2020.

\bibitem{karras2021aliasfree}
Tero Karras, Miika Aittala, Samuli Laine, Erik Härkönen, Janne Hellsten,
  Jaakko Lehtinen, and Timo Aila.
\newblock Alias-free generative adversarial networks, 2021.

\bibitem{karras2019style}
Tero Karras, Samuli Laine, and Timo Aila.
\newblock A style-based generator architecture for generative adversarial
  networks.
\newblock In {\em Proceedings of the IEEE conference on computer vision and
  pattern recognition}, pages 4401--4410, 2019.

\bibitem{karras2020analyzing}
Tero Karras, Samuli Laine, Miika Aittala, Janne Hellsten, Jaakko Lehtinen, and
  Timo Aila.
\newblock Analyzing and improving the image quality of stylegan.
\newblock In {\em Proceedings of the IEEE/CVF Conference on Computer Vision and
  Pattern Recognition}, pages 8110--8119, 2020.

\bibitem{ledig2017photo}
Christian Ledig, Lucas Theis, Ferenc Husz{\'a}r, Jose Caballero, Andrew
  Cunningham, Alejandro Acosta, Andrew Aitken, Alykhan Tejani, Johannes Totz,
  Zehan Wang, et~al.
\newblock Photo-realistic single image super-resolution using a generative
  adversarial network.
\newblock In {\em Proceedings of the IEEE conference on computer vision and
  pattern recognition}, pages 4681--4690, 2017.

\bibitem{Li2020UnicoderVLAU}
Gen Li, N.~Duan, Yuejian Fang, Daxin Jiang, and M.~Zhou.
\newblock Unicoder-{VL}: A universal encoder for vision and language by
  cross-modal pre-training.
\newblock In {\em Proc.~AAAI}, 2020.

\bibitem{Li2019VisualBERTAS}
Liunian~Harold Li, Mark Yatskar, Da~Yin, C.~Hsieh, and Kai-Wei Chang.
\newblock Visualbert: A simple and performant baseline for vision and language.
\newblock {\em ArXiv}, abs/1908.03557, 2019.

\bibitem{Li2020OscarOA}
Xiujun Li, Xi~Yin, C.~Li, X.~Hu, Pengchuan Zhang, Lei Zhang, Longguang Wang,
  H.~Hu, Li~Dong, Furu Wei, Yejin Choi, and Jianfeng Gao.
\newblock Oscar: Object-semantics aligned pre-training for vision-language
  tasks.
\newblock In {\em ECCV}, 2020.

\bibitem{li2020few}
Yijun Li, Richard Zhang, Jingwan Lu, and Eli Shechtman.
\newblock Few-shot image generation with elastic weight consolidation.
\newblock {\em arXiv preprint arXiv:2012.02780}, 2020.

\bibitem{liu2020towards}
Bingchen Liu, Yizhe Zhu, Kunpeng Song, and Ahmed Elgammal.
\newblock Towards faster and stabilized gan training for high-fidelity few-shot
  image synthesis.
\newblock In {\em International Conference on Learning Representations}, 2020.

\bibitem{Lu2019ViLBERTPT}
Jiasen Lu, Dhruv Batra, D.~Parikh, and Stefan Lee.
\newblock Vilbert: Pretraining task-agnostic visiolinguistic representations
  for vision-and-language tasks.
\newblock In {\em NeurIPS}, 2019.

\bibitem{mo2020freeze}
Sangwoo Mo, Minsu Cho, and Jinwoo Shin.
\newblock Freeze the discriminator: a simple baseline for fine-tuning gans.
\newblock In {\em CVPR AI for Content Creation Workshop}, 2020.

\bibitem{murdock2021bigsleep}
Ryan Murdock.
\newblock The big sleep, 2021.
\newblock \url{https://twitter.com/advadnoun/status/1351038053033406468}.

\bibitem{nitzan2021large}
Yotam Nitzan, Rinon Gal, Ofir Brenner, and Daniel Cohen-Or.
\newblock Large: Latent-based regression through gan semantics, 2021.

\bibitem{noguchi2019image}
Atsuhiro Noguchi and Tatsuya Harada.
\newblock Image generation from small datasets via batch statistics adaptation.
\newblock In {\em Proceedings of the IEEE/CVF International Conference on
  Computer Vision}, pages 2750--2758, 2019.

\bibitem{ojha2021few}
Utkarsh Ojha, Yijun Li, Jingwan Lu, Alexei~A Efros, Yong~Jae Lee, Eli
  Shechtman, and Richard Zhang.
\newblock Few-shot image generation via cross-domain correspondence.
\newblock In {\em Proceedings of the IEEE/CVF Conference on Computer Vision and
  Pattern Recognition}, pages 10743--10752, 2021.

\bibitem{park2019SPADE}
Taesung Park, Ming-Yu Liu, Ting-Chun Wang, and Jun-Yan Zhu.
\newblock Semantic image synthesis with spatially-adaptive normalization.
\newblock In {\em Proceedings of the IEEE Conference on Computer Vision and
  Pattern Recognition}, 2019.

\bibitem{patashnik2021styleclip}
Or~Patashnik, Zongze Wu, Eli Shechtman, Daniel Cohen-Or, and Dani Lischinski.
\newblock Styleclip: Text-driven manipulation of stylegan imagery.
\newblock {\em arXiv preprint arXiv:2103.17249}, 2021.

\bibitem{pinkney2020resolution}
Justin~NM Pinkney and Doron Adler.
\newblock Resolution dependent gan interpolation for controllable image
  synthesis between domains.
\newblock {\em arXiv preprint arXiv:2010.05334}, 2020.

\bibitem{radford2021learning}
Alec Radford, Jong~Wook Kim, Chris Hallacy, Aditya Ramesh, Gabriel Goh,
  Sandhini Agarwal, Girish Sastry, Amanda Askell, Pamela Mishkin, Jack Clark,
  et~al.
\newblock Learning transferable visual models from natural language
  supervision.
\newblock {\em arXiv preprint arXiv:2103.00020}, 2021.

\bibitem{richardson2020encoding}
Elad Richardson, Yuval Alaluf, Or~Patashnik, Yotam Nitzan, Yaniv Azar, Stav
  Shapiro, and Daniel Cohen-Or.
\newblock Encoding in style: a stylegan encoder for image-to-image translation.
\newblock {\em arXiv preprint arXiv:2008.00951}, 2020.

\bibitem{Robb2020FewShotAO}
Esther Robb, Wen-Sheng Chu, Abhishek Kumar, and Jia-Bin Huang.
\newblock Few-shot adaptation of generative adversarial networks.
\newblock {\em ArXiv}, abs/2010.11943, 2020.

\bibitem{sariyildiz2020learning}
Mert~Bulent Sariyildiz, Julien Perez, and Diane Larlus.
\newblock Learning visual representations with caption annotations.
\newblock {\em arXiv preprint arXiv:2008.01392}, 2020.

\bibitem{schonfeld2021oasis}
Edgar Sch{\"o}nfeld, Vadim Sushko, Dan Zhang, Juergen Gall, Bernt Schiele, and
  Anna Khoreva.
\newblock You only need adversarial supervision for semantic image synthesis.
\newblock In {\em International Conference on Learning Representations}, 2021.

\bibitem{rosinalitySG2}
Kim Seonghyeon.
\newblock Stylegan2-pytorch.
\newblock \url{https://github.com/rosinality/stylegan2-pytorch}, 2020.

\bibitem{shen2020interpreting}
Yujun Shen, Jinjin Gu, Xiaoou Tang, and Bolei Zhou.
\newblock Interpreting the latent space of gans for semantic face editing.
\newblock In {\em Proceedings of the IEEE/CVF Conference on Computer Vision and
  Pattern Recognition}, pages 9243--9252, 2020.

\bibitem{10.1145/3450626.3459771}
Guoxian Song, Linjie Luo, Jing Liu, Wan-Chun Ma, Chunpong Lai, Chuanxia Zheng,
  and Tat-Jen Cham.
\newblock Agilegan: Stylizing portraits by inversion-consistent transfer
  learning.
\newblock {\em ACM Trans. Graph.}, 40(4), July 2021.

\bibitem{Su2020VL-BERT:}
Weijie Su, Xizhou Zhu, Yue Cao, Bin Li, Lewei Lu, Furu Wei, and Jifeng Dai.
\newblock {VL-BERT}: Pre-training of generic visual-linguistic representations.
\newblock In {\em Proc.~ICLR}, 2020.

\bibitem{Tan2019LXMERTLC}
Hao~Hao Tan and Mohit Bansal.
\newblock {LXMERT}: Learning cross-modality encoder representations from
  transformers.
\newblock In {\em EMNLP/IJCNLP}, 2019.

\bibitem{tov2021designing}
Omer Tov, Yuval Alaluf, Yotam Nitzan, Or~Patashnik, and Daniel Cohen-Or.
\newblock Designing an encoder for stylegan image manipulation.
\newblock {\em arXiv preprint arXiv:2102.02766}, 2021.

\bibitem{tran2021data}
Ngoc-Trung Tran, Viet-Hung Tran, Ngoc-Bao Nguyen, Trung-Kien Nguyen, and
  Ngai-Man Cheung.
\newblock On data augmentation for gan training.
\newblock {\em IEEE Transactions on Image Processing}, 30:1882--1897, 2021.

\bibitem{Tseng2021RegularizingGA}
Hung-Yu Tseng, Lu~Jiang, Ce~Liu, Ming-Hsuan Yang, and Weilong Yang.
\newblock Regularizing generative adversarial networks under limited data.
\newblock {\em ArXiv}, abs/2104.03310, 2021.

\bibitem{NIPS2017_3f5ee243}
Ashish Vaswani, Noam Shazeer, Niki Parmar, Jakob Uszkoreit, Llion Jones,
  Aidan~N Gomez, \L{}ukasz Kaiser, and Illia Polosukhin.
\newblock Attention is all you need.
\newblock In {\em Advances in Neural Information Processing Systems},
  volume~30, 2017.

\bibitem{wang2021crossdomain}
Can Wang, Menglei Chai, Mingming He, Dongdong Chen, and Jing Liao.
\newblock Cross-domain and disentangled face manipulation with 3d guidance,
  2021.

\bibitem{wang2018dni}
Xintao Wang, Ke~Yu, Chao Dong, Xiaoou Tang, and Chen~Change Loy.
\newblock Deep network interpolation for continuous imagery effect transition,
  2018.

\bibitem{Wang_2020_CVPR}
Yaxing Wang, Abel Gonzalez-Garcia, David Berga, Luis Herranz, Fahad~Shahbaz
  Khan, and Joost van~de Weijer.
\newblock Minegan: Effective knowledge transfer from gans to target domains
  with few images.
\newblock In {\em The IEEE/CVF Conference on Computer Vision and Pattern
  Recognition (CVPR)}, June 2020.

\bibitem{Wang2018TransferringGG}
Yaxing Wang, Chenshen Wu, Luis Herranz, Joost van~de Weijer, Abel
  Gonzalez-Garcia, and B.~Raducanu.
\newblock Transferring gans: generating images from limited data.
\newblock In {\em ECCV}, 2018.

\bibitem{wu2021stylealign}
Zongze Wu, Yotam Nitzan, Eli Shechtman, and Dani Lischinski.
\newblock Stylealign: Analysis and applications of aligned stylegan models,
  2021.

\bibitem{xia2021gan}
Weihao Xia, Yulun Zhang, Yujiu Yang, Jing-Hao Xue, Bolei Zhou, and Ming-Hsuan
  Yang.
\newblock Gan inversion: A survey, 2021.

\bibitem{xu2021generative}
Yinghao Xu, Yujun Shen, Jiapeng Zhu, Ceyuan Yang, and Bolei Zhou.
\newblock Generative hierarchical features from synthesizing images.
\newblock In {\em CVPR}, 2021.

\bibitem{yang2021data}
Ceyuan Yang, Yujun Shen, Yinghao Xu, and Bolei Zhou.
\newblock Data-efficient instance generation from instance discrimination.
\newblock {\em arXiv preprint arXiv:2106.04566}, 2021.

\bibitem{yang2021gan}
Tao Yang, Peiran Ren, Xuansong Xie, and Lei Zhang.
\newblock Gan prior embedded network for blind face restoration in the wild.
\newblock In {\em Proceedings of the IEEE/CVF Conference on Computer Vision and
  Pattern Recognition}, pages 672--681, 2021.

\bibitem{yu2015lsun}
Fisher Yu, Ari Seff, Yinda Zhang, Shuran Song, Thomas Funkhouser, and Jianxiong
  Xiao.
\newblock Lsun: Construction of a large-scale image dataset using deep learning
  with humans in the loop.
\newblock {\em arXiv preprint arXiv:1506.03365}, 2015.

\bibitem{zhang2018perceptual}
Richard Zhang, Phillip Isola, Alexei~A Efros, Eli Shechtman, and Oliver Wang.
\newblock The unreasonable effectiveness of deep features as a perceptual
  metric.
\newblock In {\em CVPR}, 2018.

\bibitem{zhao2020differentiable}
Shengyu Zhao, Zhijian Liu, Ji~Lin, Jun-Yan Zhu, and Song Han.
\newblock Differentiable augmentation for data-efficient gan training.
\newblock {\em arXiv preprint arXiv:2006.10738}, 2020.

\bibitem{zhao2020image}
Zhengli Zhao, Zizhao Zhang, Ting Chen, Sameer Singh, and Han Zhang.
\newblock Image augmentations for gan training.
\newblock {\em arXiv preprint arXiv:2006.02595}, 2020.

\end{thebibliography}

\clearpage

\title{Supplementary Materials \\ StyleGAN-NADA: CLIP-Guided Domain Adaptation of Image Generators }

\author{Rinon Gal$^{1,2*}$ \hspace{0.07\linewidth} \and
Or Patashnik$^{1}$ \hspace{0.07\linewidth} \and
Haggai Maron$^{2}$ \hspace{0.07\linewidth} \and 
Amit Bermano$^{1}$ \and
\hspace{0.1\linewidth} Gal Chechik$^{2}$ \hspace{0.2\linewidth} \and
Daniel Cohen-Or$^{1}$ \hspace{0.1\linewidth} \and \hspace{0.9\linewidth}
\and $^{1}$Tel-Aviv University \and $^{2}$NVIDIA
}

\maketitle

\appendix
\renewcommand{\floatpagefraction}{.8}
\renewcommand{\bottomfraction}{.7}
\section{Broader impact} 
This project aims to provide an effective content creation tool so that artists and others can unleash their creativity, as well as to support machine learning efforts in areas with limited data. At the same time, our tool can also be used for nefarious purposes in the wrong hands. 

As CLIP was trained on large collections of images and text from the internet, models using it for supervision are likely to propagate the biases inherent to such data -- and our model is no exception. Attempting to guide a face-generator conversion with the text `doctor', for example, causes the generator to produce mostly males, while using the text `nurse' has the opposite effect. In \cref{sec:few_shot_clip} we propose a method for mitigating this limitation using a small set of images.

\section{CLIP-space analysis} \label{sec:clip_analysis}

\begin{figure*}[!hb]
    \centering
    \includegraphics[width=\linewidth]{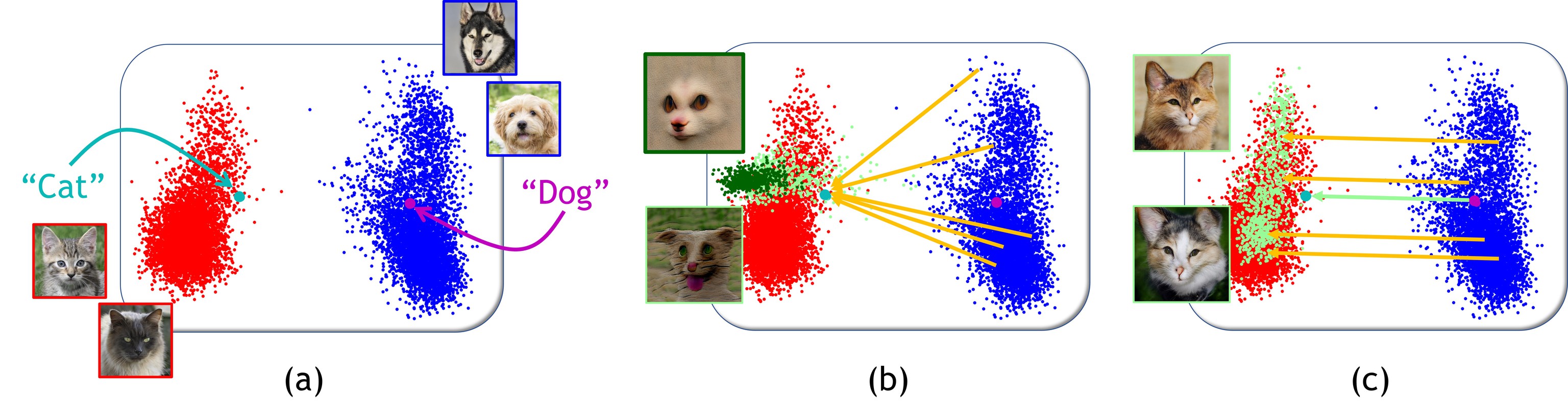}
    \caption{PCA-based visualization of the effects of the directional loss. (a) We embed real dog (blue) and cat (red) images \cite{choi2020starganv2} into CLIP's multi-modal space. We additionally embed the strings "Dog" (purple) and "Cat" (cyan). (b) A visualization of the direct, global loss. When the training objective is minimizing distances to one fixed point, we observe adversarial solutions (light green) and eventual collapse of all images to a single region (dark green). (c) A visualization of our directional loss. Generated dogs are shifted along a cross-domain direction, avoiding adversarial attacks and maintaining better diversity (light green).}
    \label{fig:clipspace_analysis}
    \vspace{-5pt}
\end{figure*}

We analyze the differences between our directional CLIP loss and the traditional global distance minimization approach by visualizing their behavior in CLIP's embedding space.
We first embed all images from the AFHQ cat and dog data sets \cite{choi2020starganv2} into CLIP's multi modal space. We then project them to 2D using PCA \cite{pearson1901pca}. We similarly embed and project the texts "Cat" and "Dog", as well as the fake images synthesized by our generator after training with both the global loss and the directional loss. The results are shown in \cref{fig:clipspace_analysis}.

Our results align with the intuition presented in \cref{sec:method}. In the case of the global CLIP loss (\cref{fig:clipspace_analysis}{\color{red}b}), we are optimizing towards a single target. There is no benefit to maintaining a diverse distribution, and results visibly collapse to a single region of the embedding space. In contrast, the directional loss (\cref{fig:clipspace_analysis}{\color{red}c}) discourages this collapse and successfully maintains a higher degree of diversity.

\section{Few-Shot CLIP-Guidance}\label{sec:few_shot_clip}
While our method focused on zero-shot domain adaptation, it is possible to leverage similar ideas for few-shot training. 
We investigate two alternative approaches for leveraging the semantic knowledge of CLIP for few-shot domain adaptation.

\paragraph{Image-based directions} In the first approach we consider the scenario where a small image set ($\sim3-5$ images) is available.
In such a scenario, rather than using a CLIP-space direction between two pairs of textual descriptions, it is possible to instead consider the CLIP-space direction between the images produced by the original generator and the domain represented by the small image set. Such an approach leverages CLIP in order to encode the \textit{semantic} difference between images from both domains. The directional loss then takes the form:
\vspace{-2pt}
\begin{equation}
\centering
\small
\hspace{-8pt}
\begin{gathered}
\Delta I_{real} = \frac{1}{N_{r}}\sum_{i=1}^{N_{r}}E_{I}\left(I_{i}\right) -\frac{1}{N_{s}}\sum_{i=1}^{N_{s}}E_{I}\left(G_{frozen}\left(w_i\right)\right)~, \\
\Delta I_{gen} = E_{I}\left(G_{train}\left(w\right)\right) - E_{I}\left(G_{frozen}\left(w\right)\right)~, \\ 
\mathcal{L}_{direction} = 1 - \frac{\Delta I_{gen} \cdot \Delta I_{real}}{\left|\Delta I_{gen}\right|\left|\Delta I_{real}\right|}~.
\end{gathered}
\end{equation}\label{eq:few_shot_loss}%
Here $N_r$ is the size of the real-image set, $I_i$ is the $i$-th image in the set, and $N_s$ is the number of images sampled from the source domain generator. Our experiments use $N_s = 16$.

This approach holds several advantages over standard few-shot methods: it better maintains the structure of the latent space, shows a higher degree of identity preservation (\cref{fig:sketch_supp}), trains in a fraction of the time, and does not require the images describing the target domain to be aligned nor preprocessed in a manner fitting the source domain.

\begin{figure}[t]
    \centering
    \setlength{\belowcaptionskip}{-5pt}
    \setlength{\tabcolsep}{1pt}
    {
    \begin{tabular}{c c}
        
        &
        \includegraphics[width=0.42\textwidth]{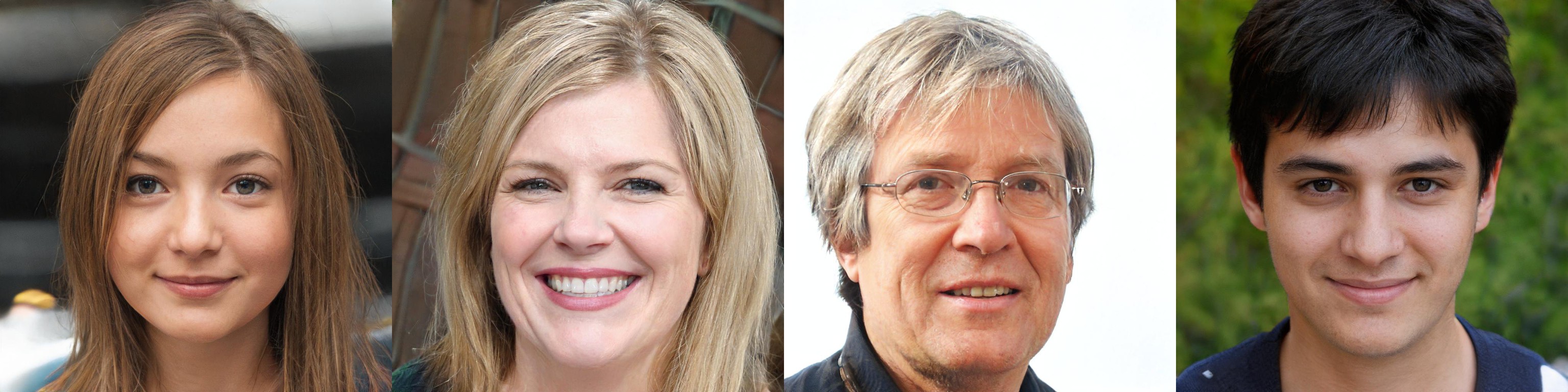} \vspace{-2pt} \\
        & Source \\
        \raisebox{0.025\textwidth}{\rotatebox[origin=t]{90}{\begin{tabular}{c@{}c@{}}Person $\rightarrow$ \\ Doctor\end{tabular}}} &
        \includegraphics[width=0.42\textwidth]{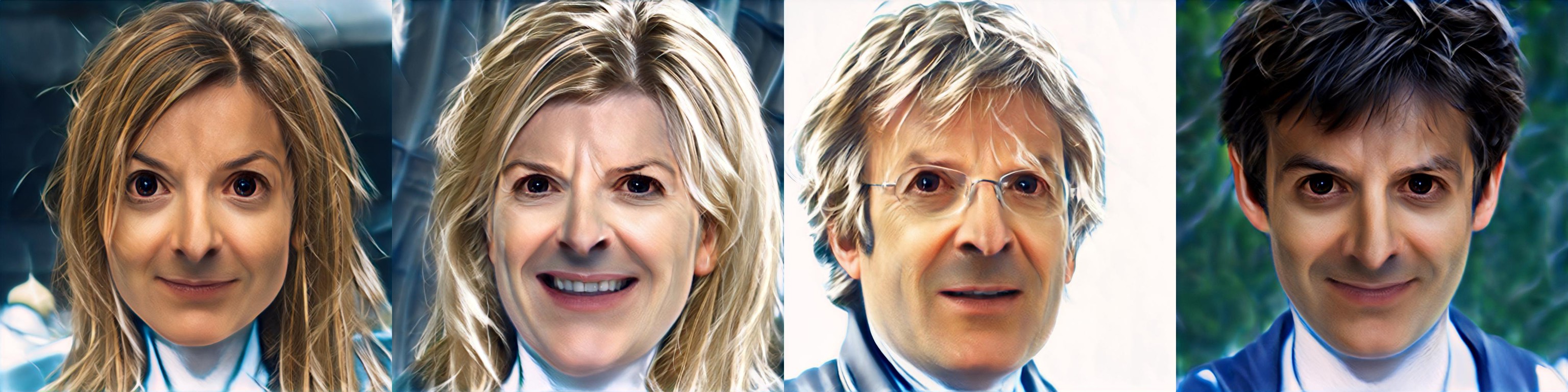} \\
        \raisebox{0.025\textwidth}{\rotatebox[origin=t]{90}{\begin{tabular}{c@{}c@{}}Person $\rightarrow$ \\ Nurse\end{tabular}}} &
        \includegraphics[width=0.42\textwidth]{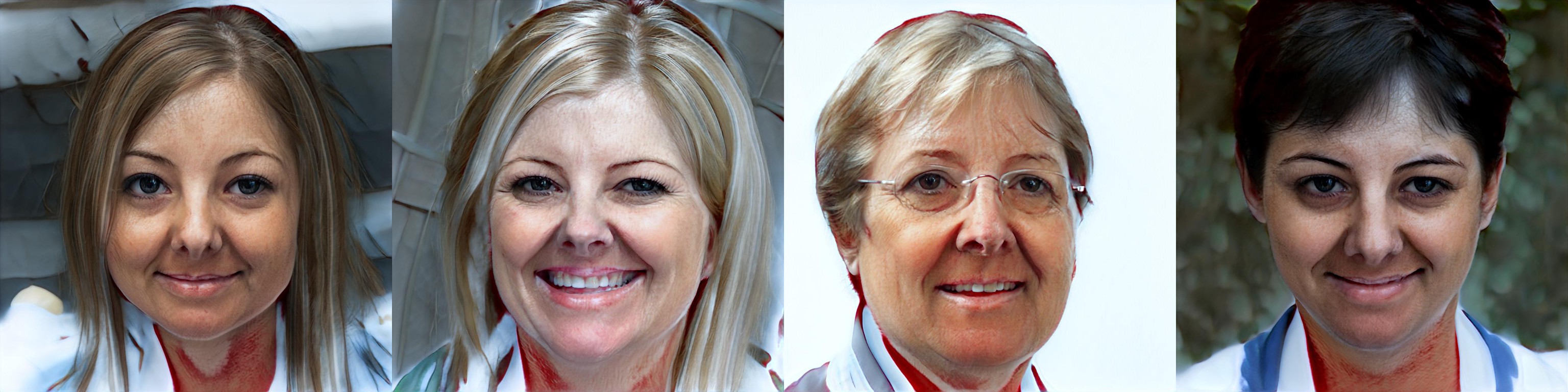} \vspace{-2pt} \\
        & Text \\
        \raisebox{0.04\textwidth}{\rotatebox[origin=t]{90}{Doctor}} &
        \includegraphics[width=0.42\textwidth]{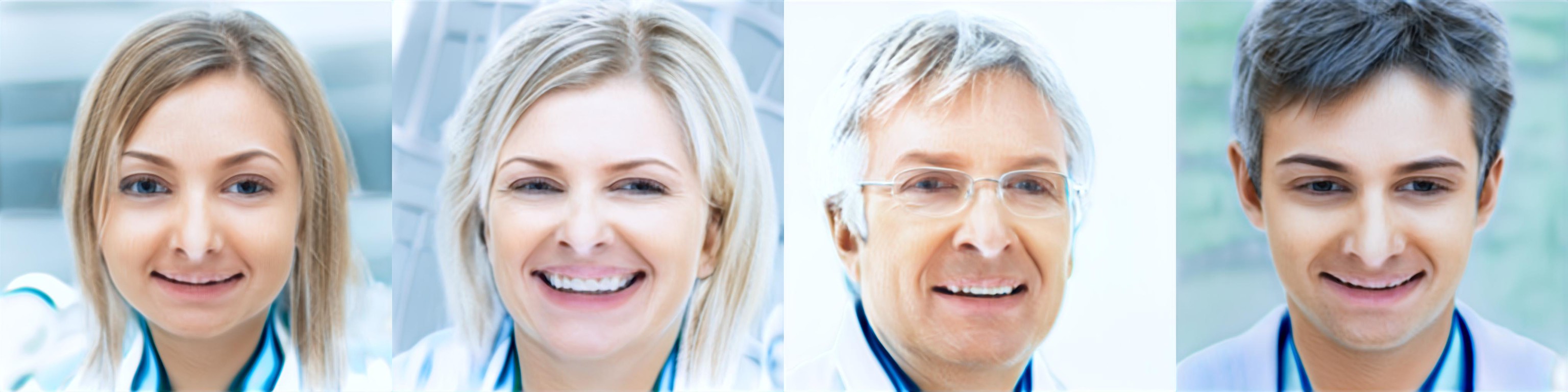} \\
        \raisebox{0.04\textwidth}{\rotatebox[origin=t]{90}{Nurse}} &
        \includegraphics[width=0.42\textwidth]{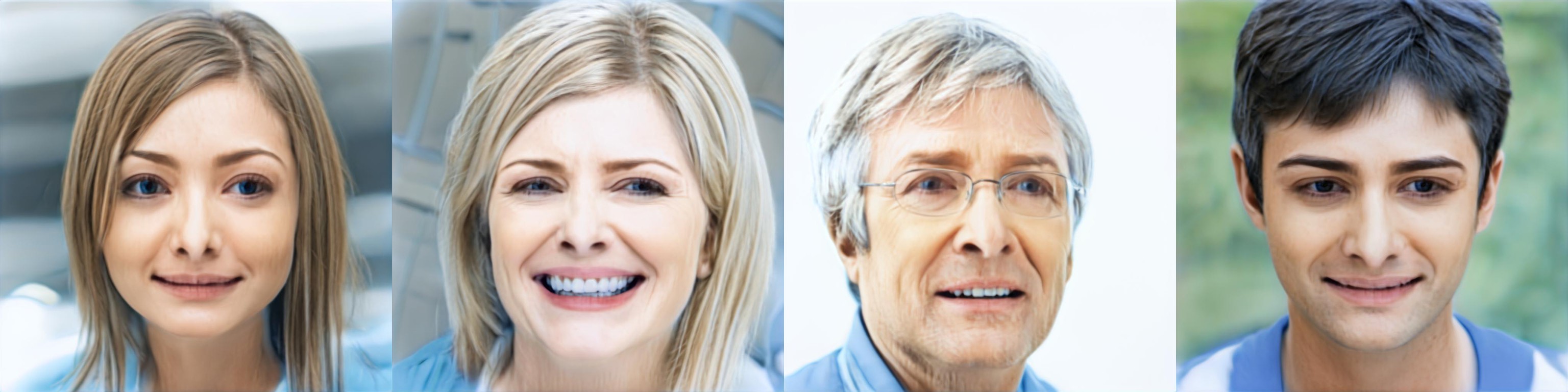} \vspace{-2pt} \\
        & Images \vspace{-3pt} \\ 
    \end{tabular}
    }
    \caption{Textual bias and ambiguity in domain adaptation. We transform an FFHQ \cite{karras2019style} model into doctors and nurses. When the conversion is done using text, CLIP's learned biases manifest in the new domain. In particular, the text `doctor' converts the individuals to males, while the text `nurse' converts them to females. Furthermore, the images generated when using the text `doctor' resemble the actor David Tennant from the television show `Doctor Who'.
    When using image targets, both issues dissapear. However, the model may instead pick up biases found in the few-shot set, such as facial expressions or hair color.}
    \label{fig:few_shot_medical}
    \vspace{-1pt}
\end{figure}

In comparison to our proposed zero-shot method, the few-shot approach can help alleviate some of the limitations of the model. In particular, it offers a way to avoid linguistic ambiguity by presenting an example of the specific domain we wish to mimic. Moreover, it can be used to better target specific styles which may be difficult to describe through text. However, as CLIP's embedding space is semantic in nature, this process is not guaranteed to converge to the exact realization of the style.
Finally, the few-shot approach can be used to combat the biases learned by CLIP. For example, by providing images of medical professionals from both sexes, one may avoid CLIP's preference for male doctors or female nurses. See \cref{fig:few_shot_medical} for an example.

\paragraph{Zero-shot pre-training} In our second approach, we consider the scenario where a few dozen or hundred images are available.
In such a scenario, few-shot models can already achieve reasonable performance. However, their performance typically depends on the similarity between the source generator and their desired target domain \cite{ojha2021few}. We show that their performance can therefore be improved (and in some situations, considerably so) if, rather than fine-tuning a model from some semi-related source domain, we first use our zero-shot approach to decrease the gap between the two domains.

Let $F$ denote a few-shot method that converts a generator between domains using an image set $\{I_{real}\}$, and let $N$ denote our zero-shot method, then we propose a few-shot adaptation of the form:

\begin{table}\setlength{\tabcolsep}{2pt}
\small
\caption{FID $\left(\downarrow\right)$ of alternative methods for applying StyleGAN-NADA as a pre-training approach for improving few-shot models. We show results for conversion of cat models from an AFHQ-Dog\cite{Karras2020ada} generator using 10 images. For each model we show the baseline results achieved without pre-training, and the results achieved when: pre-training only the generator (`Pretrained G'), fine-tuning both the generator and the discriminator on StyleGAN-NADA generated images (`Finetuned G + D'), and when allowing the first 50 steps of the few-shot training to modify only the discriminator (`catch-up'). }\label{tab:pre_training_options}

\centering 
\begin{tabular}{lcccc}
     Method & Ojha \etal & MineGAN & TGAN & TGAN + ADA \\ \hline
     Baseline        & 45.13 & 79.31 & 87.11  & 52.70  \\
     Pre-trained G    & 45.61 & \bf{48.15} & 100.61 & 51.48 \\
     Fine-tuned G + D & 48.38 & 56.89 & 56.46  & 60.82 \\
     D `catch-up'    & \bf{41.22} & 51.81 & \bf{54.49}  & \bf{49.76} \\
\end{tabular}\vspace{-2pt}
\end{table}

\begin{table}[hbt]\setlength{\tabcolsep}{3pt}
\footnotesize
\caption{FID $\left(\downarrow\right)$ for selected few-shot models when fine-tuning a source generator with and without a pre-training step using our method. Cat models were converted from an AFHQ-Dog generator. The sketch model was converted from an official FFHQ 256x256 \href{https://nvlabs-fi-cdn.nvidia.com/stylegan2-ada/pretrained/paper-fig7c-training-set-sweeps/ffhq140k-paper256-noaug.pkl}{checkpoint}. For every model and training set we compare the pre-trained and direct fine-tuning results and mark the winner in bold. For each set we further highlight the best performing model in blue.}\label{tab:pre_training}

\centering 
\begin{tabular}{lcccc} 
     \multirow{2}{*}{Model} & \multirow{2}{*}{Pre-training?} & \multicolumn{3}{c}{Set} \\
     & & Cats (10) & Cats (100) & Sketches (10) \\ \hline
     \multirow{2}{*}{Ojha \etal} & $\times$ & 45.13 & 27.54 & 72.74 \\
      & \checkmark & \bf{\color{blue}{41.22}} & \bf{\color{blue}{16.17}} & \bf{69.92} \\ \hline
     \multirow{2}{*}{MineGAN} & $\times$ & 79.31 & 27.34 & 62.27\\
      & \checkmark & \bf{51.81} & \bf{21.33} & \bf{60.55} \\ \hline
     \multirow{2}{*}{TGAN} & $\times$ & 87.11 & \bf{19.14} & 69.44 \\
      & \checkmark & \bf{54.49} & 19.41 & \bf{\color{blue}{56.52}} \\ \hline
     \multirow{2}{*}{TGAN + ADA} & $\times$ & 52.70 & \bf{19.09} & \bf{56.76} \\
      & \checkmark & \bf{49.76} & 25.47 & 57.18
\end{tabular}\vspace{-10pt}
\end{table}

\begin{equation} 
\begin{gathered}
G_{\mathrm{NADA}} = N\left(G_S, t_{source}, t_{target}\right) \\
G_T = F\left(G_{\mathrm{NADA}}, \{I_{real}\}\right)~,
\end{gathered}
\end{equation}
where $G_S$ and $G_T$ are the source and target generators respectively, $G_{\mathrm{NADA}}$ is an intermediate generator, created by applying our zero-shot method to $G_S$ using the textual prompts $t_{source}$, and $t_{target}$.

Typical few-shot adaptation methods require a well-trained discriminator for the source domain. Our method, however, modifies only the generator. We therefore investigate three possible solutions to this generator-discriminator divergence: In the first, we simply ignore the divergence and employ the few-shot methods as usual, using $G_{\mathrm{NADA}}$ along with the source discriminator, $D_S$. In the second approach, we allow the discriminator to `catch-up' by first performing a few training iterations where only the discriminator is updated, using $G_{\mathrm{NADA}}$ as the source of fake data, and the few-shot set for samples of the target domain. In the final approach, we use $G_{\mathrm{NADA}}$ and an accompanying StyleCLIP mapper to synthesize a large collection of images and then fine-tune $G_S$ and $D_S$ using this large collection before using them as the sources for the few-shot method.

We first compare all three methods using the dog-to-cat 10-image setup. In all cases we use the same number of training iterations as the baseline model (including any iterations where only the discriminator is trained). For the `Fine-tuned G + D' configuration we generated $25$ thousand images and fine-tuned the original dog model on this synthetic set for 5000 iterations. FID metrics were calculated following Ojah \etal \cite{ojha2021few}, by sampling $5000$ images from the fine-tuned model and comparing with the full (non few-shot) target set. The results are shown in \cref{tab:pre_training_options}. These results indicate that, of the three alternatives, starting with a brief discriminator-only training session yields the most consistent improvements. We hypothesize that the initial synchronization of the discriminator helps focus it on features that differentiate real cats from fake ones, rather than those that differentiate cats from dogs. In the fine-tuning case, it is possible that training on images generated with an additional StyleCLIP-mapper step leads to a reduction in diversity that harms the downstream adaptation methods.
MineGAN portrays a particularly large improvement even when using just the pre-trained generator, likely because it manages to identify the latent-regions where the good cats reside and focus the network's attention on them.

Having determined that a discriminator `catch-up' session produces the most consistent improvements, we turn to evaluating our pre-training method on additional domains and levels of supervision. 
The results are shown in \cref{tab:pre_training}. In all experiments we show the results of the `catch-up' method, even where one of the alternatives achieves superior results. In almost all cases, pre-training using our zero-shot method leads to lower FIDs. In some cases, pre-training using StyleGAN-NADA leads to remarkable improvements of more than 40\% in FID scores. These results indicate that our method can aid in reducing the domain gap before the application of the few-shot method, giving the later a much more convenient starting point.

\section{Beyond StyleGAN}
\begin{figure}[htb]\vspace{-0pt}
    \centering
    \setlength{\belowcaptionskip}{-2.5pt}
    \setlength{\tabcolsep}{1pt}
    {\scriptsize
    \begin{tabular}{c c c c}
    
        \includegraphics[width=0.24\linewidth]{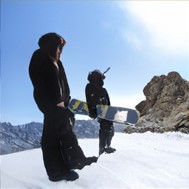} &
        \includegraphics[width=0.24\linewidth]{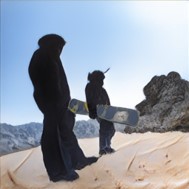} &
        \includegraphics[width=0.24\linewidth]{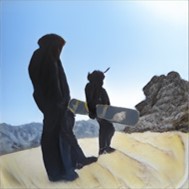} &
        \includegraphics[width=0.24\linewidth]{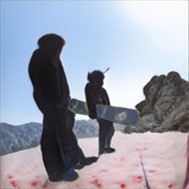} 
        \\
        
        Snow $\rightarrow$ & Chocolate ice cream & Vanilla ice cream & Cherry ice cream \\
        
        \includegraphics[width=0.24\linewidth]{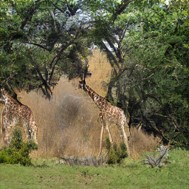} &
        \includegraphics[width=0.24\linewidth]{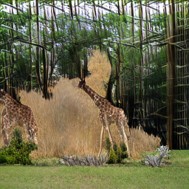} &
        \includegraphics[width=0.24\linewidth]{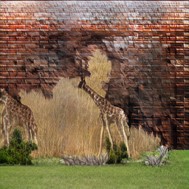} &
        \includegraphics[width=0.24\linewidth]{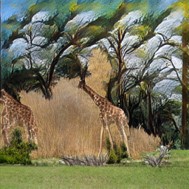}
        \\
        Tree $\rightarrow$ & Bamboo forest & Brick wall & Bob Ross painting \\
        
        \includegraphics[width=0.24\linewidth]{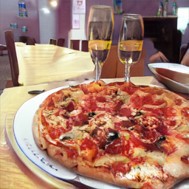} &
        \includegraphics[width=0.24\linewidth]{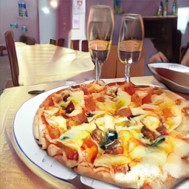} &
        \includegraphics[width=0.24\linewidth]{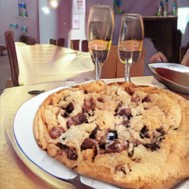} &
        \includegraphics[width=0.24\linewidth]{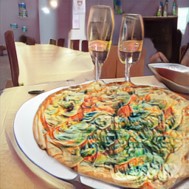} 
        \\
        Pizza $\rightarrow$ & Pizza with pineapple & {\begin{tabular}{c@{}c@{}}Chocolate chip \\ cookie\end{tabular}} & {\begin{tabular}{c@{}c@{}}Painting in the \\ style of Van Gogh\end{tabular}} \\

    \end{tabular}}
    \vspace{-5pt}
    \caption{Replacing COCO-Stuff\cite{caesar2018coco} classes with novel classes decribed through text. All images were generated using OASIS\cite{schonfeld2021oasis}. Images in the left column were created by the official pre-trained model. In each row we fine-tune the model so that one class changes its identity in the synthesized images. Target texts are provided below each modified image.}
    \label{fig:spade_edits}\vspace{-5pt}
\end{figure}

In addition to StyleGAN, we investigated our model's ability to convert existing classes in a more localized manner using OASIS~\cite{schonfeld2021oasis}, a SPADE-like~\cite{park2019SPADE} model that synthesizes images from segmentation masks. In this setup, we utilize a model pre-trained on the COCO-stuff dataset~\cite{caesar2018coco}, and aim to change one of the model classes to a novel class which shares the shape of the source class, but differs in texture. To this end, we employ the same training architecture and training losses presented in the core paper, with two modifications: First, before passing a generated image to CLIP, we mask all regions which do not belong to our designated class. In this manner, CLIP-space directions are calculated using only the regions with the class we wish to change. Second, we seek to minimize change throughout all regions outside the mask, as well as all images where the designated class does not appear. As such, we employ both an L2 and an LPIPS \cite{zhang2018perceptual} loss between all the masked regions in the source and target generator outputs. Qualitative results are shown in \cref{fig:spade_edits}.

These results demonstrate that our framework can be readily applied to other generative models with minimal effort. In this sense, our method is not a StyleGAN-specific tool, but rather a general framework for training generative models without data.

\section{Identity Preservation}\label{sec:identity}
\begin{figure*}[hbt]
\setlength{\tabcolsep}{1pt}
    \centering
    \includegraphics[width=0.99\textwidth]{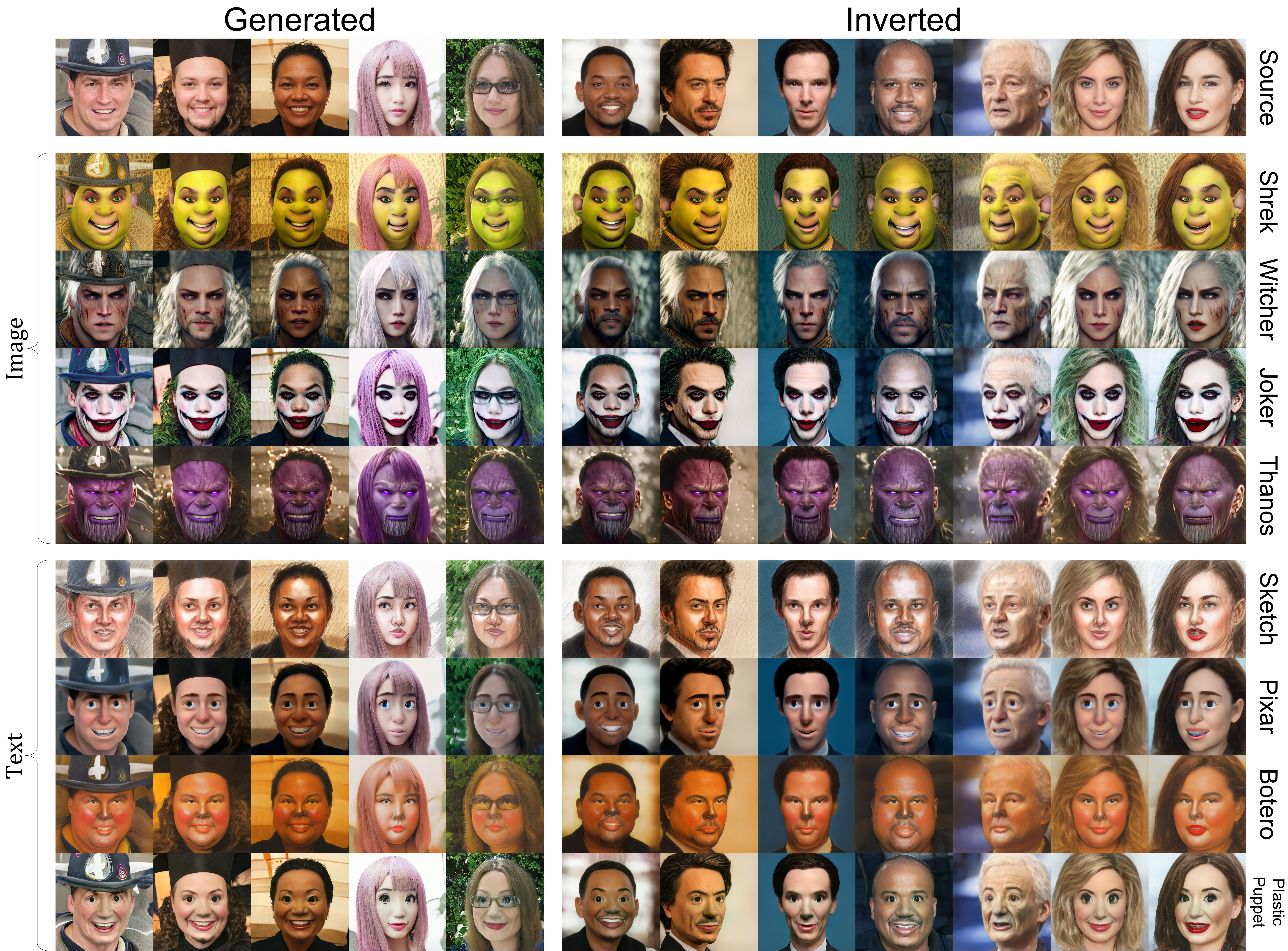}
    \caption{Cross-domain identity transfer for generated and real (inverted) images. Our method maintains a high degree of identity preservation, including accessories such as hats and eyeglasses. The method works equally well with both synthesized and real images. Furthermore, it can be employed using both textual and image targets.}
    \label{fig:edit_supp}
\end{figure*}

We provide additional examples that showcase our method's ability to preserve identity between different domains. \cref{fig:edit_supp} shows domain adaptation results using both synthetic and real images (inverted using e4e\cite{tov2021designing}). StyleGAN-NADA succesfully preserves the source identity in the new domain and even converts appropriate accessories such as hats and eyeglasses.

\section{Cross-Model interpolation}

\begin{figure*}[hbt]
\setlength{\tabcolsep}{1pt}
    \centering
    \includegraphics[width=0.98\textwidth]{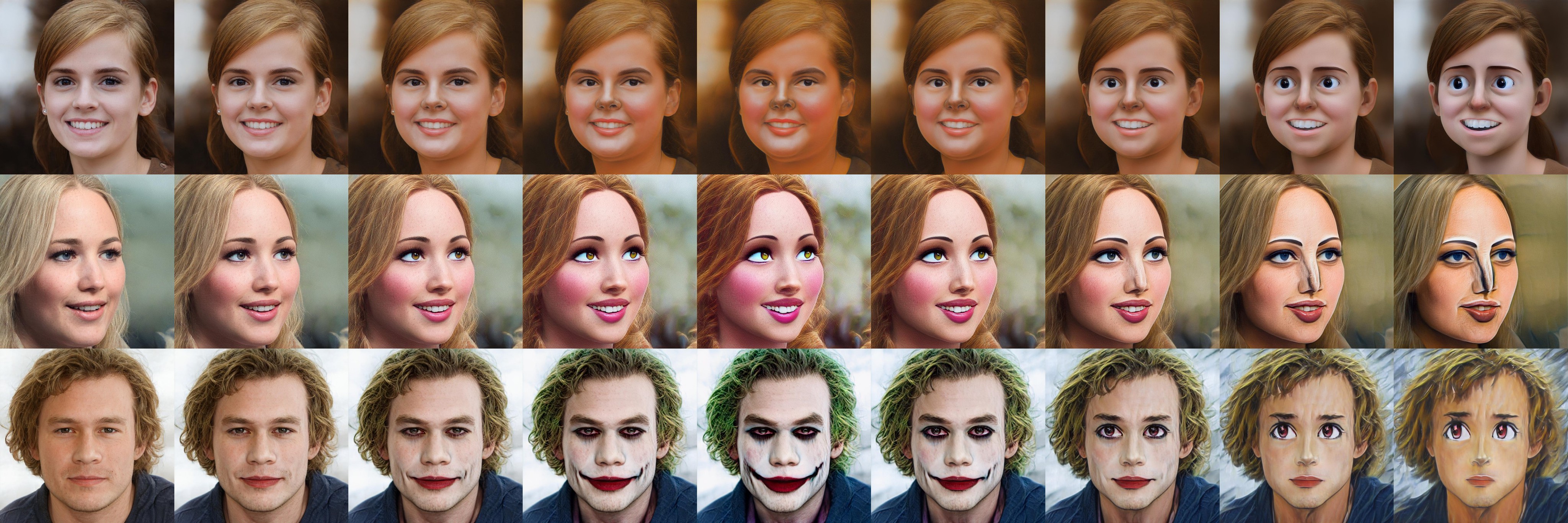}
    \caption{Cross-domain image interpolations. Our models can be used to smoothly transform an image between domains by interpolating the model weights, rather than latent codes.
    The left-most image in each row is a reconstruction in the original StyleGAN2-FFHQ~\cite{karras2020analyzing} model, obtained using e4e~\cite{tov2021designing}. The other images demonstrate model-based interpolation through two different domains. Interpolation works not only from the source domain to new targets, but also between different target domains. }
    \label{fig:model_interp}
    \vspace{-5pt}
\end{figure*}

In addition to supporting latent-space interpolations and editing in the new domains, we observe that our models allow for an additional form of interpolation - between the model parameters in two different domains. This serves as further demonstration for the strong coupling of the latent spaces across our transformed models. Furthermore, doing so enables additional applications such as generating videos of smooth transitions across a wide range of domains. \cref{fig:model_interp} demonstrates such transitions. Our project page includes a video with further demonstrations.

For more applications of network interpolations, see \cite{wang2018dni} or the concurrent work by Wu \etal \cite{wu2021stylealign}.

\section{Additional samples}\label{sec:more_samples}

\begin{figure*}[!hbt]
    \centering
    \setlength{\belowcaptionskip}{-5pt}
    \setlength{\tabcolsep}{1pt}
    {\footnotesize
    \begin{tabular}{c c}
    
        \raisebox{0.08\textwidth}{\rotatebox[origin=t]{90}{\begin{tabular}{c@{}c@{}}Photo $\rightarrow$ \\ Amedeo Modigliani painting\end{tabular}}} &
        \includegraphics[width=0.8\textwidth]{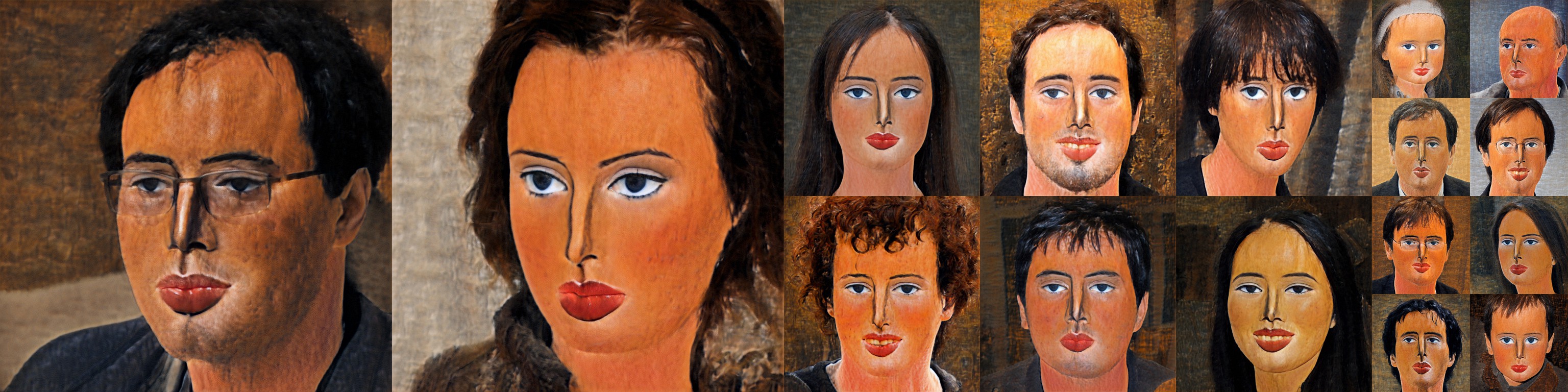} \\
        
        \raisebox{0.08\textwidth}{\rotatebox[origin=t]{90}{\begin{tabular}{c@{}c@{}}Human $\rightarrow$ \\ Tolkien elf\end{tabular}}} &
        \includegraphics[width=0.8\textwidth]{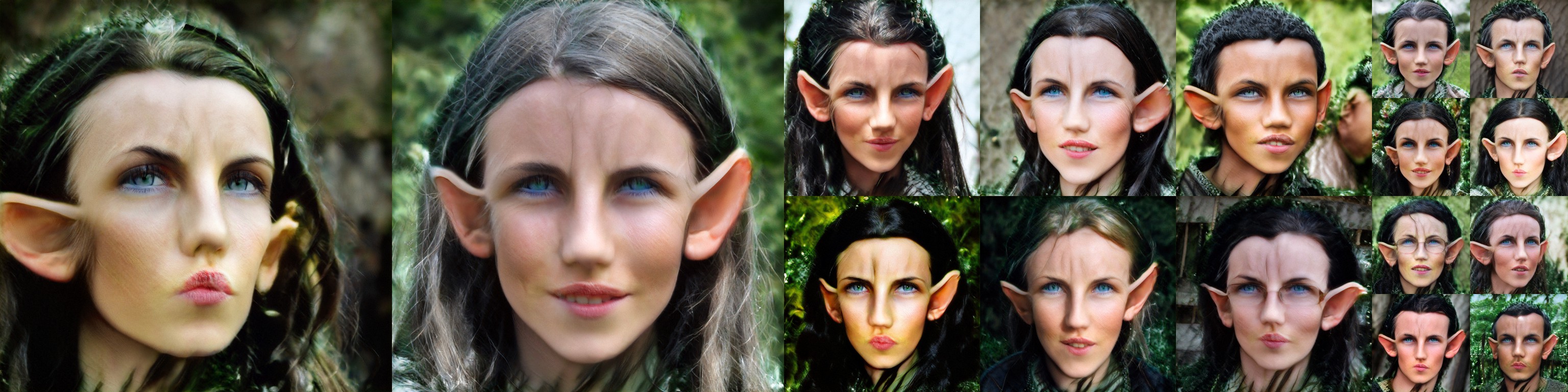} \\
        
        \raisebox{0.08\textwidth}{\rotatebox[origin=t]{90}{\begin{tabular}{c@{}c@{}}Human $\rightarrow$ \\ Zombie\end{tabular}}} &
        \includegraphics[width=0.8\textwidth]{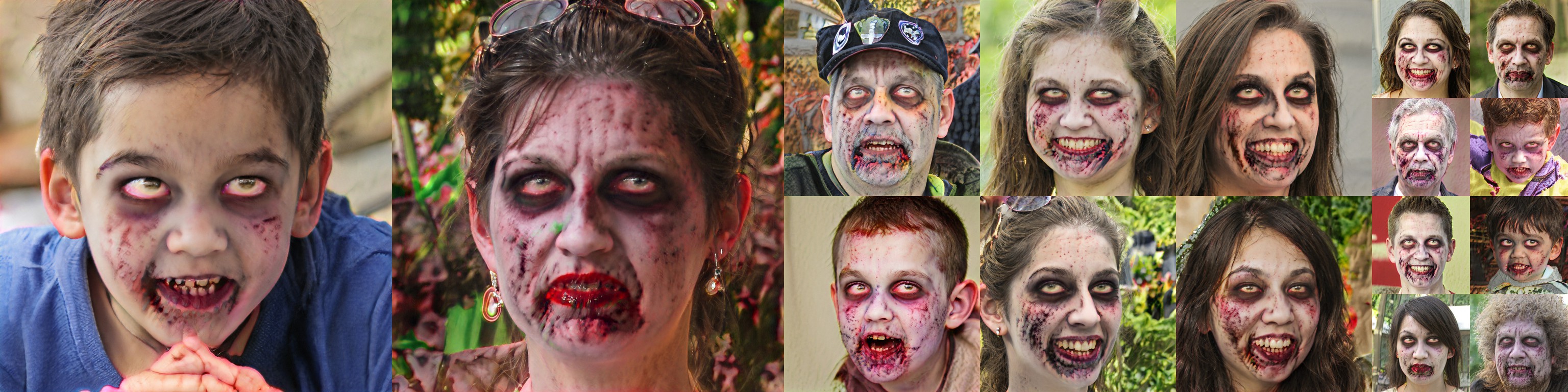} \\
        
        \raisebox{0.08\textwidth}{\rotatebox[origin=t]{90}{\begin{tabular}{c@{}c@{}}Human $\rightarrow$ \\ Neanderthal\end{tabular}}} &
        \includegraphics[width=0.8\textwidth]{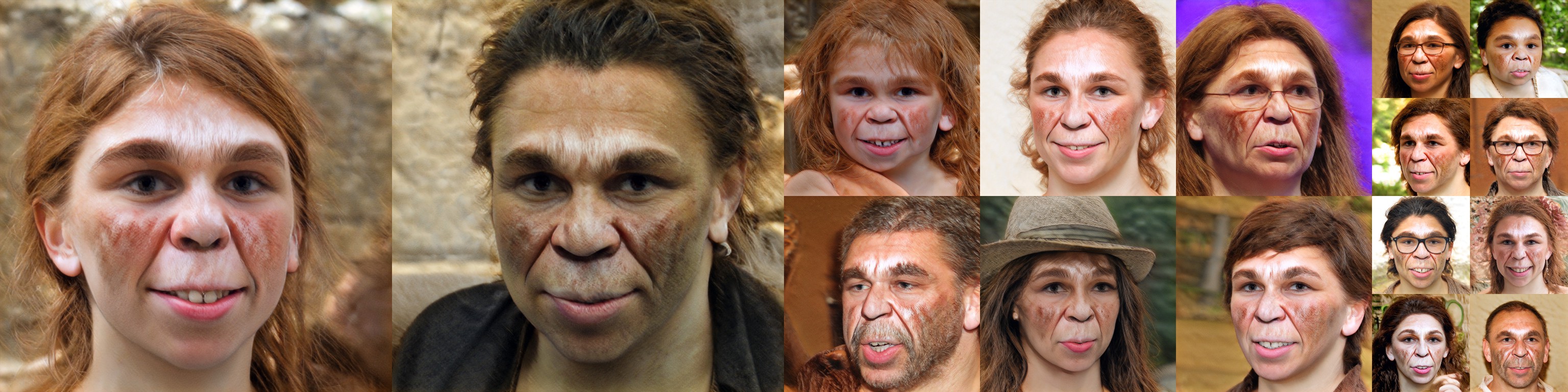} \\
        
        \raisebox{0.08\textwidth}{\rotatebox[origin=t]{90}{\begin{tabular}{c@{}c@{}}Human $\rightarrow$ \\ Mark Zuckerberg\end{tabular}}} &
        \includegraphics[width=0.8\textwidth]{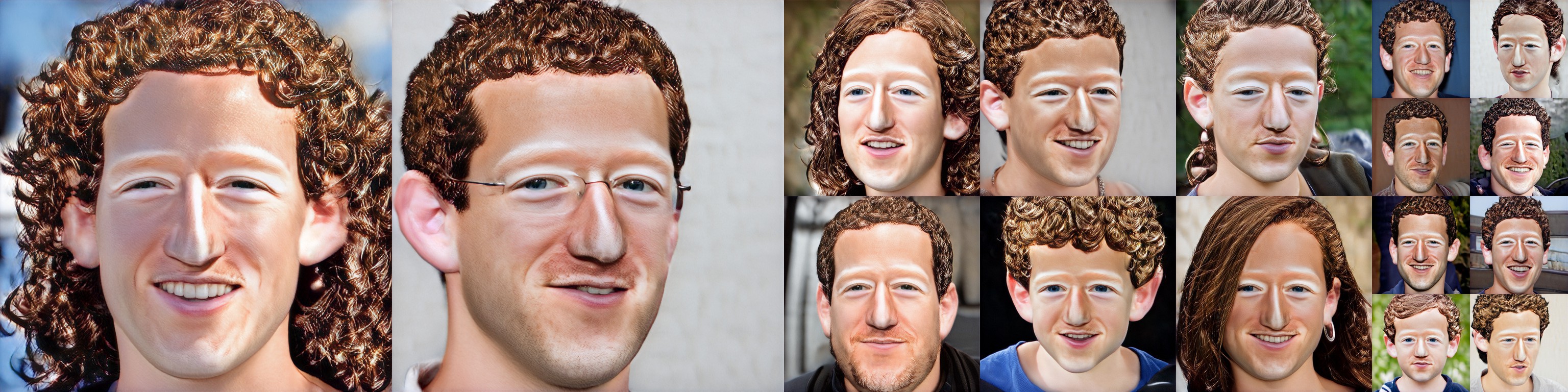} \\

    \end{tabular}
    }
    \vspace{0.1cm}
    \caption{Additional images synthesized using models adapted from StyleGAN2-FFHQ \cite{karras2020analyzing} to a set of textually-prescriped target domains. All images were sampled randomly, using truncation with $\psi = 0.7$. The driving texts appear to the left of each row.}
    \label{fig:generated_ffhq_2_appendix}
\end{figure*}

\begin{figure*}[!hbt]
    \centering
    \setlength{\belowcaptionskip}{-5pt}
    \setlength{\tabcolsep}{1pt}
    {\footnotesize
    \begin{tabular}{c c}
        
        \raisebox{0.08\textwidth}{\rotatebox[origin=t]{90}{\begin{tabular}{c@{}c@{}}Photo $\rightarrow$ \\ Mona Lisa Painting\end{tabular}}} &
        \includegraphics[width=0.8\textwidth]{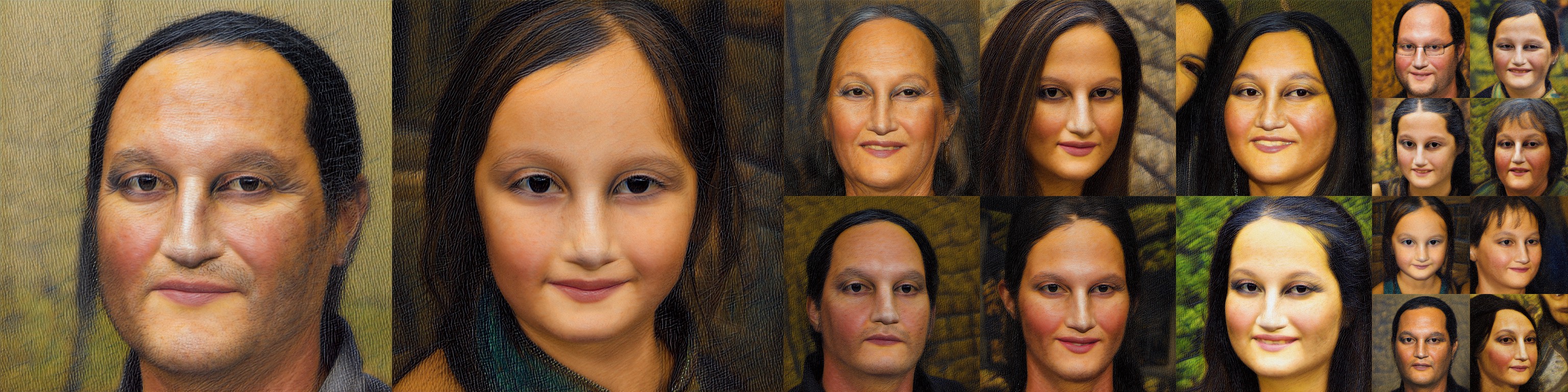} \\
        
        \raisebox{0.08\textwidth}{\rotatebox[origin=t]{90}{\begin{tabular}{c@{}c@{}}Photo $\rightarrow$ 3D render in \\ the style of Pixar\end{tabular}}} &
        \includegraphics[width=0.8\textwidth]{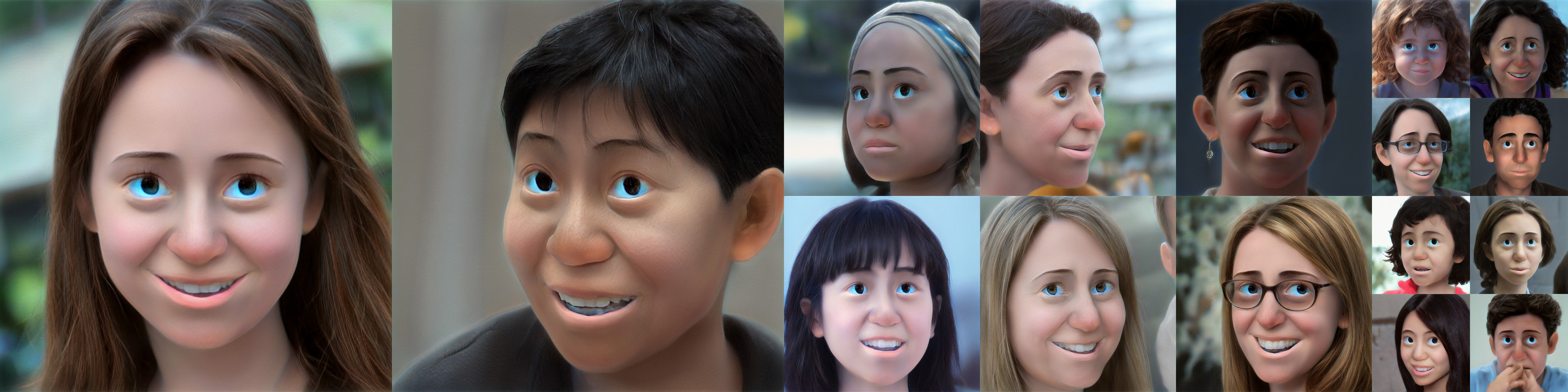} \\
        
        \raisebox{0.08\textwidth}{\rotatebox[origin=t]{90}{\begin{tabular}{c@{}c@{}}Photo $\rightarrow$ \\ A painting by Raphael \end{tabular}}} &
        \includegraphics[width=0.8\textwidth]{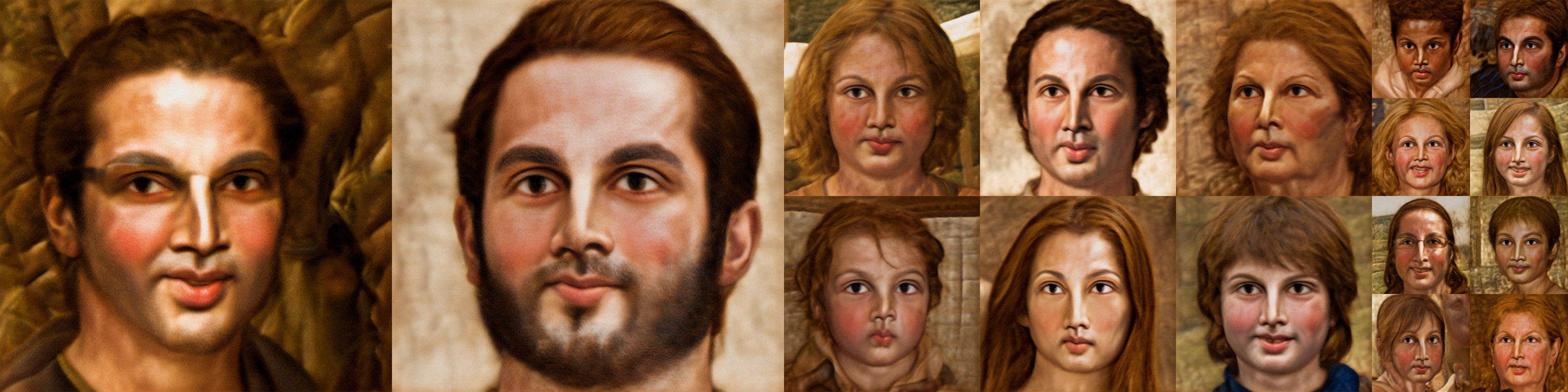} \\
        
        \raisebox{0.08\textwidth}{\rotatebox[origin=t]{90}{\begin{tabular}{c@{}c@{}}Photo $\rightarrow$ \\ Old-timey photograph\end{tabular}}} &
        \includegraphics[width=0.8\textwidth]{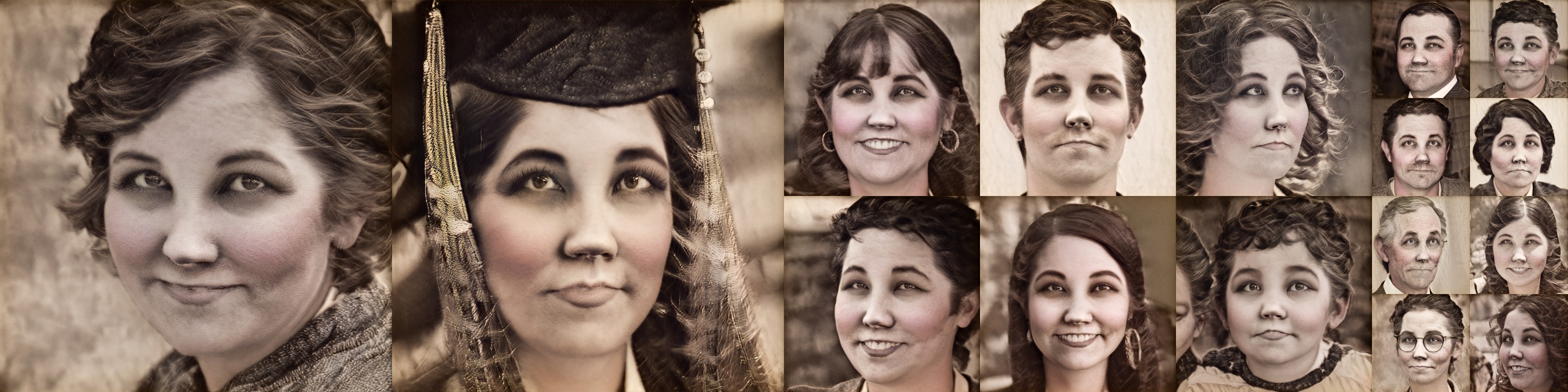} \\
        
        \raisebox{0.08\textwidth}{\rotatebox[origin=t]{90}{\begin{tabular}{c@{}c@{}}Photo $\rightarrow$ A painting \\ in Ukiyo-e style\end{tabular}}} &
        \includegraphics[width=0.8\textwidth]{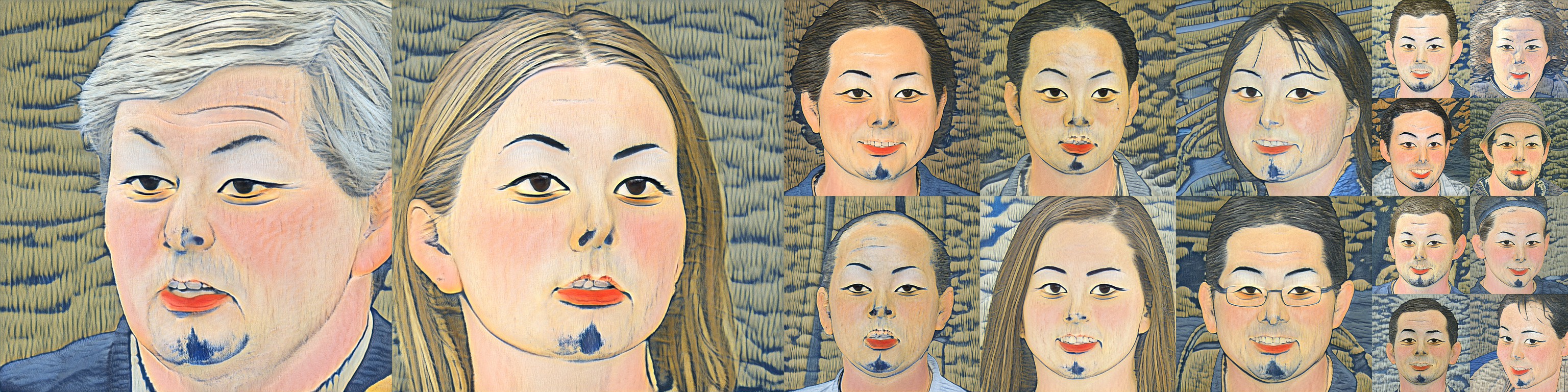} \\

    \end{tabular}
    }
    \vspace{0.1cm}
    \caption{Additional images synthesized using models adapted from StyleGAN2-FFHQ \cite{karras2020analyzing} to a set of textually-prescriped target domains. All images were sampled randomly, using truncation with $\psi = 0.7$. The driving texts appear to the left of each row.}
    \label{fig:generated_ffhq_appendix}
\end{figure*}

\begin{figure*}[!hbt]
    \centering
    \setlength{\belowcaptionskip}{-5pt}
    \setlength{\tabcolsep}{1pt}
    {\footnotesize
    \begin{tabular}{c c}
        
        \raisebox{0.08\textwidth}{\rotatebox[origin=t]{90}{Church $\rightarrow$ Hut}} &
        \includegraphics[width=0.77\textwidth]{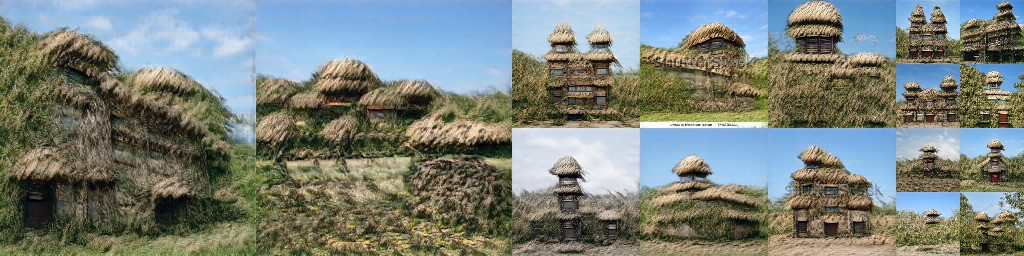} \\
        
        \raisebox{0.08\textwidth}{\rotatebox[origin=t]{90}{\begin{tabular}{c@{}c@{}}Church $\rightarrow$ \\ Snowy Mountain\end{tabular}}} &
        \includegraphics[width=0.77\textwidth]{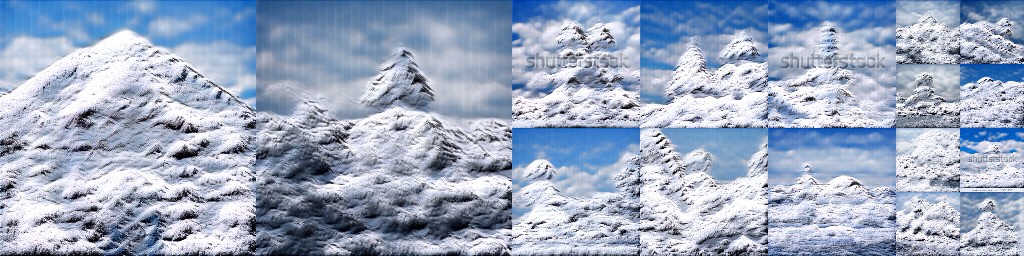} \\
        
        \raisebox{0.08\textwidth}{\rotatebox[origin=t]{90}{\begin{tabular}{c@{}c@{}}Church $\rightarrow$ \\ Ancient underwater ruin\end{tabular}}} &
        \includegraphics[width=0.77\textwidth]{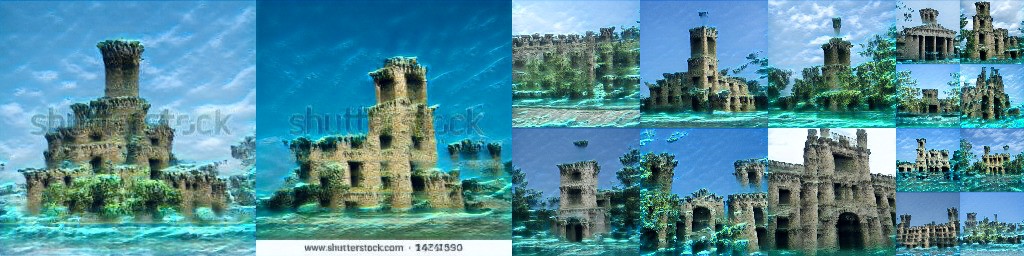} \\
        
        \raisebox{0.08\textwidth}{\rotatebox[origin=t]{90}{Church $\rightarrow$ The Shire }} &
        \includegraphics[width=0.77\textwidth]{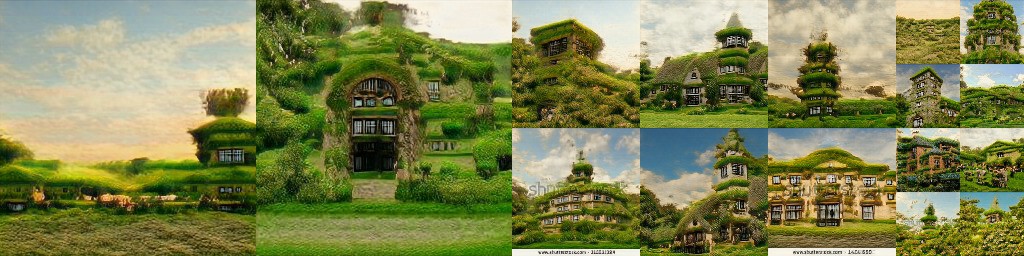} \\
        
        \raisebox{0.08\textwidth}{\rotatebox[origin=t]{90}{\begin{tabular}{c@{}c@{}}Photo of a church $\rightarrow$ Cryengine \\ render of Shibuya at night\end{tabular}}} &
        \includegraphics[width=0.77\textwidth]{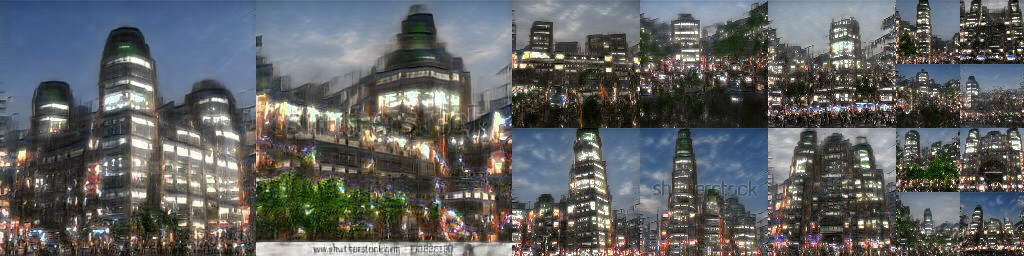} \\
        
        \raisebox{0.08\textwidth}{\rotatebox[origin=t]{90}{\begin{tabular}{c@{}c@{}}Photo of a church $\rightarrow$ Cryengine \\ render of New York\end{tabular}}} &
        \includegraphics[width=0.77\textwidth]{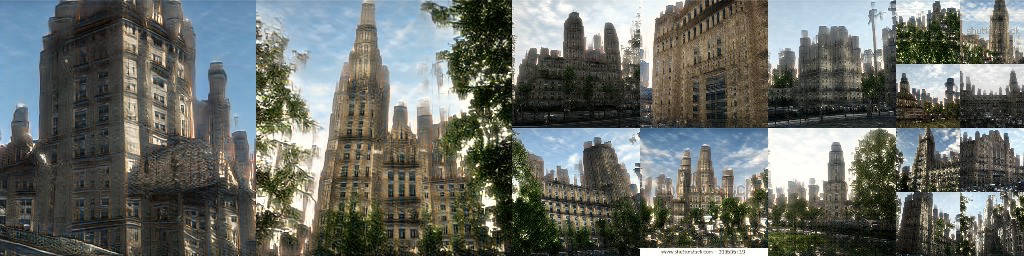} \\

    \end{tabular}
    }
    \vspace{0.1cm}
    \caption{Additional images synthesized using models adapted from StyleGAN2 \cite{karras2020analyzing} LSUN Church \cite{yu2015lsun} to a set of textually-prescriped target domains. All images were sampled randomly, using truncation with $\psi = 0.7$. The driving texts appear to the left of each row.}
    \label{fig:generated_church_appendix}
\end{figure*}

\begin{figure*}[!hbt]
    \centering
    \setlength{\belowcaptionskip}{-5pt}
    \setlength{\tabcolsep}{1pt}
    {\footnotesize
    \begin{tabular}{c c}

        \raisebox{0.08\textwidth}{\rotatebox[origin=t]{90}{\begin{tabular}{c@{}c@{}}Photo $\rightarrow$ \\ Pixel Art\end{tabular}}} &
        \includegraphics[width=0.78\textwidth]{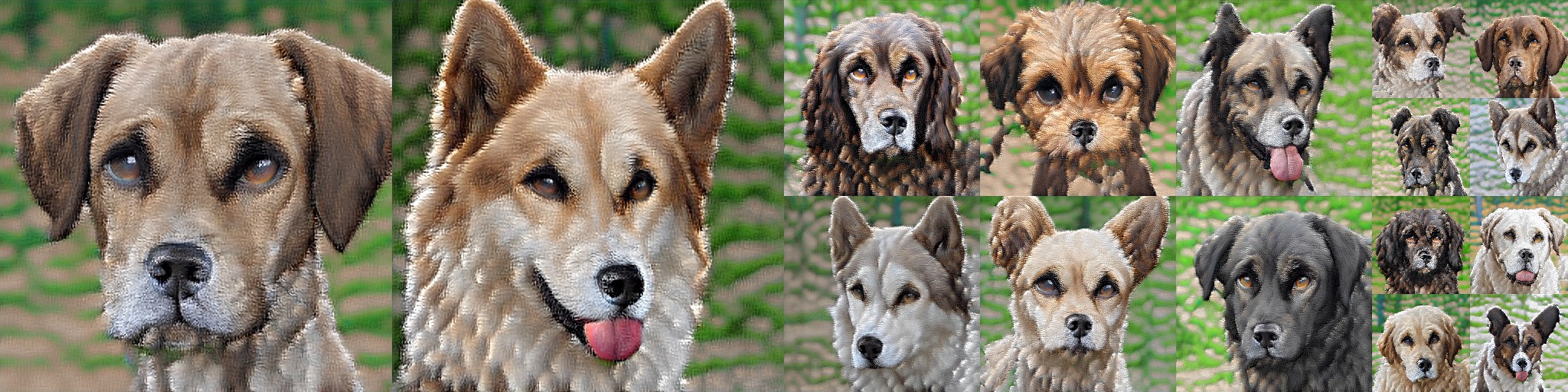} \\
        
        \raisebox{0.08\textwidth}{\rotatebox[origin=t]{90}{\begin{tabular}{c@{}c@{}}Dog $\rightarrow$ \\ The Joker\end{tabular}}} &
        \includegraphics[width=0.78\textwidth]{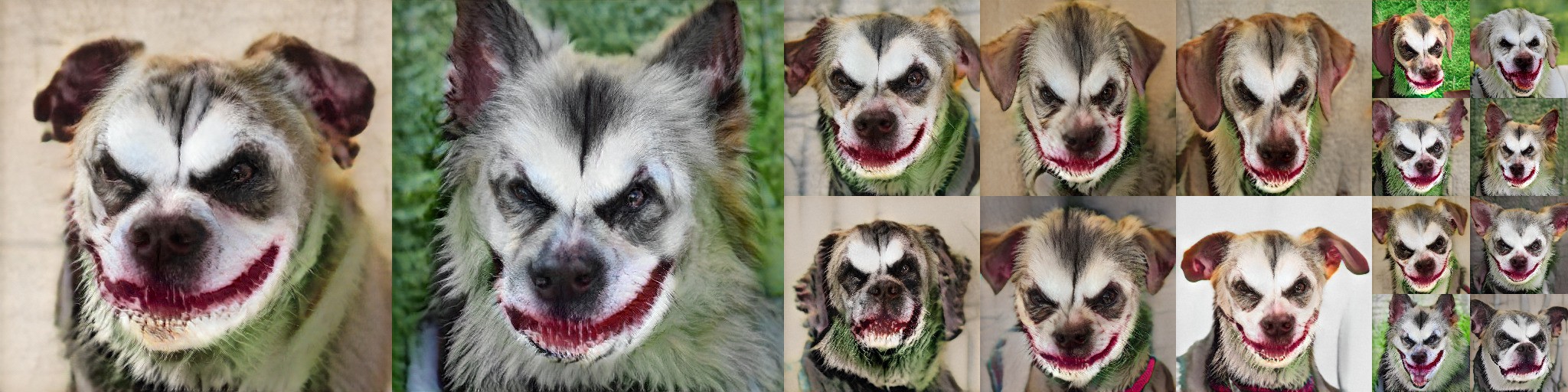} \\
        
        \raisebox{0.08\textwidth}{\rotatebox[origin=t]{90}{\begin{tabular}{c@{}c@{}}Dog $\rightarrow$ \\ Bugs Bunny\end{tabular}}} &
        \includegraphics[width=0.78\textwidth]{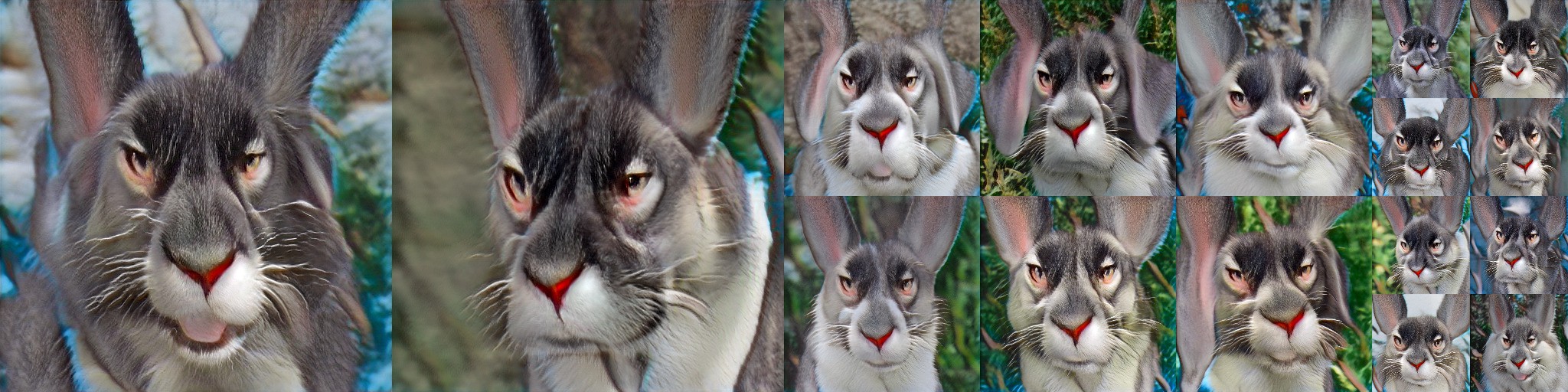} \\

        \raisebox{0.08\textwidth}{\rotatebox[origin=t]{90}{\begin{tabular}{c@{}c@{}}Dog $\rightarrow$ \\ Nicolas Cage \end{tabular}}} &
        \includegraphics[width=0.78\textwidth]{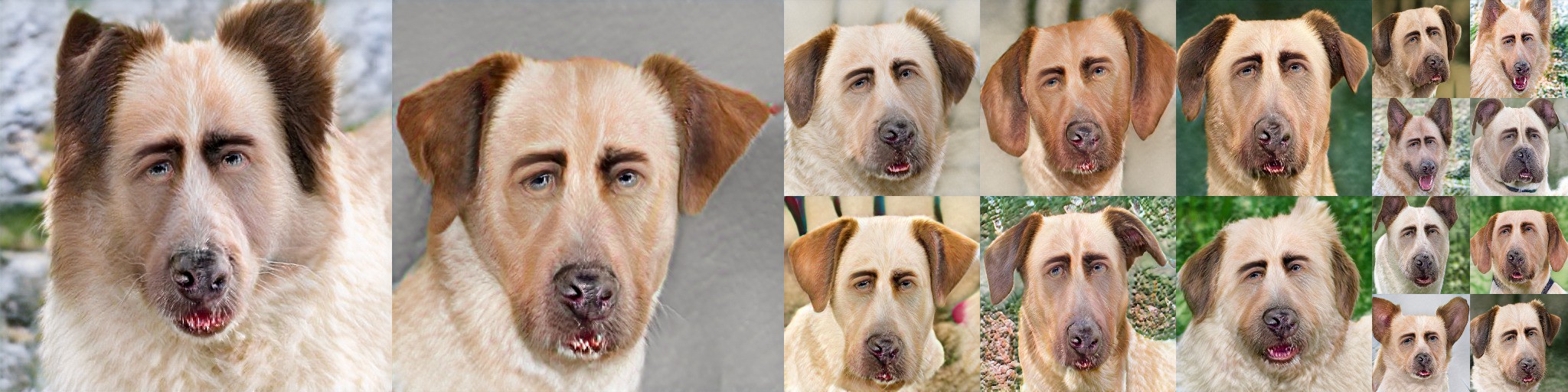} \\
        
        \raisebox{0.08\textwidth}{\rotatebox[origin=t]{90}{\begin{tabular}{c@{}c@{}}Photo $\rightarrow$ \\ Watercolor Art \end{tabular}}} &
        \includegraphics[width=0.78\textwidth]{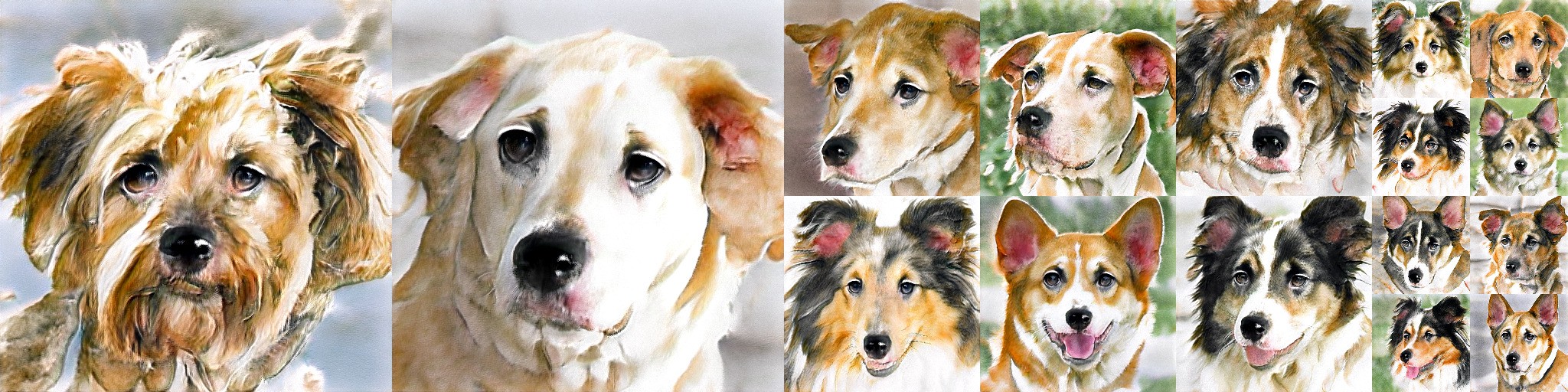} \\
        
        \raisebox{0.08\textwidth}{\rotatebox[origin=t]{90}{\begin{tabular}{c@{}c@{}}Photo $\rightarrow$ Watercolor Art \\  with Thick Brushstrokes\end{tabular}}} &
        \includegraphics[width=0.78\textwidth]{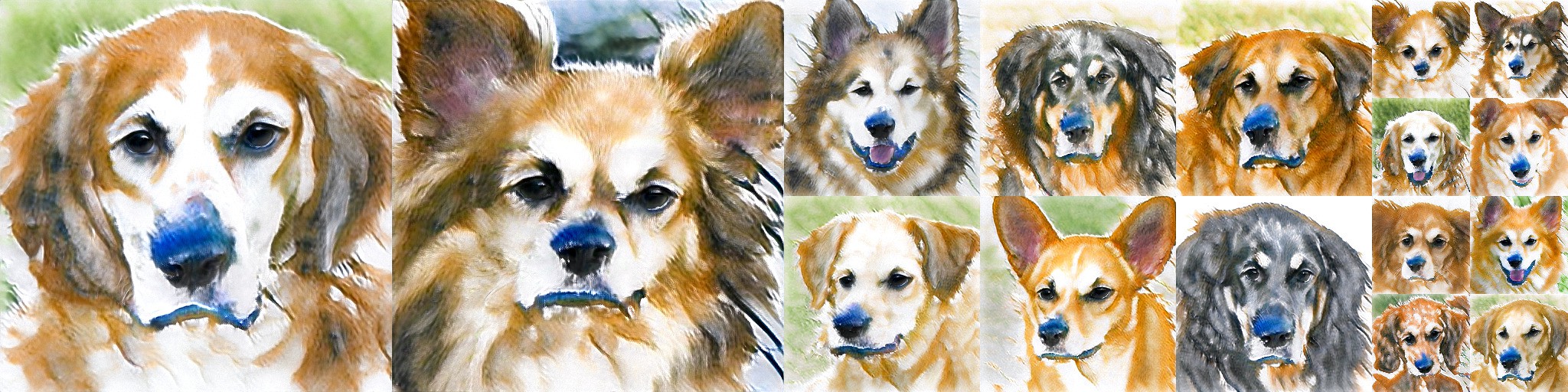} \\

    \end{tabular}
    }
    \vspace{0.1cm}
    \caption{Additional images synthesized using models adapted from StyleGAN-ADA \cite{Karras2020ada} AFHQ-dog \cite{choi2020starganv2} to a set of textually-prescriped target domains. All images were sampled randomly, using truncation with $\psi = 0.7$. The driving texts appear to the left of each row.}
    \label{fig:generated_dog_appendix}
\end{figure*}

\begin{figure*}[!hbt]
    \centering
    \setlength{\belowcaptionskip}{-5pt}
    \setlength{\tabcolsep}{1pt}
    {\footnotesize
    \begin{tabular}{c c c}
        
        \includegraphics[width=0.33\textwidth]{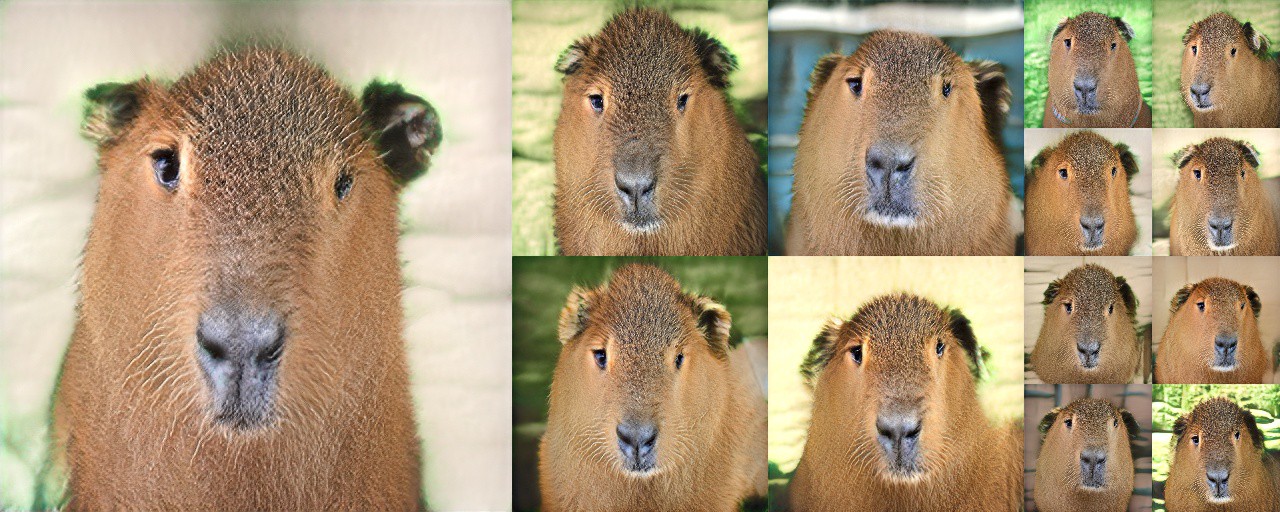} &
        \includegraphics[width=0.33\textwidth]{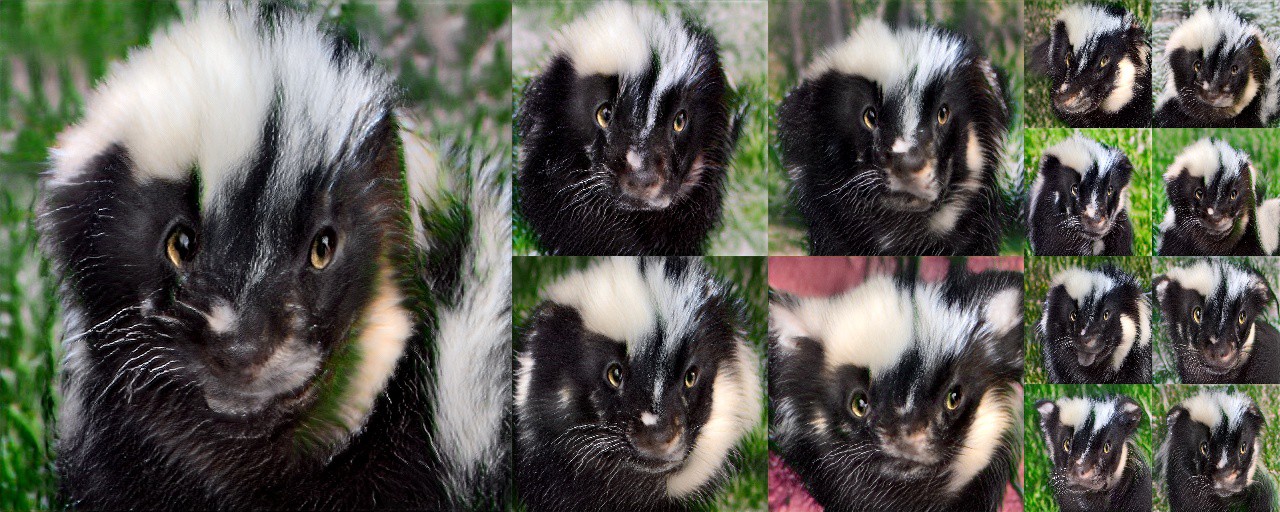} &
        \includegraphics[width=0.33\textwidth]{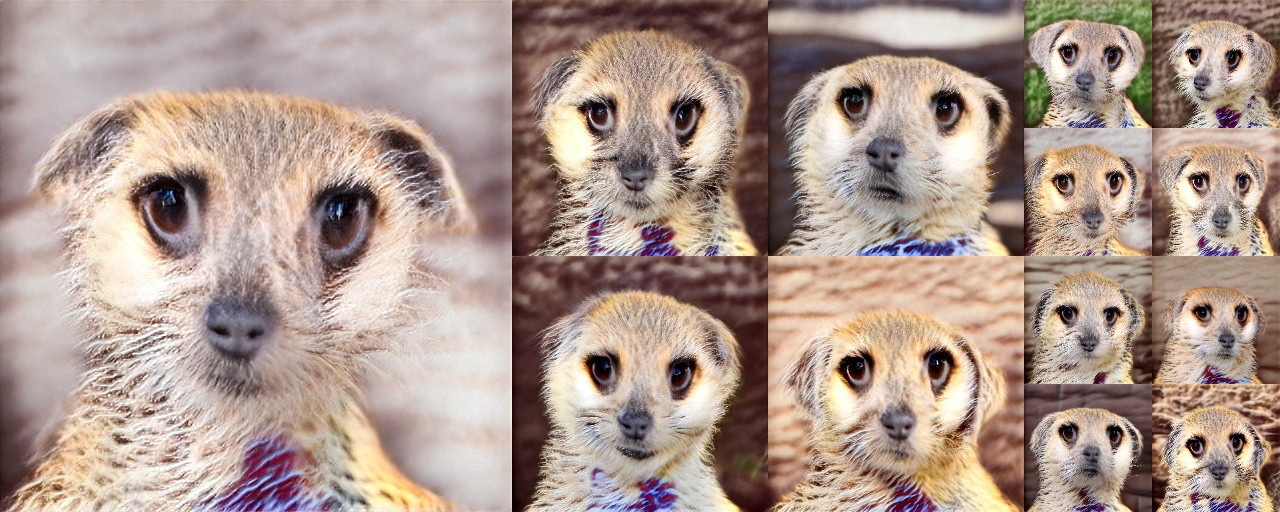} \\
        "Capybara" & "Skunk" & "Meerkat" \\
        
        \includegraphics[width=0.33\textwidth]{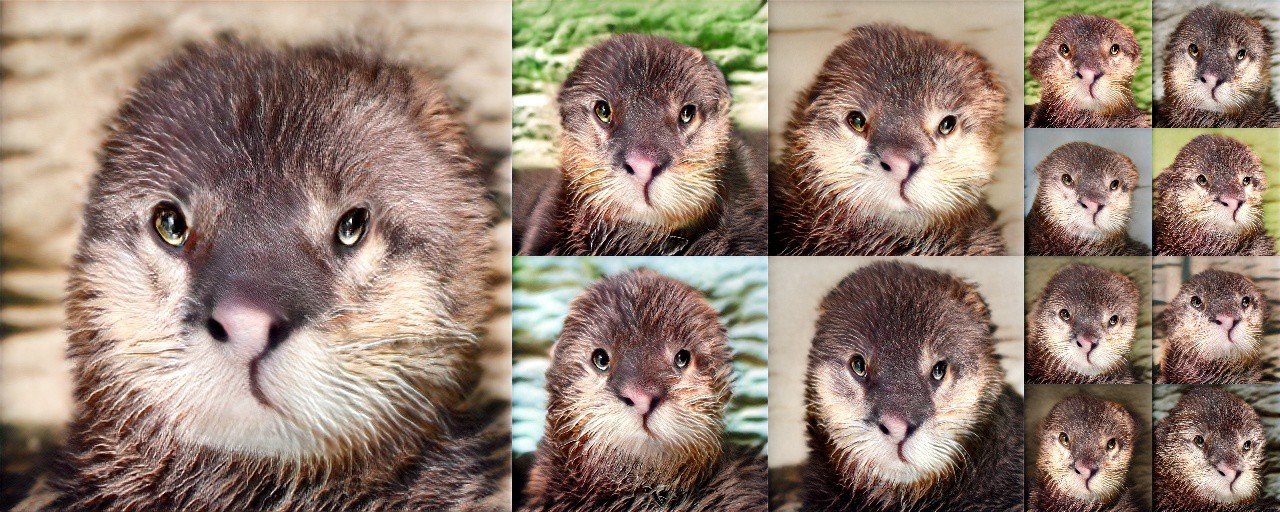} &
        \includegraphics[width=0.33\textwidth]{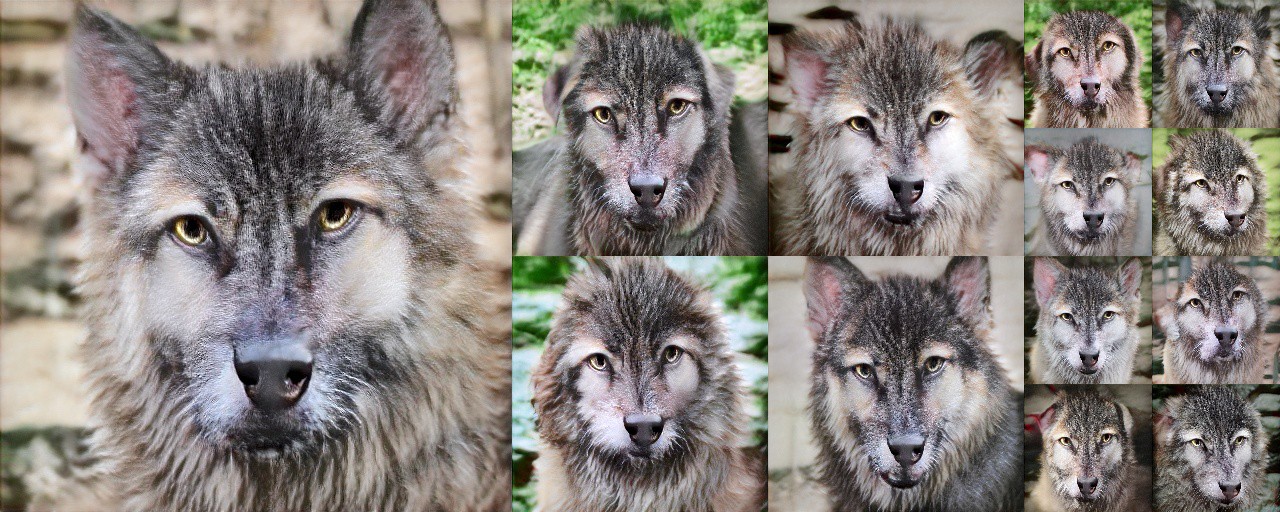} & 
        \includegraphics[width=0.33\textwidth]{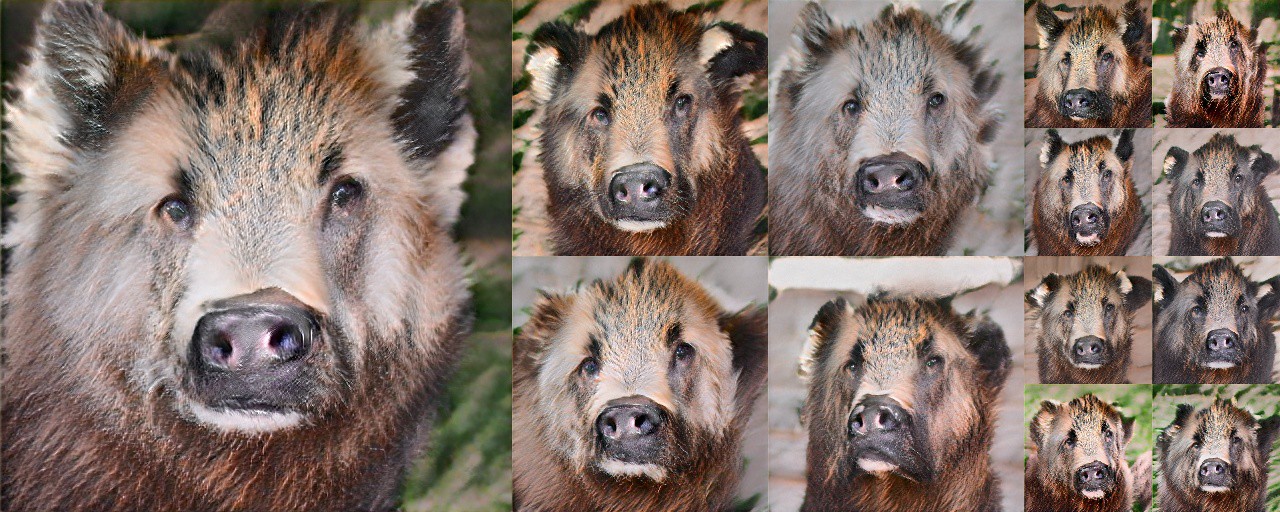} \\
        "Otter" & "Wolf" & "Boar" \\
        
    \end{tabular}
    }
    \vspace{-6pt}
    \caption{Additional generator transformations to multiple animal domains. In all cases we begin with a StyleGAN-ADA~\cite{Karras2020ada} AFHQ-Dog~\cite{choi2020starganv2} model. All generators are adapted using our method and a StyleCLIP \cite{patashnik2021styleclip} latent mapper. For all experiments, the source domain text was `Dog'. The target domain text is shown below each image.}
    \label{fig:animals_supp}\vspace{-5pt}
\end{figure*}

We provide additional synthesized results from a large collection of source and target domains. In \cref{fig:generated_ffhq_appendix,fig:generated_ffhq_2_appendix} we show results from models converted from the face domain. In \cref{fig:generated_church_appendix} we show results from models converted from the church domain. In \cref{fig:generated_dog_appendix} we show additional results from the dog domain. Finally, in \cref{fig:animals_supp} we show additional animal transformations.


\section{Qualitative few-shot comparisons}\label{sec:few_shot_qual}

\begin{figure*}[hbt]
\setlength{\tabcolsep}{1pt}
    \centering
    \includegraphics[width=0.95\textwidth]{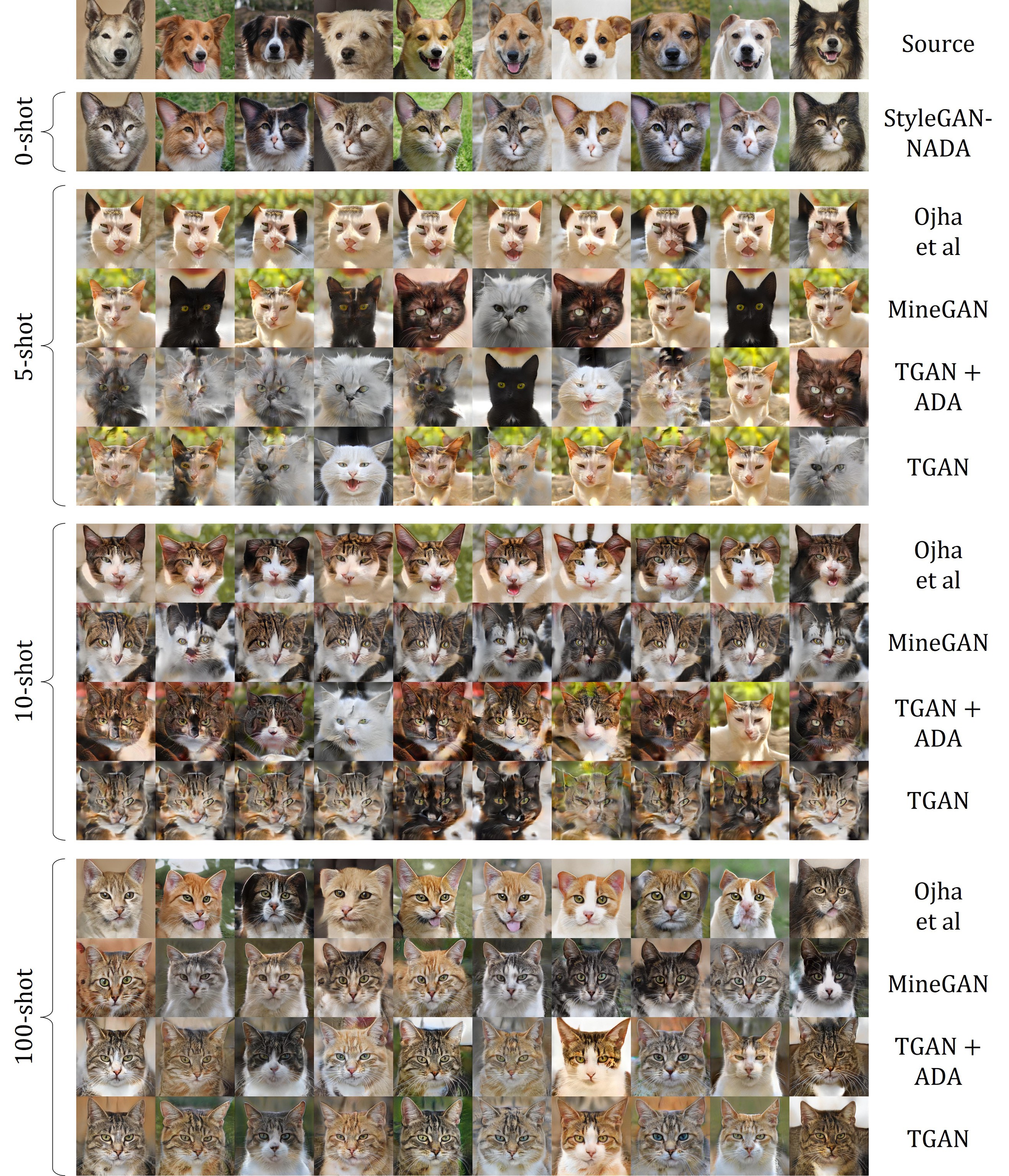}
    \caption{Comparison to few-shot methods on a challenging shape-modifying domain adaptation task. In all cases we begin with a StyleGAN-ADA \cite{Karras2020ada} model pre-trained on AFHQ-Dog \cite{choi2020starganv2}. Our method transforms the domain of the generator in a 0-shot manner. For the remaining approaches, we train using 5, 10 and 100 image sets. In low data settings, competing methods produce considerable artifacts or simply memorize the training set. Even with a hundred images, most competing methods fail to maintain correspondence between the source and target domain beyond broad attributes such as pose.}
    \label{fig:cat_supp}
\end{figure*}

We provide qualitative comparisons to few shot methods.
In \cref{fig:cat_supp} we transform the official StyleGAN-ADA \cite{Karras2020ada} AFHQ-Dog \cite{choi2020starganv2} model to the cat domain using our zero-shot method, and using few-shot models trained on samples from the AFHQ-Cat \cite{choi2020starganv2} set.

In the extreme low-data regime, competing methods either fail to produce meaningful images when faced with such a domain gap, or they simply memorize the training set. Moreover, only our method and the $100$-image variant of Ojha \etal\cite{ojha2021few} maintain consistent correspondence between images in the source and target domain.

\begin{figure*}[hbt]
\setlength{\tabcolsep}{1pt}
    \centering
    \includegraphics[width=0.95\textwidth]{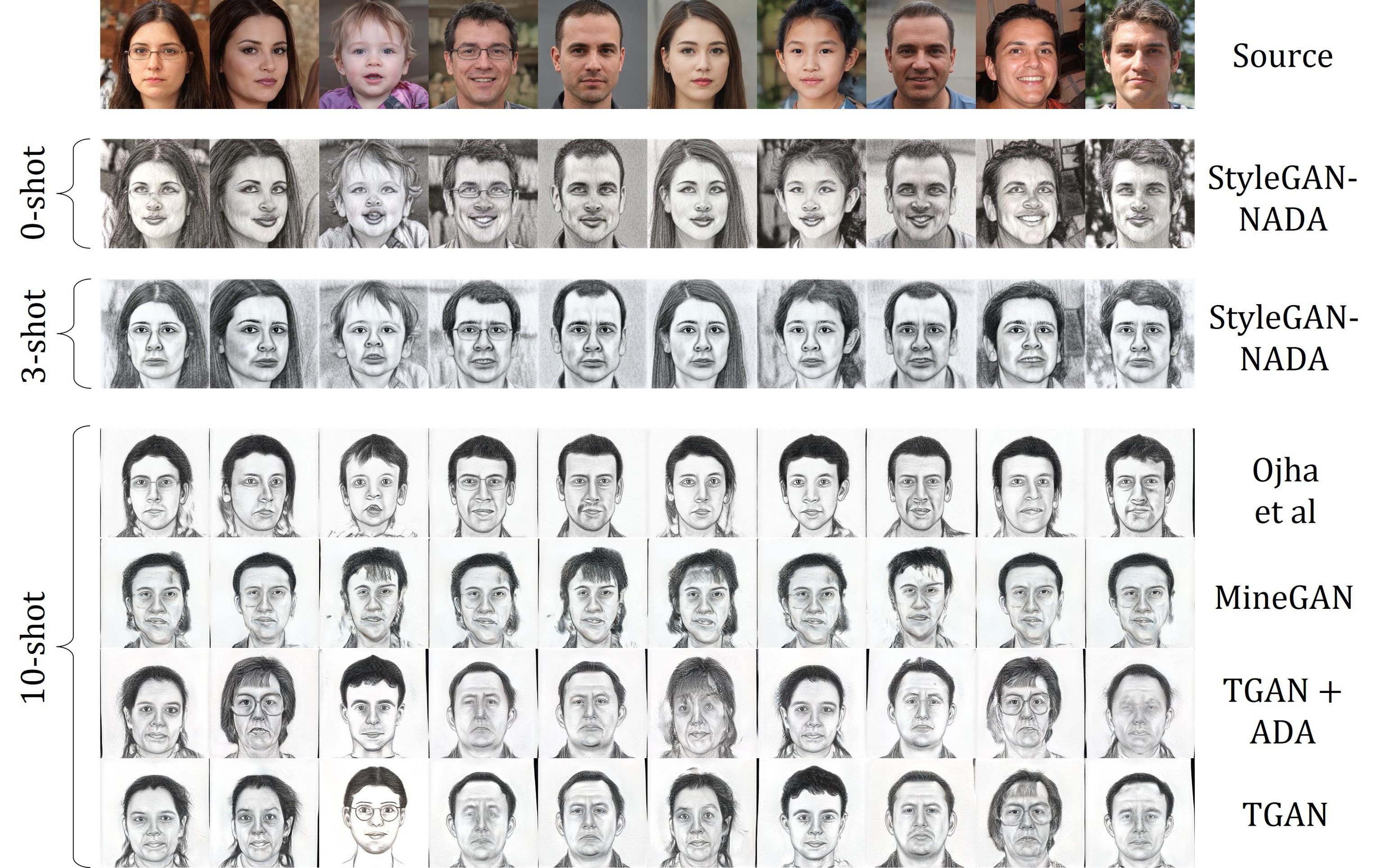}
    \caption{Comparison to few-shot methods on an image-to-sketch translation task. Our method creates high quality sketches and maintains considerably more diversity and better correspondence with the source domain. Targeting the exact artistic style using natural language prompts, however, is difficult. By guiding domain adaptation using a few samples from the set (\cref{sec:few_shot_clip}), we can adapt more features from the specific style. However, a gap still remains. }
    \label{fig:sketch_supp}
\end{figure*}

In \cref{fig:sketch_supp} we transform the official StyleGAN-ADA\cite{Karras2020ada} FFHQ\cite{karras2019style} 256x256 \href{https://nvlabs-fi-cdn.nvidia.com/stylegan2-ada/pretrained/paper-fig7c-training-set-sweeps/ffhq140k-paper256-noaug.pkl}{checkpoint} to sketches, using the few-shot set of Ojha \etal\cite{ojha2021few}. Our zero-shot transformation used the texts ``Photo'' to ``Black and white pencil sketch''. Our method maintains significantly higher quality and preserves more detail from the source domain. However, designing a textual prompt which targets the exact style portrayed in those images is difficult. By using $3$ images from the target set (\cref{sec:few_shot_clip}) rather than text, we are able to reduce the gap between the styles, but not completely eliminate it.
Even when transferring across `close' domains, most competing methods produce considerable artifacts, display significant mode collapse, or maintain very limited correspondence with the source domain.

\section{Training details}\label{sec:train_details}

\paragraph{Hyper parameter choices.}
For texture-based changes we find that a training session typically requires 300 iterations with a batch size of 2, or roughly 3 minutes on a single NVIDIA V100 GPU. In some cases ("Photo" to "Sketch") training can converge in 50 iterations and using less than a single minute, an improvement of two orders of magnitude compared to recent adversarial methods designed for training speed \cite{10.1145/3450626.3459771}.

In practice, calibrating the number of training iterations is trivial. Since training converges quickly, one may simply set the number of iterations arbitrarily high and investigate intermediate model outputs to identify the point where the model produced the most pleasing results. 

For animal changes, training typically lasts 2000 iterations. We then train a StyleCLIP mapper using the modified $G_{train}$ as a base model. The entire process takes roughly 6 hours on a single NVIDIA V100 GPU.

For all experiments, we use an ADAM Optimizer with a learning rate of 0.002. 

When using our adaptive layer-freezing approach, we set the optimization batch size $N_{w}$ to $8$, and the number of optimization iterations $N_{i}$ to $1$.
For texture-based changes we allow the network to modify all layers. For small shape changes (\eg `Human' to `Werewolf') we set the number of trainable layers $k$ to $\frac{2}{3}$ of the number of network layers (\ie $k=12$ for FFHQ). When modifying animals we set $k=3$.

We use the Vision-Transformer~\cite{dosovitskiy2020vit} based CLIP models, `ViT-32/B' and `ViT-B/16'. We observe that the two models tend to focus on different levels of detail. For texture-based changes, the choice of a model leads to minor variations in artistic styles. As such, the `optimal' choice is a matter of individual preference.
For shape changes we obtain the best results by using both models in tandem, allowing the model to better focus on both global shape and more local texture. 

In a similar manner, we observe that the choice of style mixing probability in StyleGAN2's training can affect the artistic style of the generated images. All animal change experiments were conducted without mixing. For other experiments, we report our hyperparameter choices in \cref{tab:param_choices}.

All remaining StyleGAN2 parameters were unmodified.

\begin{table}[!hbt]\setlength{\tabcolsep}{2pt}
\small
\caption{Hyper parameter choices for select models shown in the paper. See \cref{sec:train_details} for more details on parameter choices.}\label{tab:param_choices}

\centering 
\begin{tabular}{lccccc}
     Experiment & Iterations & ViT-B/32 & ViT-B/16 & Mixing & Adaptive $k$ \\ \hline
     White Walker & 200  & \checkmark & \checkmark & 0.9 & 18 \\
     Werewolf     & 300  & \checkmark & \checkmark & 0.9 & 12 \\
     Elf          & 200  & \checkmark & \checkmark & 0.9 & 18 \\
     Edvard Munch & 300  & \checkmark & \checkmark & 0.9 & 18 \\
     Sketch       & 300  & \checkmark & $\times$   & 0.0 & 18 \\
     Pixar        & 130  & \checkmark & \checkmark & 0.9 & 18 \\
     Zombie       & 150  & \checkmark & $\times$   & 0.9 & 18 \\
     Cubism       & 300  & \checkmark & $\times$   & 0.0 & 18 \\
     Princess     & 200  & \checkmark & \checkmark & 0.9 & 18 \\
     Modigliani   & 400  & \checkmark & \checkmark & 0.0 & 18 \\ 
     Shire        & 300  & \checkmark & $\times$   & 0.9 & 14 \\ 
     Nicolas Cage & 300  & \checkmark & $\times$   & 0.9 & 12 \\ 
     Cat          & 2000 & \checkmark & \checkmark & 0.0 & 3  \\
     Bear         & 2000 & \checkmark & \checkmark & 0.0 & 3  \\
\end{tabular}
\end{table}

\paragraph{StyleCLIP mapper}
As discussed in \cref{sec:experiments}, in some scenarios we use the latent mapper from StyleCLIP \cite{patashnik2021styleclip}, to better identify latent-space regions which match the target domain. Unfortunately, the mapper occasionally induces undesirable semantic artifacts on the image, such as opening an animal's mouth and enlarging tongues. We observe that these artifacts correlate with an increase in the norm of the CLIP-space embeddings of the generated images. We thus discourage the mapper from introducing such artifacts by constraining these norms by introducing an additional loss during mapper training:
\begin{equation}
    \mathcal{L}_{norm} = \left|E_{I}\left(G\left(w\right)\right) - E_{I}\left(G\left(M\left(w\right)\right)\right)\right|^2~,
\end{equation}\label{eq:norm_loss}%
where $E_I$ is the CLIP image encoder, $G$ is the fine-tuned generator, $w$ is a sampled latent code and $M$ is the latent mapper.
When training a latent mapper, we set $\lambda_{L2} = 0.5$ and $\lambda_{embedding-norm} = 0.2$.

\section{Licenses and data privacy}
\Cref{tbl:data_licenses,tbl:model_licenses} outline the models used in our work, the datasets used to train them, and their respective licenses.

The FFHQ \cite{karras2019style} dataset contains biometric data in the form of face images. Images in this set were crawled from Flickr, without reaching out to their owners. However, they were all uploaded under permissive licenses which allow free use, redistribution, and adaptation for non-commercial purposes. The FFHQ curators provide contact details for individuals that want their images removed from the set.

\begin{table}
    \small
    \centering
    \caption{Models used in our work, their sources and licenses.}
    \begin{tabular}{l c c}
        Model & Source & License \\ \hline
        StyleGAN2 & \cite{karras2020analyzing} & \nvsrc \\ 
        scikit-learn-extra & \cite{sklearn2019extra} & \bsd \\
        pSp & \cite{richardson2020encoding} & \mitlic \\
        e4e & \cite{tov2021designing} & \mitlic \\
        ReStyle & \cite{alaluf2021restyle} & \mitlic \\
        InterFaceGAN & \cite{shen2020interpreting} & \mitlic \\
        StyleCLIP & \cite{patashnik2021styleclip} & \mitlic \\
        StyleFlow & \cite{10.1145/3447648} & \ccbyncsa \\
        CLIP & \cite{radford2021learning} & \mitlic \\
        StyleGAN2-pytorch & \cite{rosinalitySG2} & \mitlic \\
        StyleGAN-ADA & \cite{Karras2020ada} & \href{https://nvlabs.github.io/stylegan2-ada-pytorch/license.html}{Nvidia Source Code License} \\
        MineGAN & \cite{Wang_2020_CVPR} & \mitlic \\
        Ojha \etal & \cite{ojha2021few} & \adblic \\
        OASIS & \cite{schonfeld2021oasis} & \href{https://github.com/boschresearch/OASIS/blob/master/LICENSE}{GNU Affero GPL}
    \end{tabular}\label{tbl:model_licenses}
\end{table}

\begin{table}
    \small
    \centering
    \caption{Datasets used in our work, their sources and licenses.}
    \begin{tabular}{l c c}
        Dataset & Source & License \\ \hline
        FFHQ & \cite{karras2019style} & \ccbyncsa \footnote{Individual images under different licenses. See \url{https://github.com/NVlabs/ffhq-dataset}}\\
        LSUN & \cite{yu2015lsun} & No License \\ 
        AFHQ & \cite{choi2020starganv2} & \ccbync \\
        Sketches & \cite{ojha2021few} & \adblic \\ 
        COCO-Stuff & \cite{caesar2018coco} & \href{https://creativecommons.org/licenses/by/4.0/}{CC BY 4.0}
    \end{tabular}
    \label{tbl:data_licenses}
\end{table}

\end{document}